%% file: main.tex
\documentclass[10pt,twocolumn,letterpaper]{article}

\usepackage[pagenumbers]{cvpr} 

\usepackage{graphicx} \graphicspath{{figures/}}
\usepackage[utf8]{inputenc} 
\usepackage[T1]{fontenc}    
\usepackage{amsmath, amsthm, amssymb, amsfonts, bbm}
\usepackage{microtype}      
\usepackage{pifont}
\usepackage{url}            
\usepackage{subcaption}
\usepackage{verbatim}
\usepackage[ruled, vlined]{algorithm2e}
\usepackage{algorithmic}
\usepackage{booktabs}       
\usepackage{multirow, multicol}
\usepackage[most]{tcolorbox}
\usepackage{balance}
\usepackage{xargs}

\definecolor{datasheetcolor}{rgb}{0.2,0.08,0.45}
\newcommand{\datasheetsection}[1]
{\vspace{0.5em}\noindent\textcolor{datasheetcolor}
{\fbox{\parbox{0.97\linewidth}{\centering\textsf{\large\textbf{#1}}}}}\vspace{0.8em}}
\newenvironmentx{datasheetitem}[2][usedefault,2={}]{\footnotesize\noindent\ignorespaces\textcolor{datasheetcolor}{\textsf{\textbf{#1} #2}}\par\normalsize}{\par\vspace{0.8em}\noindent\ignorespacesafterend}

\makeatletter 
\renewcommand{\paragraph}{\@startsection{paragraph}{4}
{\z@}{1.25ex \@plus 1ex \@minus .2ex}{-1em} 
{\normalfont\normalsize\bfseries}} 
\makeatother

\newcommand{\cmark}{\ding{51}}
\newcommand{\xmark}{\ding{55}}
\newcommand{\TeamCraft}{\textsl{TeamCraft}\xspace}

\definecolor{cvprblue}{rgb}{0.21,0.49,0.74}
\usepackage[pagebackref,breaklinks,colorlinks,allcolors=cvprblue]{hyperref}
\usepackage{wrapfig}
\def\confName{CVPR}
\def\confYear{2025}

\title{TeamCraft: A Benchmark for Multi-Modal Multi-Agent Systems in Minecraft}

\author{
    Qian Long\textsuperscript{\rm 1}\stepcounter{footnote},
    Zhi Li\textsuperscript{\rm 1}, 
    Ran Gong\textsuperscript{\rm 1},
    Ying Nian Wu\textsuperscript{\rm 1},
    Demetri Terzopoulos\textsuperscript{\rm 1},
    Xiaofeng Gao\textsuperscript{\rm 2}\thanks{~This work does not relate to the author's position at Amazon.}~
    \\[10pt]
    \textsuperscript{\rm 1}University of California, Los Angeles\\ \textsuperscript{\rm 2}Amazon AGI \\
    {\tt teamcraftbench@gmail.com} \\
    {\tt \url{https://teamcraft-bench.github.io/}}
}

\begin{document}
\maketitle
\begin{abstract}
Collaboration is a cornerstone of society. In the real world, human teammates make use of multi-sensory data to tackle challenging tasks in ever-changing environments. It is essential for embodied agents collaborating in visually-rich environments replete with dynamic interactions to understand multi-modal observations and task specifications. To evaluate the performance of generalizable multi-modal collaborative agents, we present \TeamCraft, a multi-modal multi-agent benchmark built on top of the open-world video game Minecraft. The benchmark features 55,000 task variants specified by multi-modal prompts, procedurally-generated expert demonstrations for imitation learning, and carefully designed protocols to evaluate model generalization capabilities. We also perform extensive analyses to better understand the limitations and strengths of existing approaches. Our results indicate that existing models continue to face significant challenges in generalizing to novel goals, scenes, and unseen numbers of agents. These findings underscore the need for further research in this area. The \TeamCraft platform and dataset are publicly available at\\ \footnotesize\url{https://github.com/teamcraft-bench/teamcraft}
\end{abstract}

\section{Introduction}

Developing collaborative skills is essential for embodied agents, as collaboration is a fundamental aspect of human intelligence \cite{smith2005development}. In the AI community, multi-agent collaboration is frequently studied using grid-world environments \citep{leibo2021meltingpot, nmmo_neurips, stone2000multiagent, gong2023mindagent, dong2024villageragent, puig2021watchandhelp, park2023generative, zhang2024proagent, wu2021too,long2024inverseattentionagentmultiagent}. However, agents in these environments lack multi-modal understanding. By contrast, learning within visually-rich environments enables agents to develop useful representations of multi-agent dynamics \cite{chen2020visual, jaderberg2019human}, as vision facilitates implicit communication, coordination, and collaborative execution \cite{jain2020cordial, jain2019body}.

\begin{table*}
\centering
\caption{Comparison with other benchmarks. \TeamCraft features visual observation for multi-agent control with widely-varied tasks specified by multi-modal prompts, targeting various types of generalization essential for multi-agent teaming. \textbf{MM Spec.}: multi-modal task specification. \textbf{Observation}: \textbf{V} for visual observation and \textbf{S} for state-based observation. \textbf{MA}: multi-agent control, \textbf{C} for centralized and \textbf{D} for decentralized. \textbf{Interaction}: object interaction. \textbf{Tool}: tool use. \textbf{Generalization}: types of generalization targeted, \textbf{E} for generalization on novel environments or scenes, \textbf{G} for novel goals, \textbf{A} for novel numbers of agents. \textbf{\# Variants}: number of task variants involved. }
\resizebox{\linewidth}{!}{
\begin{tabular}{lccccccccccc}
\toprule
Benchmark & MM Spec. & 3D & Observation & MA & Interaction & Tool & Generalization & \# Agents & \# Variants & \# Demonstrations \\
\midrule
ALFRED~\citep{shridhar2020alfred} & \xmark & \cmark & V & \xmark & \cmark & \cmark & E & 1 & 2,600+ & 8,000+ \\
FurnMove~\citep{jain2020cordial} & \xmark & \cmark & V & CD & \cmark & \xmark & E & 2 & 30 & \xmark \\ 
Marlo~\citep{perez2019multi} & \xmark & \cmark & V & D & \cmark & \xmark & \xmark & 4+ & 14 & \xmark \\
MineDojo~\citep{fan2022minedojo} & \xmark & \cmark & V & \xmark & \cmark & \cmark & EG & 1 & 3,000+ & 740,000+ \\ 
MindAgent~\citep{gong2023mindagent} & \xmark & \cmark & VS & C & \cmark & \cmark & \xmark & 4+ & 39 & \xmark \\
Neural MMO 2.0~\citep{suarez2024neural} & \xmark & \xmark & S & CD & \cmark & \cmark & EGA & 128+ & 25+ & \xmark \\
Overcooked-AI~\citep{carroll2020utility} & \xmark & \xmark & VS & C & \cmark & \cmark & \xmark & 2 & 5 & 80 \\
PARTNR \cite{chang2024partnr} & \xmark & \cmark & VS & CD & \cmark & \cmark & E & 2 & 100,000+ & 100,000+ \\
RoCoBench~\citep{mandi2024roco} & \xmark & \cmark & S & CD & \cmark & \cmark & G & 2 & 6 & \xmark \\
VIMA-Bench~\citep{jiang2022vima} & \cmark & \cmark & V & \xmark & \cmark & \cmark & EG & 1 & 1,000+ & 600,000+ \\
Watch\&Help \citep{puig2021watchandhelp} & \xmark & \cmark & S & CD & \cmark & \cmark & EG & 2 & 1,200+ & 6,300+ \\
\midrule
\textbf{\TeamCraft} & \cmark & \cmark & VS & CD & \cmark & \cmark & EGA & 4+ & 55,000+ & 55,000+ \\
\bottomrule
\end{tabular}
}
\label{tab:comparison}
\end{table*}

Learning vision-based, multi-task, multi-agent systems is a challenging objective that presents several difficulties. These systems must develop detailed scene understanding to handle the diverse visual appearances of scenes. The complexity is further heightened by the numerous combinations of task configurations, such as object spatial arrangements, goal configurations, arbitrary numbers of agents, and heterogeneous agent capabilities. Consequently, it is essential for multi-agent systems to acquire generalizable skills that can be effectively transferred across different settings.

An important step in addressing these challenges is to develop simulation systems that support multi-modal multi-agent learning. Recent advances in simulated environments have significantly facilitated progress in embodied vision-based  systems \cite{yu2024mineland, jain2020cordial, chen2020visual, perez2019multi, das2019tarmac}. Despite notable progress, these systems have several limitations: (1) many of them target one or two-agent scenarios \cite{jain2019body, mandi2024roco, wang2023voyager}, (2) they are often limited to indoor settings with a narrow range of tasks \cite{puig2021watchandhelp, zhang2024combo}, and (3) the task specifications are generally purely in text \cite{liu2022embodied, mandi2024roco}, making it hard to specify subtle task differences accurately and efficiently.

To drive progress in this area, we have developed a comprehensive benchmark, named \TeamCraft, that features procedurally generated large-scale datasets specifically designed for multi-modal multi-agent systems. This benchmark utilizes the widely acclaimed open-world video game Minecraft as an experimental platform to engage with the complex dynamics of multi-modal multi-agent interactions. Inspired by the work of \cite{jiang2022vima}, we also leverage multi-modal prompts as task specifications to guide agent interactions, as language often fails to effectively convey spatial information \cite{cai2024rocket}. Our benchmark offers rich visual backgrounds, diverse object categories, complex crafting sequences, and varying task dynamics. These features enable systematic exploration of out-of-distribution generalization challenges for multi-modal, multi-task, multi-agent systems at scale. In particular, our benchmark evaluates a model's ability to generalize to novel goal configurations, unseen number of agents, novel agent capabilities, and new types of visual backgrounds. To evaluate existing techniques using our benchmark, we have designed several baseline models to work within the framework and compare their performance. Our results highlight that current approaches to vision-conditioned collaboration and task planning encounter significant challenges when tested within \TeamCraft's complex and dynamic environment, especially when it comes to generalizations.

In summary, the main contributions of this paper are:
\begin{enumerate}
\item \TeamCraft, a new multi-modal multi-agent benchmark with its associated large-scale dataset encompassing complex tasks challenging multi-agent systems in a wide variety of generalization scenarios.
\item  Extensive experiments and analyses on state-of-the-art multi-modal multi-agent models, uncovering their strengths and weaknesses to inform and inspire future research.
\item To ensure reproducibility and encourage future works from the community, we open source the entire platform, its training and evaluation code, and release the model checkpoints and training data at\\ {\footnotesize{\url{https://github.com/teamcraft-bench/teamcraft}}}
\end{enumerate}

\section{Related work}
\label{gen_inst}

\subsection{Platforms for Multi-Agent Systems}

The recent success of multi-agent reinforcement learning (MARL) methods \citep{lowe2020multiagent, yu2021surprising, long2020evolutionary, long2024socialgfs} has attracted attention, as these methods explore cooperation and competence behaviors among agents. However, many of the methods are evaluated in simplified 2D environments \citep{leibo2021meltingpot, nmmo_neurips, mordatch2017emergence, vinyals2019grandmaster, carroll2020utility}. Recent work on embodied multi-agent benchmarks has considered more realistic tasks and environments \cite{liu2022multi, liu2022embodied, gong2023lemma, park2023generative, chang2024partnr}, but it often relies on certain privileged sensor information of the environment \citep{zhang2024building, puig2021watchandhelp, puig2023nopa}. Additionally, subject to environmental constraints, these works often have limited set of tasks \cite{jain2019body, tan2020multi} related to navigation and simple interactions such as object rearrangement \cite{szot2021habitat}. By comparison, \TeamCraft is based on Minecraft, a three-dimensional, visually rich open-world realm characterized by procedurally generated landscapes and versatile game mechanics supporting an extensive spectrum of object interactions, providing rich activities ripe for intricate collaborations.

\subsection{Embodied Language-Guided Benchmarks}

Several researchers have looked at the problem of using natural language as the interface between embodied agents, either in the form of task specifications \citep{shridhar2020alfworld, shridhar2020alfred, zheng2022vlmbench, gong2023arnold}, question answering \citep{das2018embodied, gordon2018iqa, ma2023sqa3d, majumdar2024openeqa}, instruction following \citep{anderson2018vision, narayan2019collaborative, jayannavar2020learning, gao2022dialfred, padmakumar2022teach, wan2022handmethat, gao2023alexa}, or as means of task coordination \citep{li2023camel, mandi2024roco}. VIMA-Bench \citep{jiang2022vima} builds on previous efforts in language-guided robotic manipulation \citep{zeng2020transporter, shridhar2021cliport, mees2022calvin} and uses multi-modal prompts as uniform task specifications for object manipulation. \TeamCraft extends multi-modal prompts to the multi-agent domain and uses them to specify a wide variety of collaborative tasks that require object interaction and navigation. 

\begin{figure}
    \centering
    \includegraphics[width=\linewidth]{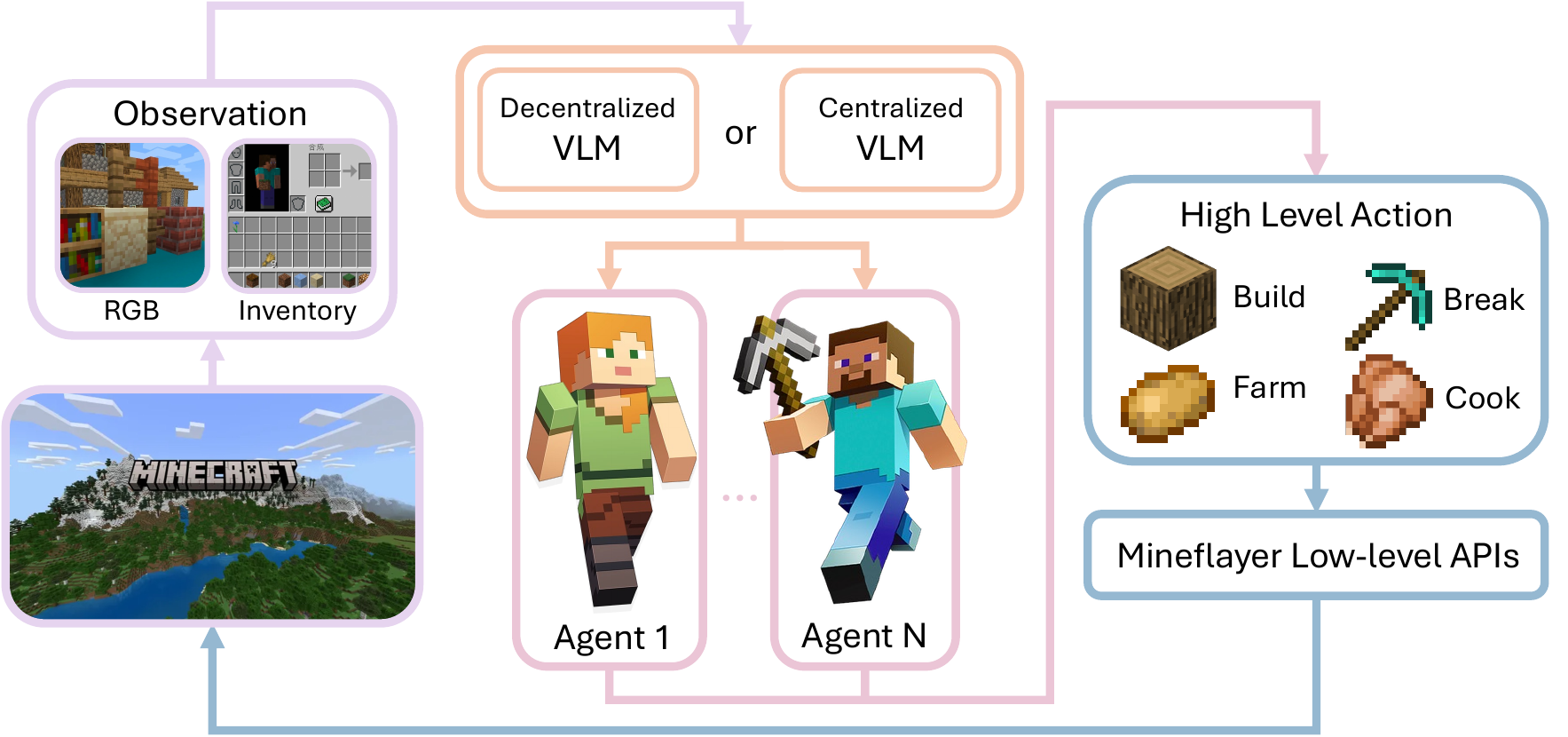}
    \caption{The \TeamCraft platform consists of three main components: (1) a Minecraft server that hosts the game as an online platform, (2) Mineflayer, which serves as the interface for controlling agents in the server, and (3) a Gym-like environment that provides RGB and inventory observations to the models, allowing control of multiple agents through high-level actions.}
    \label{fig:MAMC-arch}
\end{figure}

\begin{figure*}
    \centering
        \includegraphics[width=\linewidth]{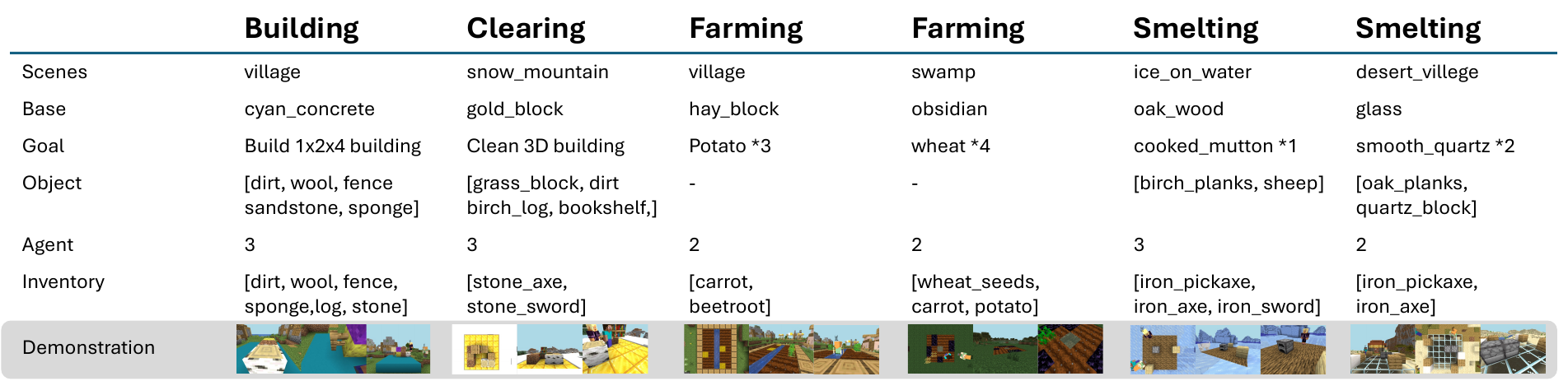}
        \caption{We present example task configurations, as a combination of distinct biomes, playground base blocks, task goals, target blocks materials and agent counts. Agents are initialized with unique inventories, which provide them with different capabilities to complete various activities. A detailed distribution is provided in \autoref{appendix:stat_table}.}
        \label{fig:task_all}
\end{figure*}

\begin{figure*}
    \centering
    \includegraphics[width=\linewidth]{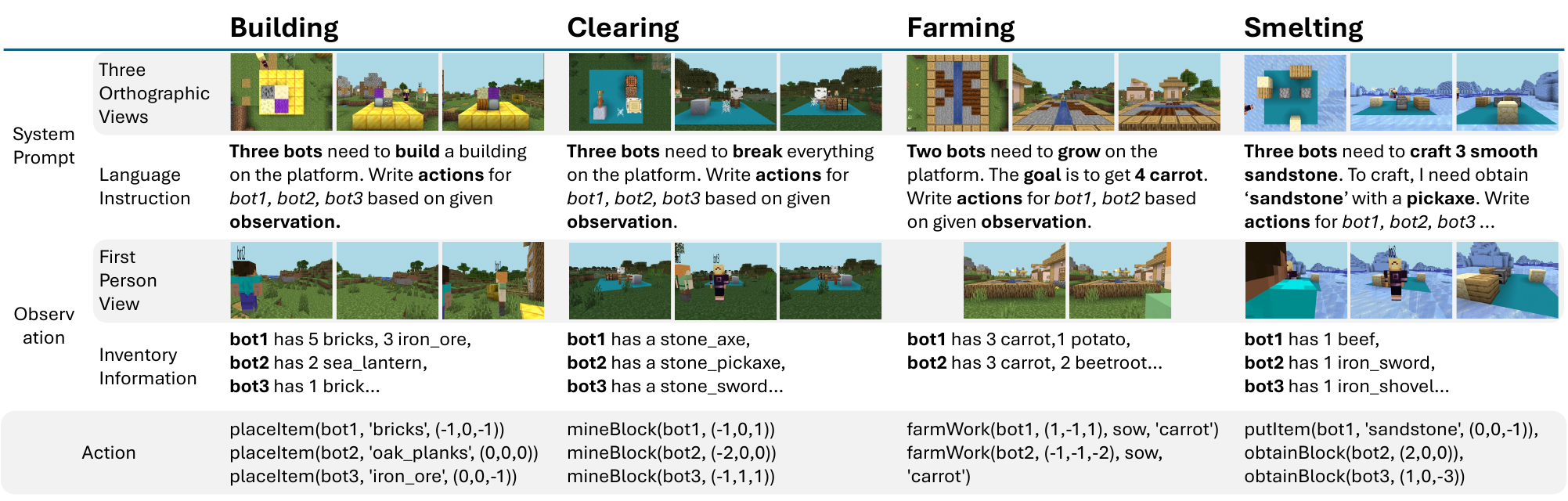}
    \caption{Multi-modal prompts are provided for all tasks. The system prompt includes both the three orthographic views and specific language instructions. Observations consist of first-person views from different agents, along with agent-specific information.}
    \label{fig:mm-prompt}
\end{figure*}

\subsection{Benchmarks Based on Minecraft}

Malmo \citep{johnson2016malmo} marks the advent of a Gym-style platform tailored to Minecraft games. It paves the way for subsequent single-agent works such as MineRL \citep{guss2019minerl}, Voyager \cite{wang2023voyager}, and MineDojo \citep{fan2022minedojo}. Marlo \citep{perez2019multi} extends Malmo to multi-agent scenarios, but the small number of task variations limit generalizations. Similar to our work, MindAgent \citep{gong2023mindagent} and VillagerBench \citep{dong2024villageragent} focus on multi-agent collaboration in a multi-task setting. However, both of these use purely state-based observations, while \TeamCraft tackles the more challenging problem of learning to collaborate from multi-modal perceptions. \autoref{tab:comparison} compares \TeamCraft with prior benchmarks.

\section{TeamCraft Benchmark}

\subsection{Problem Formulation}

Assume that an embodied multi-agent system comprised of $N$ agents needs to complete a complex task involving navigation and object manipulation. The task is specified in a multi-modal prompt $x_L=\{x_{l}\}_{l=1}^{L}$, which is a sequence of interleaved language and image tokens with length $L$. At time step $t$, each agent receives partial observation $o_n^t \in O$ from the full observation space $O$. To complete the task, each agent can choose to perform a high level action $a_t \in A$ from the full set of action $A$. The action can be further decomposed into a sequence of low level control signals.   

\subsection{Simulation Environment}

\TeamCraft utilizes Minecraft as its foundational simulation environment, offering a complex, open-world setting for multi-agent interactions. With a Gym-like environment, it facilitates the execution of intricate multi-agent commands via self-explanatory skills. \autoref{fig:MAMC-arch} illustrates the platform architecture. High level skills from the model can be translated into low level control signals via nested API calls through Mineflayer\footnote{\url{https://github.com/PrismarineJS/mineflayer}}. After execution, visual observation of each agent are rendered and provided as input to the model. 

\paragraph{Multi-Modal Prompts:} 
In our work, the multi-modal prompt $x_L$ consisting of a language instruction interleaved with a set of orthographic projection images (i.e., top, left, and front views) for task specification. Depending on the specific task, the images can specify either the initial states, intermediate states or the goal states. 

\paragraph{Observation and Actions:} 
To mimic real world settings of embodied visual agent teaming, we use first-person view RGB image and inventory information as the observation $o_n$. The action space $A$ involves high-level self-explanatory skills such as \textit{obtainBlock} to obtain a block and \textit{farmWork} to farm a crop. Most actions take three input parameters, including (1) agent name such as \textit{bot1}, as the action-executing entity, (2) item name such as \textit{dirt}, and (3) a 3D vector indicating the position of the target. There are 8 types of actions in total. A complete list of actions are described in \autoref{appendix:high_level_skills}.

\subsection{Task Design} 

\TeamCraft introduces a variety of complex and interactive tasks that challenge the agents' capabilities in planning, coordination, and execution within a collaborative and dynamic environment. Each task is designed to test different facets of MA interaction, including role distribution, real-time decision-making, and adaptability to changing environments. Task examples are shown in \autoref{fig:task_all} and the corresponding prompt examples are shown in \autoref{fig:mm-prompt}.

\paragraph{Building:} 
Agents erect a structure based on a provided orthographic view blueprint. Each agent possesses a unique inventory of building blocks necessary for the construction. Successful completion requires visual cognition to associate blueprint components with inventory items, spatial reasoning to reconstruct a 3D structure from 2D images and map it to 3D coordinates for action targets, and collaborative coordination with other agents to resolve action dependencies. For example, an agent cannot place a floating block and should wait for another agent to build the supporting block first.

\paragraph{Clearing:} 
Agents are required to remove all blocks from a specified area. Besides spatial understanding and awareness of action dependencies, agents must employ appropriate tools to break blocks, which vary in durability, thereby requiring multiple interactions for complete removal. The use of correct tools can dramatically reduce the time required to remove blocks. Thus agents must coordinate task assignments to optimize block-breaking efficiency. Strategic coordination is essential in this task as agents need to dynamically decide which blocks to target based on their current tools, and assist each other even without the optimal tools when necessary.

\paragraph{Farming:} 
Agents sow and harvest crops on designated farmland plots. They must monitor crop growth stages, from newly planted to fully grown, and harvest only when crops reach maturity. Efficient task completion requires spatial reasoning to select appropriate farmland, visual cognition to assess crop maturity, and continuous updating of farmland states based on other agents' actions. As the available farmland exceeds what is needed, understanding other agents' actions to avoid redundancy, and dynamically allocating sub-tasks based on positions, available seeds, and crop maturity are essential. For example, some agents can sow while others are harvesting, stop when the total yield meets the goal.

\paragraph{Smelting:} 
Agents obtain processed items using furnaces by gathering materials and coordinating actions. They collect resources from the environment, by harvesting blocks or killing mobs, or use existing inventory items to produce goal items like cooked food or refined materials. Agents also need to gather fuel before they can make use of furnaces. Efficient task completion requires spatial understanding to locate furnaces and resources, and coordinating actions with inter-agent dependencies. For instance, if one agent is collecting beef, others should focus on gathering fuel rather than duplicating efforts. Working as a team to use limited furnaces efficiently is crucial, rather than each agent independently smelting their own goal item.

\subsection{Centralized and Decentralized Agents}

\TeamCraft supports centralized and decentralized control.

\paragraph{Centralized Agents:} 
The centralized model is given the observational data of all agents, including the first-person view, action history, and inventory information. Based on these comprehensive data, the model generates the actions for all agents simultaneously. This approach leverages the full scope of information available in the environment to coordinate and optimize the actions of all agents collectively.

\paragraph{Decentralized Agents:} 
The decentralized models do not receive information about other agents except for the initial inventory of the team. Each model generates actions solely for the individual agent based on its limited view. This setting simulates a more realistic scenario where agents operate independently with restricted information, focusing on their actions absent of any centralized coordination.

\subsection{Diversity}

The tasks are complex and challenging, testing multi-agent systems in diverse settings. \autoref{table:diversity} provides task statistics and variants, with visual diversity detailed in \autoref{appendix:visual_diversity}.

\paragraph{Object Diversity:}
More than 30 target object or resource are used in tasks. Objects, such as a fence, an anvil, or a stone block, have different shapes and textures. Farm crops have different visual appearances during growth stage. The smelting task has resources with different appearances, such as chickens or rabbits.

\paragraph{Inventory Diversity:}
Agent's inventory might include essential items mixed with non-essential ones (i.e., distractors), realistically simulating scenarios where agents must choose the right materials for specific tasks while managing inventory constraints. Agents are also provided with random tools for each task. Having the appropriate tools significantly enhances efficiency in tasks like clearing. For smelting, some resources must be collect by agent with specific tools.

\paragraph{Scene Diversity:}
More than 10 scenes are included in the tasks, covering biomes such as village, mountain, forest, swamp, desert, etc. 
Tasks take place on grounds with diverse textured bases such as glass, concrete, and quartz. Certain tasks may involve additional complexity, including farmland which are intermixed with non-plantable blocks.

\paragraph{Goal Diversity:}
Each task requires achieving a varying number of goal targets. Building requires different blocks placed into various shapes, categorized based on dimensionalities, e.g., 2D (all blocks are at the same level) or 3D (some blocks are on top of others). Farming requires various target crops and yields. For the smelting task, the target object is sampled from various food or processed items. 

\subsection{Tasks and Expert Demonstrations Generation}

To create a rich learning environment and effective imitation learning dataset, systematic scenario design and data collection methods are employed, as follows:

\paragraph{Task Generation:} 
Variables from a diversity pool, such as agent counts, scenes, and goals, are sampled to establish task configurations. Specifically, a solvable task is formulated by rejection sampling of the essential task variables. "Solvable" implies that the task can be completed within the Minecraft world rules and is within the agents' capabilities. For example, in smelting tasks, fuel must either be available to collect in the scene or directly accessible in the inventory. 
 
\paragraph{Planner-Based Demonstrations Generation:} 
Given the task specifications, a planner assigns actions to agents at every time step, utilizing privileged information of the environment. Assume agent $i$ performing action $j$, the planner optimizes a cost function designed to minimize total task completion time $T$, idle actions $E_i$, action dependencies $D$, redundant actions $U$, and the cost $c_{ij}$ for agent $i$ performing action $j$:
\begin{equation}
    C = w_1 T + w_2 \sum_{i=1}^N E_i + w_3 D + w_4 \sum_{i=1}^N \sum_{j \in A_i} c_{ij} + w_5 U
\end{equation}
where $w_1$, $w_2$, $w_3$, $w_4$, and $w_5$ are weighting coefficients. Details of the weights are available in \autoref{appendix:planner_details}.

\begin{table}
\centering
\caption{Task variants and dataset statistics}
\label{table:diversity}
\resizebox{\linewidth}{!}{
\begin{tabular}{lcccc}
\toprule
                       & \textbf{Building} & \textbf{Clearing} & \textbf{Farming} & \textbf{Smelting} \\ 
\midrule
\# Action Sequences     & 2 -- 6             & 2 -- 9             & 2 -- 7            & 2 -- 8             \\
\# Agents               & 2 -- 3             & 2 -- 3             & 2 -- 3            & 2 -- 3             \\
\# Tools                & --                 & 1 -- 4             & --                & 1 -- 4             \\
\# Scenes               & 6                 & 5                 & 4                & 5                 \\
\# Base Types           & 10                & 11                & 9                & 11                \\
\# Furnaces             & --                 & --                 & --                & 1 -- 2             \\
\# Target Block Types         & 19                & 16                & 3                & 13                \\
\# Target Block Counts        & 5 -- 12            & 4 -- 9             & 2 -- 14           & 1 -- 4             \\
\# Fuel Types           & --                 & --                 & --                & 12                \\
\# Resource Types       & --                 & --                 & --                & 20                \\
\# Dimensional Shapes   & 2                 & 2                 & 2                & 1                 \\
\# Placement Shapes     & 7715              & 12724             & 13188            & 8885              \\
\# Total Demonstrations & 14998             & 14641             & 14815            & 10803             \\
\# Test Set              & 50               & 50               & 50                & 50               \\
\# Generalization Set   & 200              & 200               & 150               & 200               \\
\# Generalization Conditions   & 4              & 4               & 3               & 4               \\
\bottomrule
\end{tabular}
}
\end{table}

As shown in \autoref{table:diversity}, we generated 55,000 unique task variants, each with one demonstration. A demonstration consists of a multi-modal prompt as task specification, including three orthographic view images representing task initial states or goal states and the corresponding language instructions. At each time step, agent inventories, first-person RGB observations and actions are recorded.

\begin{figure*}
    \centering
    \includegraphics[width=\linewidth]{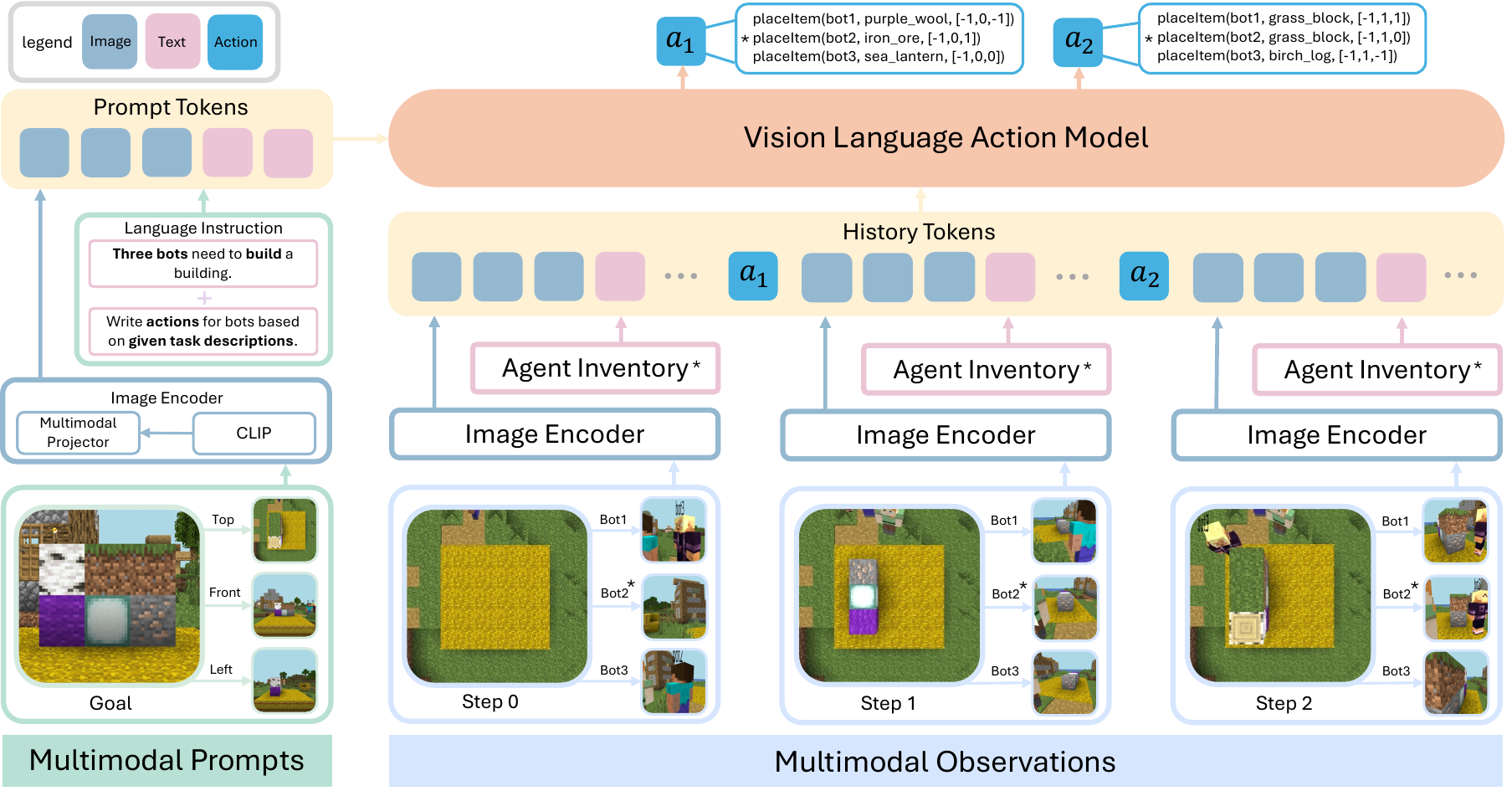}
    \caption{The architecture of the TeamCraft-VLA model. Multi-modal task specifications combining three orthographic views images of the task goal states and corresponding language instructions are encoded as initial input to the model. Agents inventories and visual observations are further encoded in each step to generate actions for agents. For decentralized setting, the model only has access to one agent's information, exampled by Bot2: items associated with a \textbf{*} represent the fact that only the data associated with agent 2 are available. }
    \label{fig:model-arch}
\end{figure*}

\subsection{Test Set and Generalization Set}

\TeamCraft features a test set, where agents are initialized with random position, orientation, and inventory. Other variables follow the same distribution as training. To evaluate the model generalization, we further designed a generalization set with hold-out elements excluded from training data. In general, we withheld test cases involving \textit{four agents}, whereas the training data include only two or three agents. We also introduced unseen \textit{scenes} not present during training. In addition to these general hold-outs, we implemented task-specific exclusions as following: 1) Building: novel \textit{shapes} and \textit{materials} to build. We exclude 8 block placement \textit{shapes}, defining how target blocks are arranged on the ground. These shapes varied in complexity, containing 5 to 12 blocks in both 2D and 3D configurations. Additionally, we omitted 3 block \textit{materials} appeared in clearing but not in building. 2) Clearing: novel \textit{shapes} and \textit{materials} to clear. We held out 6 block placement \textit{shapes} with block counts ranging from 4 to 9. We also excluded 3 block \textit{materials} present in building but absent in clearing. 3) Farming: novel \textit{crops} to farm and collect. 4) Smelting: novel number of \textit{furnaces} and \textit{goal} objects. We excluded 4 unseen \textit{goal} objects and introduced scenarios with novel number of \textit{furnaces} in the scene. As shown in \autoref{table:diversity}, with 50 samples per task for the test set and each generalization condition, our benchmark contains a total of 950 test cases.

\section{Experiments}

\subsection{Baselines and Ablations}

\paragraph{TeamCraft-VLA:} 
We introduce TeamCraft-VLA (Vision-Language-Action), a multi-modal vision-language action model designed for multi-agent collaborations. As shown in \autoref{fig:model-arch}, the model first encodes multi-modal prompts specifying the task, then encodes the visual observations and inventory information from agents during each time step to generate actions. Following \cite{liu2024visual}, the VLA model architecture consisting of a CLIP encoder for images, a projector to align the image features with language model. We use CLIP ViT-L/14 as the visual encoder and a linear projector for modality alignment. The model is trained on the demonstration data for three epochs before convergence. 

\paragraph{Proprietary VLA:} 
We use GPT-4o as the proprietary VLA. Specifically, we use similar prompt structures as the centralized finetuned TeamCraft-VLA model, with additional task information in the initial system prompt to provide background knowledge of the task. The system prompt contains recipes, input, output formats, all available blocks, items, workspace limitations, and one successful rollout of a similar task in the same task family. At the first step, we additionally provide the first user prompt, where the model is given a specific multi-modal task specification accompanied by initial visual observations and inventory details of the agents. Based on the system prompts and user prompts, the model predicts the actions. As the interaction progresses with subsequent prompts, the context is maintained and expanded with the addition of prior responses and updated visual data. Detailed prompts are available in \autoref{appendix:gpt4o}. 

\paragraph{Grid-World Settings:}
In order to the understand the impact of learning in multi-modal environment as opposed to purely text-based or state-based environment, we perform an ablation study by translating the \TeamCraft environments into a 3D grid-world. We retain the same prompt structure of the training data used in the TeamCraft-VLA models, with the main difference being that environmental information (i.e. visual observations and three orthographic view images) is now represented in text, describing the voxel coordinate of each block, e.g. "brick is at (2,3,0), stone is at (2,3,1)...".  We fine-tuned a LLM in centralized setting with variance in the dataset size (10\%, 50\%, and 100\% of the total data) for three epochs before convergence.  

\begin{figure*}
    \centering
    \includegraphics[width=1.0\linewidth]{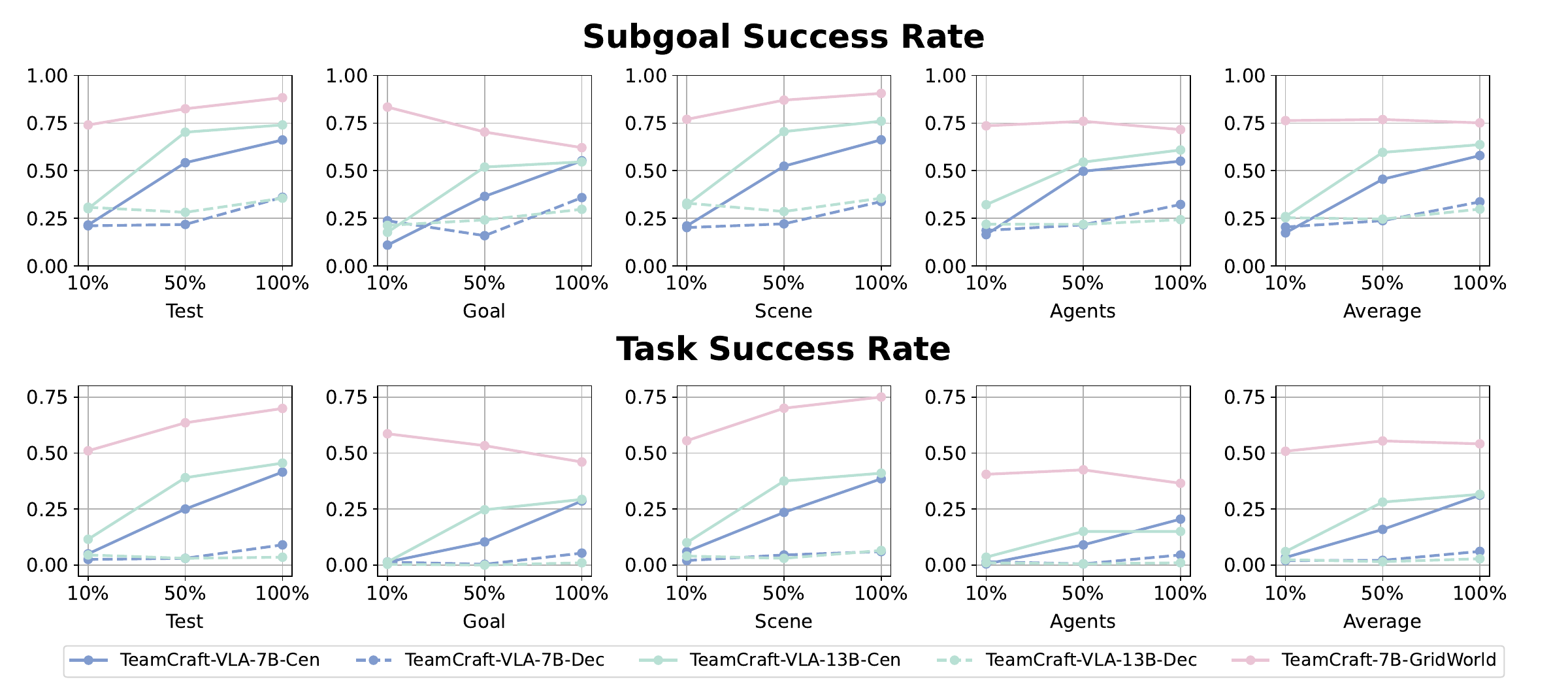}
    \caption{Subgoal success rate and task success rate across centralized, decentralized and grid-world settings. The leftmost column displays the \textit{Test} category, which shares similar data distribution as training. The \textit{Goal}, \textit{Scene} and \textit{Agents} categories represent generalization tasks involving unseen goals, scenes, and tasks involving four agents, respectively. Average performance is presented in the rightmost column.}
    \label{fig:data-scale}
\end{figure*}

\paragraph{Ablations:} 
We performed a total of 15 ablation studies, varying in dataset sizes (10\%, 50\%, and 100\% of the total data), control settings (centralized and decentralized), experiment settings (Multi-modal and Grid-World) and sizes of the VLA model (7B and 13B).

\subsection{Evaluation Metrics}

We evaluated the performance of the methods based on three key metrics: task success rate, subgoal success rate and redundancy rate. 

\paragraph{Subgoal Success Rate:} 
This metric evaluates the effectiveness of agents in completing tasks. Given $M$ test cases, each test case $m$ has $s^g_m$ subgoals, and agents complete $s^d_m$ subgoals. The subgoal success rate $SGS$ is defined as
\begin{equation}
    SGS = \frac{1}{M} \sum_{m=1}^{M} \frac{s^d_m}{s^g_m}
\end{equation}
Specifically, subgoals are designed based on the task requirements, i.e. the number of blocks to be built for building and the number of target objects to be created for smelting. 

\paragraph{Task Success Rate:} 
This metric indicates the proportion of test cases that the model can successfully complete from start to finish. Specifically, the task success rate $TS$ is defined as: 
\begin{equation}
    TS = \frac{1}{M} \sum_{m=1}^{M} \mathbbm{1} \left[ s^d_m = s^g_m \right]
\end{equation}
A higher success rate reflects the model's ability to consistently achieve the desired outcomes in various scenarios.

\paragraph{Redundancy Rate:} 
This metric assesses whether multiple agents are performing the same action at the same time, which would lead to conflicts. Assume $p_m$ is the total number of actions for test case $m$ and $q_m$ the number of conflicts between agents, the redundancy rate $RR$ is defined as: 
\begin{equation}
    RR = \frac{1}{M} \sum_{m=1}^{M} \frac{q_m}{p_m}
\end{equation}
A lower redundancy rate indicates better task allocation among agents and a higher level of cooperative efficiency.

\subsection{Evaluation Results}

We evaluated the subgoal success rate and task success rate of the models. As illustrated in \autoref{fig:data-scale}, 
our analysis and findings are discussed below:

\paragraph{Success Rate:} 
For both the 7B and 13B models, the subgoal success rate and task success rate fall short of optimal performance. This is particularly evident in challenging tasks such as smelting, with both subgoal and task success rates below $40\%$. This highlights the inherent difficulty of the designed tasks and underscores the current limitations of VLA models in handling multi-step, sequentially dependent processes.

\paragraph{Across Model Size:} 
In \autoref{fig:data-scale}, we observe that as training data increased, the performance of the 7B model approaches that of the 13B model, especially when generalizing to novel goals and number of agents. This suggests that scaling up model sizes blindly do not guarantee success. 

\begin{figure*}
    \centering
    \includegraphics[width=\linewidth]{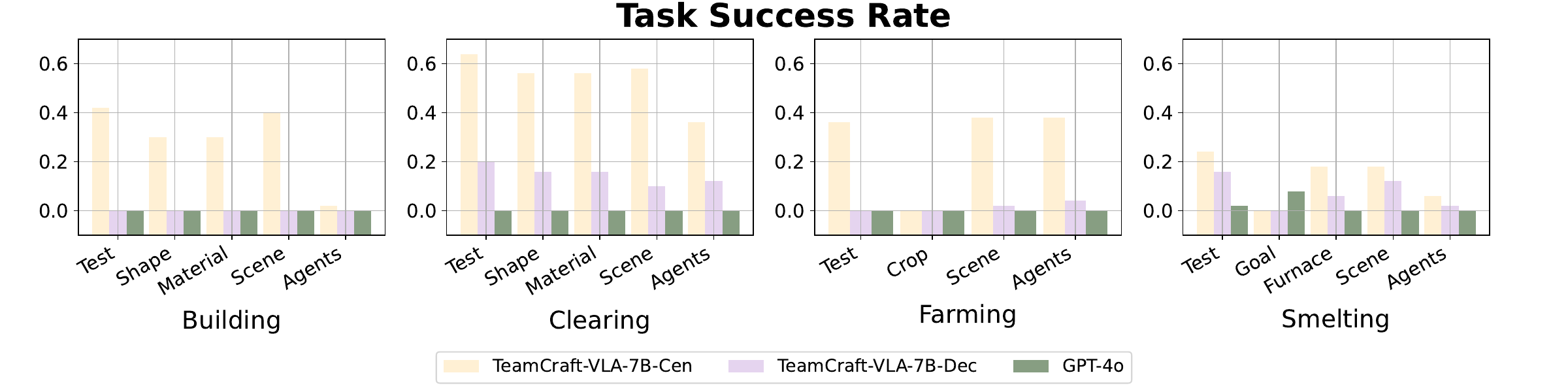}
    \caption{Task success rates of centralized and decentralized VLA model and GPT-4o. Models are trained with the full data except GPT-4o. TeamCraft-VLA-7B-Cen outperforms the other two methods by a significant margin across nearly all variants.}
    \label{fig:success-rate}
\end{figure*}

\paragraph{Multi-Modal Environment vs. Grid-World:} 
The performance of the language model in the text-based Grid-World significantly surpasses VLA models in multi-modal settings. This suggests that state descriptions provided purely in text format are less challenging for models than multi-modal inputs, underscoring a notable gap in current VLA models’ ability to effectively interpret visual information. For the language model, we observe a surprising trend in the \textit{Goal} and \textit{Agents} splits: training with more data lower the success rate. This decline suggests that the generalization capacity for certain task categories actually diminishes as training goes on. One possible cause is when exposed to more data, the model relies more heavily on patterns specific to the training examples, limiting its ability to adapt to unseen scenarios. 

\paragraph{Generalization Splits:} 
For VLA models, performance generally drops when models transfer to novel generalization splits, especially in the \textit{Goal} and \textit{Agents} categories. The \textit{Scene} split primarily tests image understanding, while the \textit{Goal} and \textit{Agents} splits emphasize task planning and allocation, critical factors in multi-agent systems. 
This indicates that VLA models still struggle with planning for unseen goals and adapting to variable numbers of agents.

\paragraph{Scaling Law:} 
As training data increases, we observe significant improvements in both subgoal and task success rates across centralized and decentralized settings, underscoring the importance of dataset size for achieving better performance. The improvement is particularly pronounced when the training data increases from $10\%$ to $50\%$ in centralized settings. This suggests that while more data generally leads to better performance, gains diminish beyond a certain point, especially in the decentralized setting.

\paragraph{Centralized vs. Decentralized:} 
\autoref{fig:success-rate} compares centralized and decentralized settings in terms of subgoal and task success rates across all task variants. Centralized tasks exhibit significantly better performance across nearly all variants, highlighting the challenge of effective planning with partial information. This finding also demonstrates that multi-agent systems cannot be simplistically modeled as single agents interacting with environments containing other agents. In decentralized settings, the absence of comprehensive agent modeling is particularly impactful, especially for cooperation-intensive tasks like "Farming" or "Building". 

\paragraph{Redundancy Rate:} 
\autoref{t:redundancy} compares redundancy rates between centralized and decentralized settings. Our results show that decentralized agents exhibit significantly higher redundancy rates than centralized agents, indicating reduced efficiency in task planning and allocation. This inefficiency becomes even more pronounced as the number of agents increases, creating greater challenges for effective task allocation. In decentralized settings, the absence of centralized control complicates the avoidance of redundant work, as each agent must independently infer the intentions of others to prevent duplication. By contrast, a centralized controller can efficiently assign distinct tasks to each agent, minimizing overlap and enhancing overall efficiency. These findings suggest that VLA models lacking explicit mechanisms to understand or infer the actions of other agents, highlighting a critical need for improved inter-agent communication and awareness within decentralized systems.

\begin{table}
    \centering
    \caption{Comparison of TeamCraft-VLA model redundancy rates.}
    \resizebox{\linewidth}{!}{\begin{tabular}{ccccccc}
    \toprule
               & Test & Goal & Scene & Agents & Average \\ 
               \midrule
        TeamCraft-VLA-7B-Cen & 0.01 & 0.02 & 0.01& 0.01 & 0.01 \\
        TeamCraft-VLA-7B-Dec & 0.13 & 0.12 & 0.13& 0.24 & 0.15 \\       
        \bottomrule
    \end{tabular}}
    \label{t:redundancy}
\end{table}

\paragraph{GPT-4o Result:} 
We evaluate GPT-4o on in a one-shot prompt learning setup and it failed on almost all test cases. A detailed analysis reveals that GPT-4o struggles with mapping block coordinates based on visual inputs, demonstrate a lack of 3D spatial reasoning needed for accurate task execution. This shortcoming severely impacts performance, since most of our tasks require precise spatial orientation and alignment. For example, in building task, a brick should be placed at (8,0,8), while the output of model is "placeItem(bot1, 'bricks', (7,0,9))" which leads to wrong execution.

\subsection{Qualitative Analysis}

We performed a qualitative analysis across three generalization splits, examining how models handle novel goals, new scenes, and novel number of agent:

\paragraph{Goals:} 
When faced with novel goals, the models struggle to generalize beyond familiar items and often fail to adapt to specific, unseen objectives. For example, in the "farming" task, if instructed to farm beetroot—a crop not encountered in training—the model might generate a command like "farm\_work(bot1, (9,3,3), 'sow', 'beef')," causing Bot1 to sow "beef", which appears in the training data for "smelting". This behavior reflects the model’s reliance on similar, previously seen items in the training data and reveals its limited ability to infer new tasks based solely on partial similarity.

\balance

\paragraph{Object State Recognition:} 
VLA models show strong generalization to new scenes, performing comparably to the \textit{Test} set. However, errors often arise in recognizing object states. For example, in "farming" tasks, agents may harvest crops before they are fully grown due to challenges in identifying crop states, especially when encountering new scenes. This highlights limitations in precise object state recognition when operating within unseen environments.

\paragraph{Agents:} 
For generalization to four agents, models frequently ignoring the fourth agent and assigning tasks inefficiently only to two or three agents. For example, for the building task, the model predicts the action sequences \{"placeItem(bot1, 'birch\_log', (4,4,7))", "placeItem(bot2, 'sandstone', (4,4,6))", "placeItem(bot3, 'dirt', (3,4,6))"\} with the fourth agent overlooked, reducing productivity and sometimes preventing timely task completion. This limitation becomes especially evident in tasks requiring full coordination, such as "Building." In these tasks, each of the four agents holds unique blocks in their inventory, and all agents must contribute their specific block to a shared platform to complete the structure. The model’s inability to distribute tasks effectively across all agents often leads to incomplete structures or outright task failure. This highlights a significant limitation in precise coordination and workload distribution necessary for successful multi-agent collaboration.

\section{Conclusions}

We have presented \TeamCraft, a benchmark for multi-modal multi-agent collaborative task planning in Minecraft. The benchmark consists of challenging collaborative tasks and evaluation splits designed to systematically test multi-modal agents across novel goal configurations, unseen numbers of agents, and unseen scenes. We conducted extensive experiments and analyses to pinpoint the limitations of the current models and identified promising research directions for collaborative multi-modal agents. 

\subsection{Limitations and Future Work}

\begin{enumerate}
\item Given the limited capacity of existing multi-agent VLA models, \TeamCraft relies on MineFlayer as an oracle controller to execute skills predicted by the models. Enabling VLA models to directly control multiple agents via low-level control \cite{wang2023describe, wang2023jarvis} would be important future research. 
\item We have trained the models using procedurally generated multi-agent demonstration data. Learning from noisy but more diverse real-world demonstrations of human players can potentially further strengthen model generalization \cite{baker2022video, fan2022minedojo}.
\item Currently decentralized \TeamCraft agents rely solely on implicit communication \cite{jain2019body}; i.e., through passively perceiving other agents and the environment, to gather information and to collaborate. Enabling agents to communicate explicitly via natural language \cite{narayan2019collaborative, jayannavar2020learning, mandi2024roco} has great potential in avoiding redundant actions and increasing efficiency. 
\item Multi-player video games have been widely used as testbeds for human-AI collaboration \cite{carroll2020utility, gao2020joint, amresh2023minecraft}. Extending \TeamCraft with human players is a promising research direction.
\end{enumerate}

\clearpage
\input{appendices}

\clearpage
\bibliographystyle{IEEEtranS}
{\small\bibliography{main}}

\end{document}

%% file: appendices.tex
\section*{\Large Appendices}
\appendix

\begin{table*}
    \centering
    \caption{Action space of \TeamCraft.}
    \resizebox{\textwidth}{!}{
        \begin{tabular}{lll}
            \toprule
            \textbf{Type} & \textbf{Arguments} & \textbf{Description} \\
            \midrule
            placeItem & BotID, ItemType, Location & BotID places an item of ItemType at the specified 3D Location. \\
            \midrule
            mineBlock & BotID, Location & BotID mines a block at the specified 3D Location. \\
            \midrule
            farmWork & BotID, Location, Action, ItemType & BotID performs an Action (sow or harvest) on ItemType at the specified 3D Location. \\
            \midrule
            obtainBlock & BotID, Location & BotID obtains a block from the specified 3D Location. \\
            \midrule
            putFuelFurnace & BotID, ItemType, Location & BotID places an ItemType as fuel into a furnace at the specified 3D Location. \\
            \midrule
            putItemFurnace & BotID, ItemType, Location & BotID inserts an ItemType into a furnace at the specified 3D Location. \\
            \midrule 
            takeOutFurnace & BotID, ItemType, Location & BotID removes an ItemType from a furnace at the specified 3D Location. \\
            \midrule 
            killMob & BotID, Location & BotID engages and eliminates a mob at the specified 3D Location. \\
            \bottomrule
        \end{tabular}
    }
    \label{tab:atomic_actions}
\end{table*}

\section{High Level Skills}
\label{appendix:high_level_skills}

The action space of agents mainly involves high-level self-explanatory skills such as \textit{obtainBlock} and \textit{farmWork}. We provided 8 such skills. Most skills take three input parameters, including 1) agent name such as \textit{bot1}, as the action executing entity, 2) item name such as \textit{dirt}, which strongly associated with task goal or agent's inventory, 3) a vector indicating the position of the target on the test field.

For example, \texttt{obtainBlock(bot1, new Vec3(1, 0, 1))} takes the agent name \texttt{bot1} and a 3D vector \texttt{(1, 0, 1)} as its arguments. It directs \texttt{bot1} to perform multiple actions in Minecraft via APIs provided by Mineflayer. First, it controls \texttt{bot1} to \texttt{goto} a diggable position for block \texttt{(1, 0, 1)}, then has \texttt{bot1}'s vision ray cast to the block at \texttt{(1, 0, 1)} using the \texttt{lookAt} action. Next, it commands \texttt{bot1} to \texttt{equip} a proper tool that can dig the block at \texttt{(1, 0, 1)} most efficiently, and then instructs \texttt{bot1} to dig the target block. Once the target block has been mined, \texttt{bot1} will \texttt{goto} the position where the block item dropped and collect it.

Similarly, \texttt{farmWork(bot2, "sow", "potato", new Vec3(2, 0, 4))} takes the agent name \texttt{bot2}, action type \texttt{"sow"} (as opposed to \texttt{"harvest"}), crop seed item \texttt{"potato"}, and a 3D vector \texttt{(2, 0, 4)} as its arguments. It directs \texttt{bot2} to \texttt{goto} a placeable position for farmland at \texttt{(2, 0, 4)}, then \texttt{check} if the seed is a valid item—that is, a crop seed available within \texttt{bot2}'s inventory. It then \texttt{checks} if the farmland at \texttt{(2, 0, 4)} is plantable. Finally, it instructs \texttt{bot2} to \texttt{lookAt} the farmland and \texttt{sow} it with the seed \texttt{"potato"}.

\autoref{tab:atomic_actions} documents all the skills, which are implemented in JavaScript code with Mineflayer APIs.

\section{Low Level Atomic Actions}

High level skills are processed through multiple stages before reaching the final execution APIs. At each time step, \TeamCraft accepts a list of skills as input, with a maximum length equal to the number of agents involved in the current task and a minimum length of zero. Each agent can perform at most one skill per time step. The updated list of skills is then passed into the JavaScript environment along with the predefined atomic actions. Each atomic action is processed simultaneously, meaning that agents' actions are executed concurrently rather than sequentially. This avoid the dependency issue that might occur in sequential execution. For example, if one agent's action is executed ahead of another's, the first agent may block the location where the next agent intends to place a block. The agent whose atomic action is executed first will have a higher chance of success, potentially altering the dynamics of the multi-agent setting. Executing actions concurrently ensures fairness among agents and maintains the equivalence of the multi-agent environment. 

\section{Visual Diversity}
\label{appendix:visual_diversity}

\TeamCraft uses a set of visual variate to provide a visual rich environment. Each task is constructed from a random number of agents, in a randomly selected scene, achieving different goal on playground built by different base block.

\subsection{Shared Elements}

\begin{figure*}
    \centering
        \includegraphics[width=1.0\linewidth]{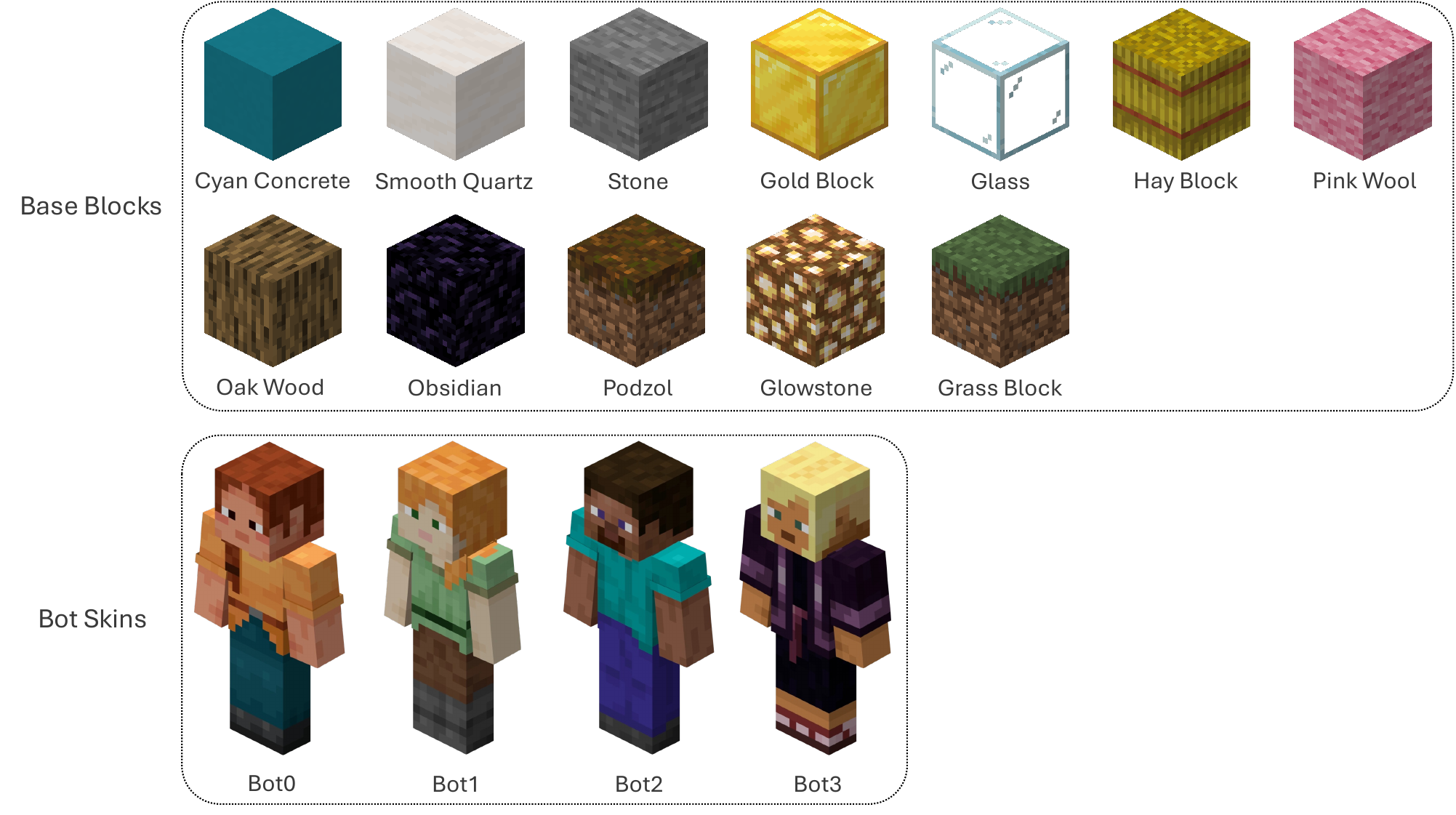}
        \caption{A close-up view of the shared visual diversity in every tasks.}
        \label{fig:diversity_1}
\end{figure*}

Each task begins with a basic setting involving multiple agents on a playground. Each agent has a unique skin, as illustrated in \autoref{fig:diversity_1}, and is rendered as a two-block-high character. The playground combines a base platform spawned within a Minecraft biome. The base block is also randomly selected from a pool,shown in \autoref{fig:diversity_1}. Each biome offers variations in special surrounding blocks, designs, and environments.

\begin{figure*}
    \centering
        \includegraphics[width=1.0\linewidth]{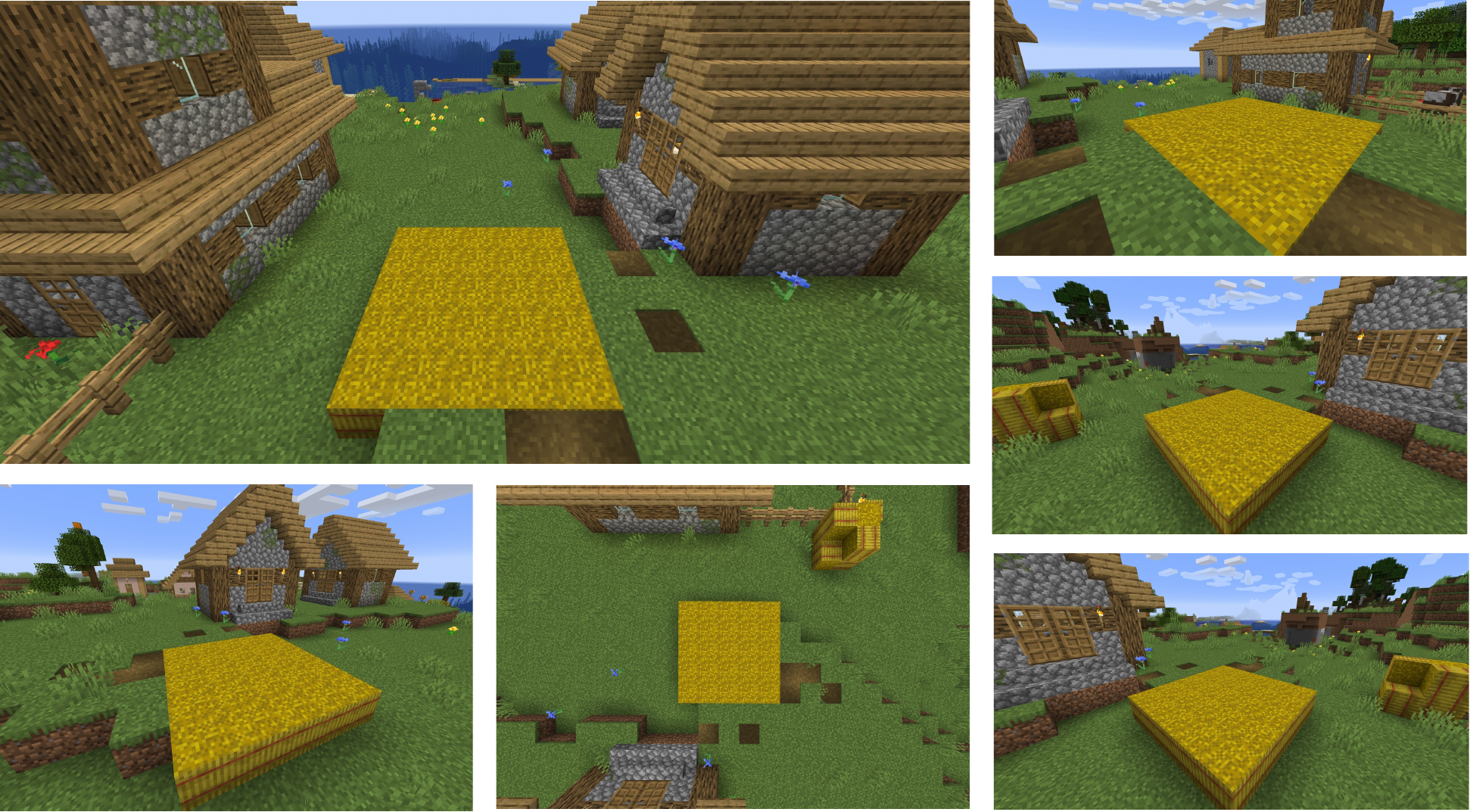}
        \caption{An example scene in the seaside village biome.}
        \label{fig:diversity_scenes_1}
\end{figure*}

\begin{figure*}
    \centering
        \includegraphics[width=1.0\linewidth]{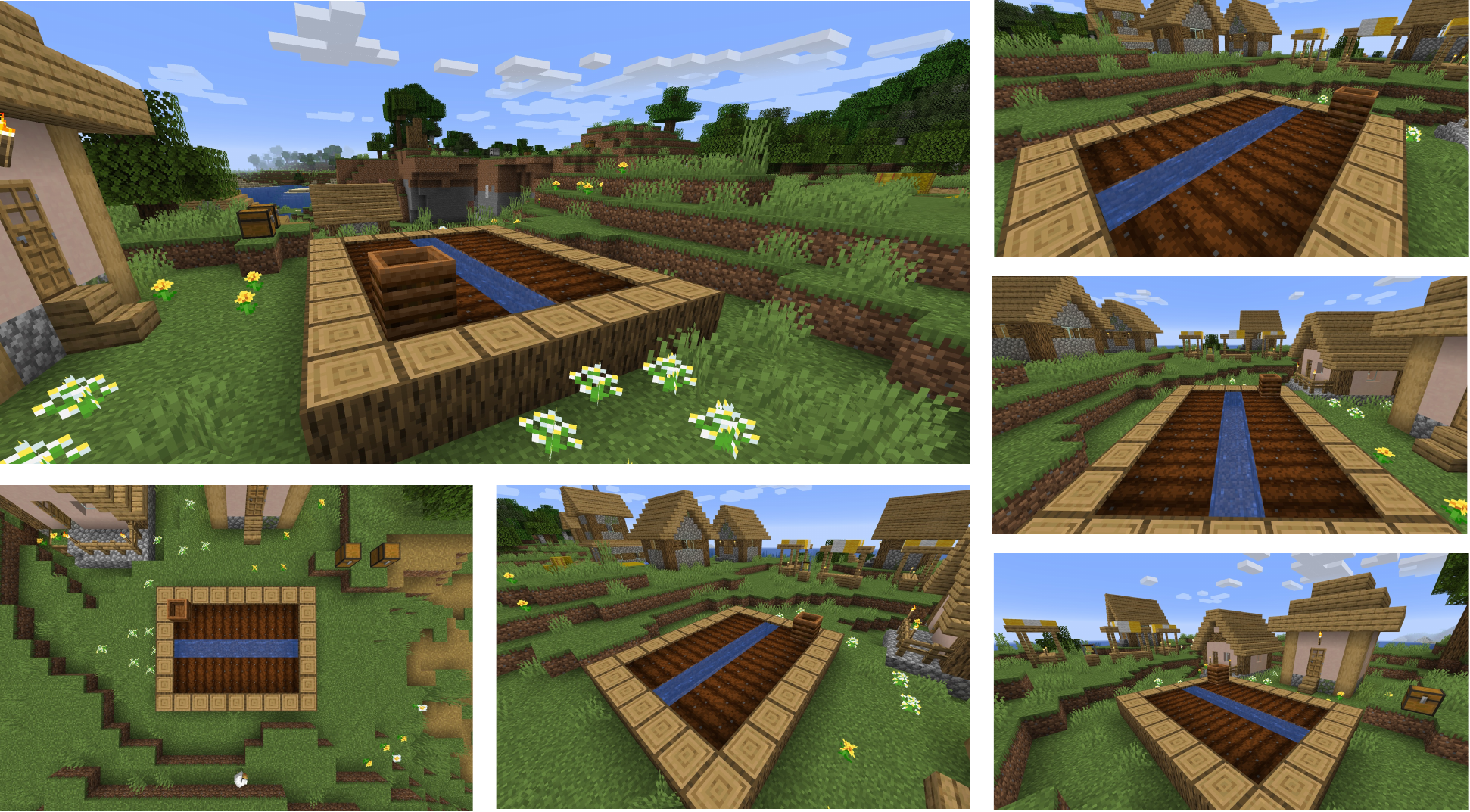}
        \caption{An example scene in the grass village biome.}
        \label{fig:diversity_scenes_2}
\end{figure*}

\begin{figure*}
    \centering
        \includegraphics[width=1.0\linewidth]{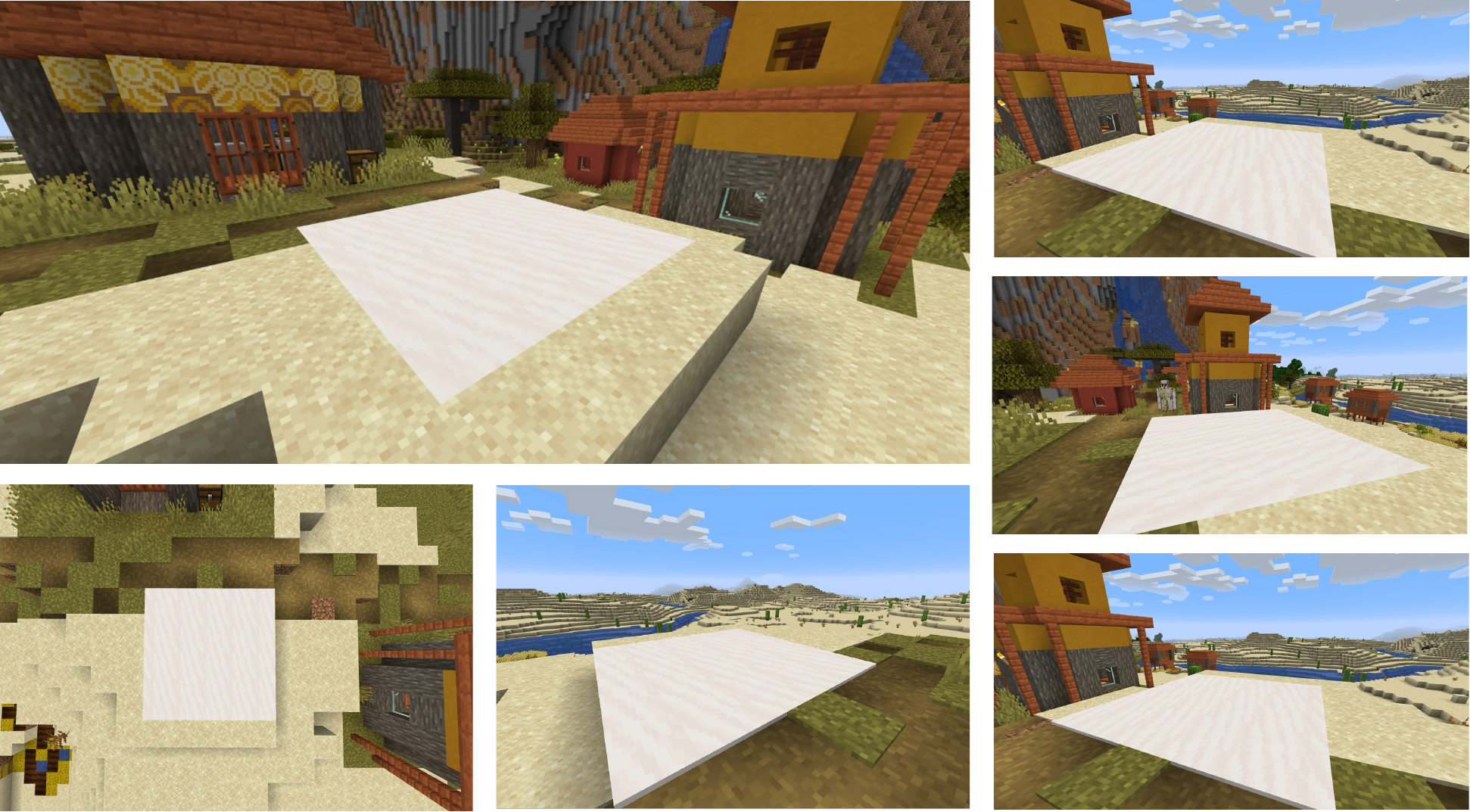}
        \caption{An example scene in the dessert village biome.}
        \label{fig:diversity_scenes_3}
\end{figure*}

\begin{figure*}
    \centering
        \includegraphics[width=1.0\linewidth]{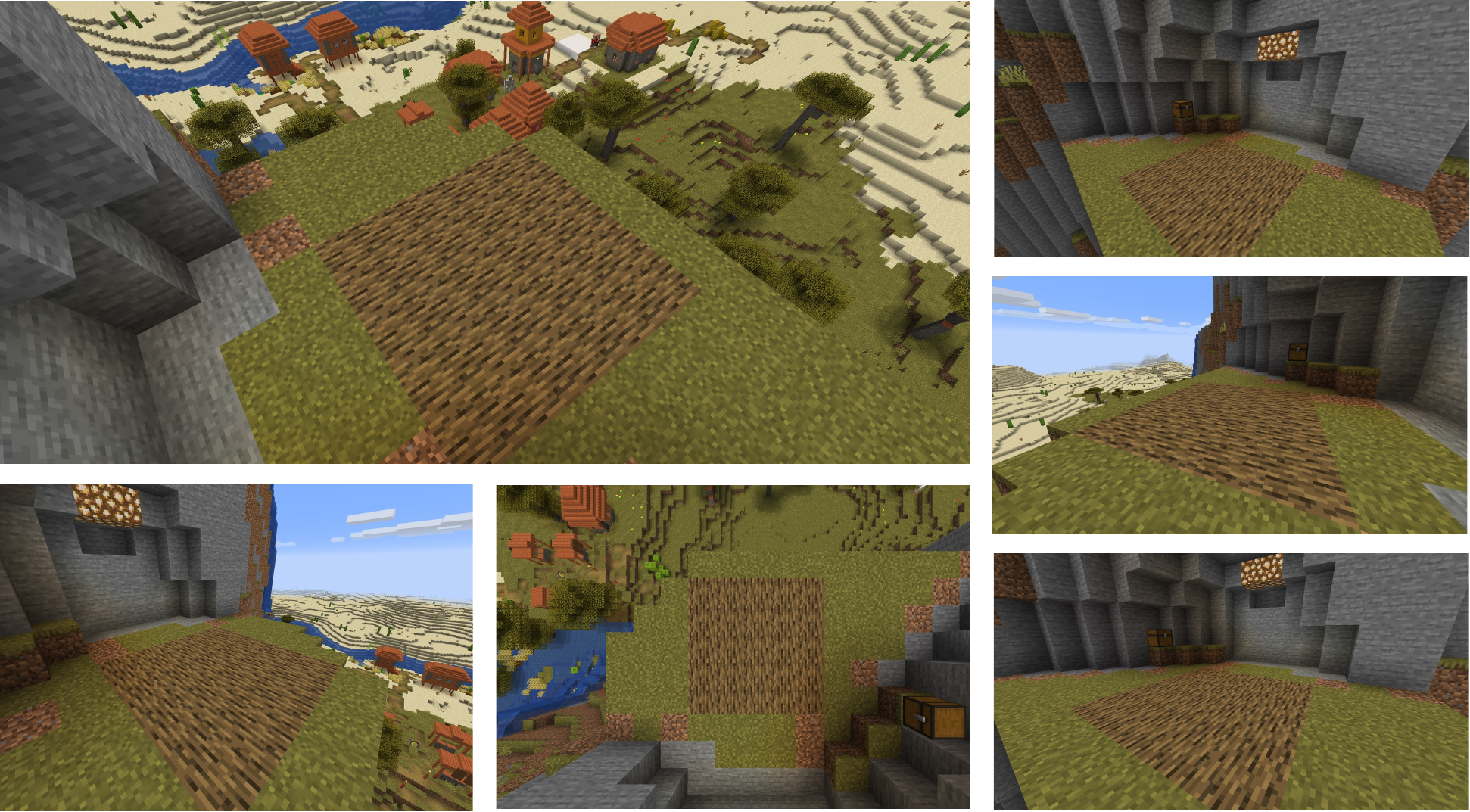}
        \caption{An example scene in the half mountain biome.}
        \label{fig:diversity_scenes_4}
\end{figure*}

\begin{figure*}
    \centering
        \includegraphics[width=1.0\linewidth]{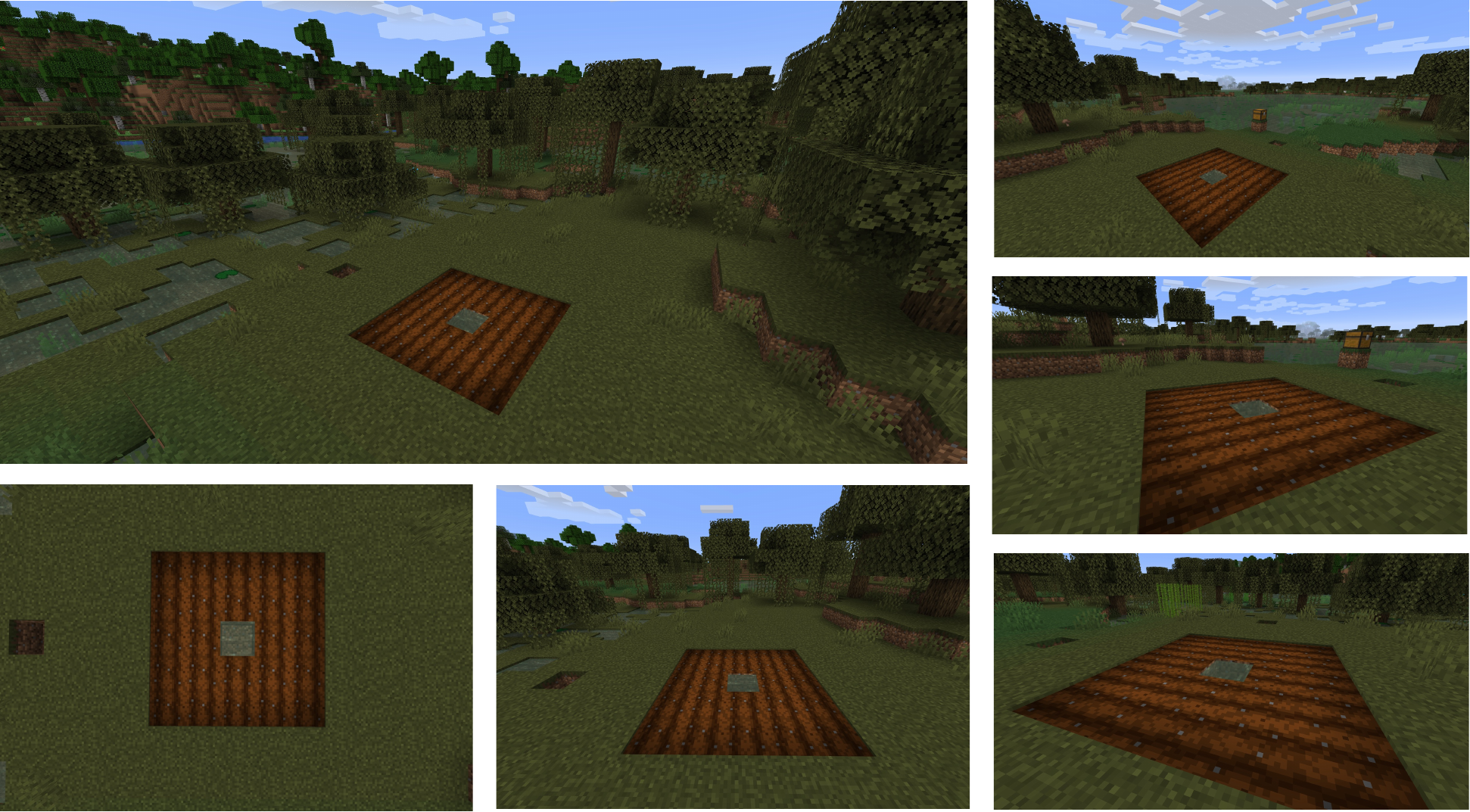}
        \caption{An example scene in the swamp biome.}
        \label{fig:diversity_scenes_5}
\end{figure*}

\begin{figure*}
    \centering
        \includegraphics[width=1.0\linewidth]{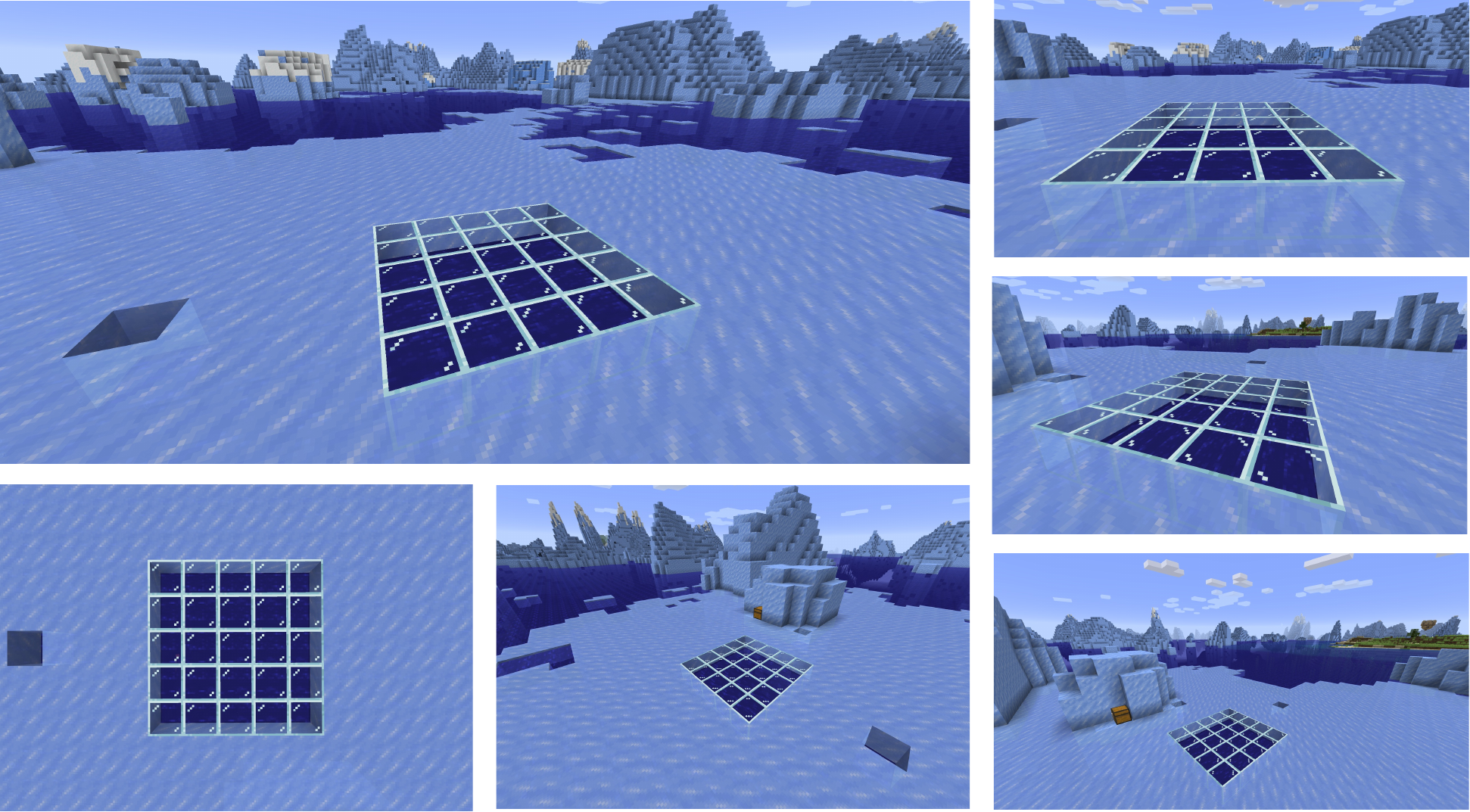}
        \caption{An example scene in the iceberg biome.}
        \label{fig:diversity_scenes_6}
\end{figure*}

\begin{figure*}
    \centering
        \includegraphics[width=1.0\linewidth]{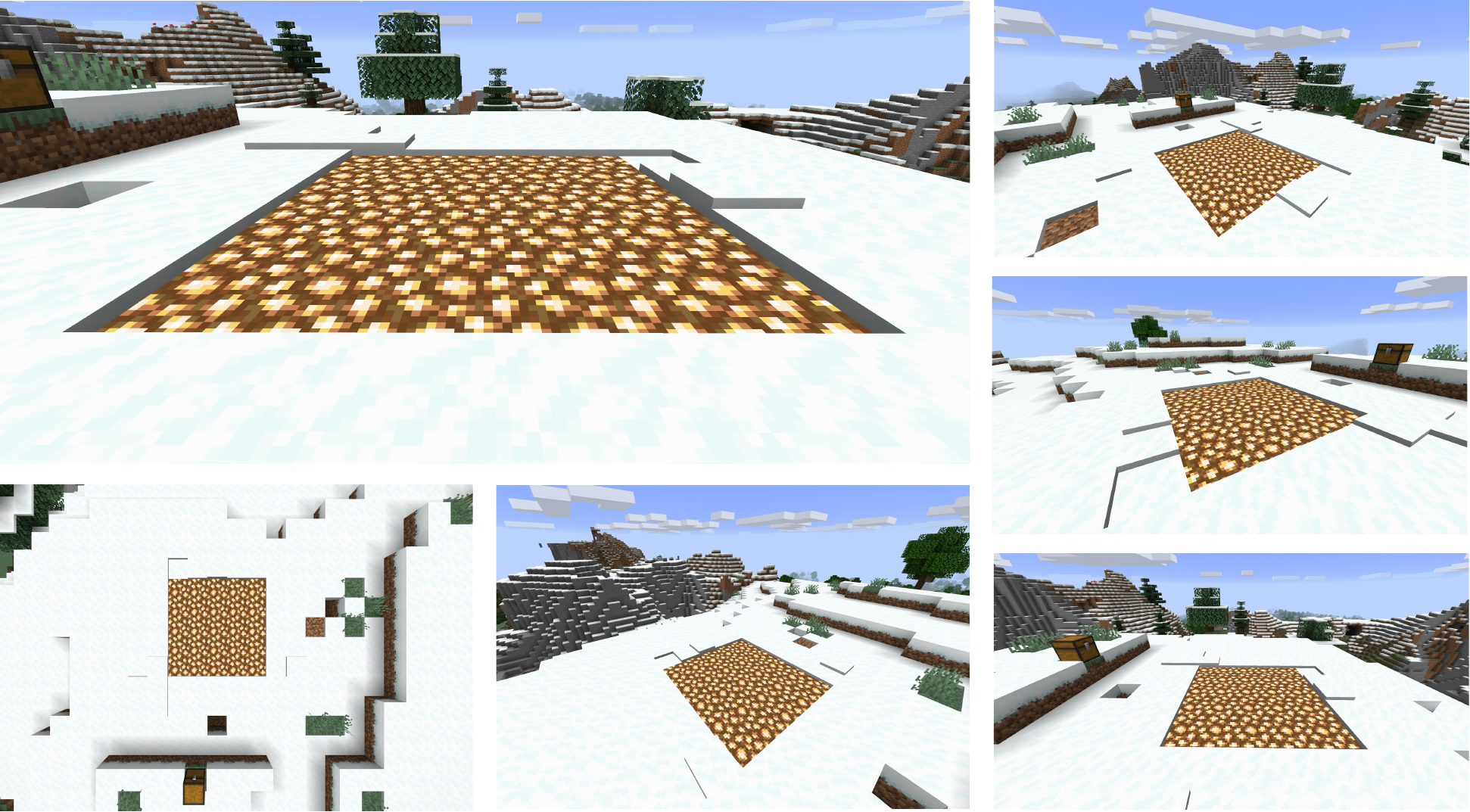}
        \caption{An example scene in the snow mountain biome.}
        \label{fig:diversity_scenes_7}
\end{figure*}

For example, the \texttt{seaside village} biome is a village near the sea with houses made of oak wood and cobblestone, decorated with flowers and cow sheds, as shown in \autoref{fig:diversity_scenes_1}. It also features a nearby farm surrounded by oak logs (\autoref{fig:diversity_scenes_2}). Another variation of village is the \texttt{desert village} biome, built from acacia planks, acacia logs, and sandstone, blending seamlessly with the desert's arid terrain, shown  in \autoref{fig:diversity_scenes_3}. \autoref{fig:diversity_scenes_4} illustrates a biome that is located on half of the mountains, where a small flat land protruding from a cliff. Additional examples of biomes used are shown in \autoref{fig:diversity_scenes_5}, \autoref{fig:diversity_scenes_6}, and \autoref{fig:diversity_scenes_7}.

\subsection{Task Specific Diversity}

\begin{figure*}
    \centering
        \includegraphics[width=1.0\linewidth]{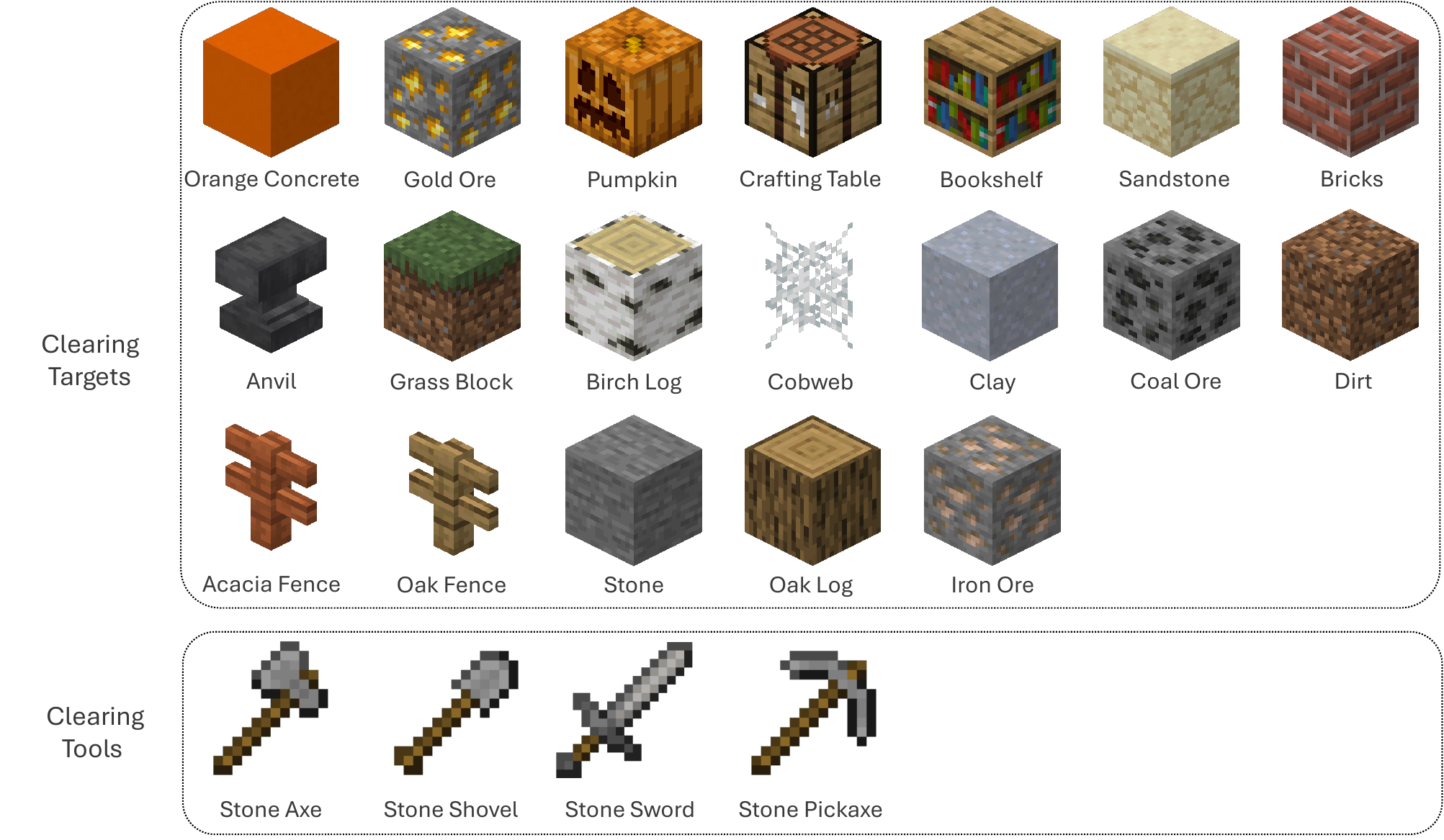}
        \caption{A close-up view of the visual diversity in clearing tasks.}
        \label{fig:diversity_2}
\end{figure*}

\begin{figure*}
    \centering
        \includegraphics[width=1.0\linewidth]{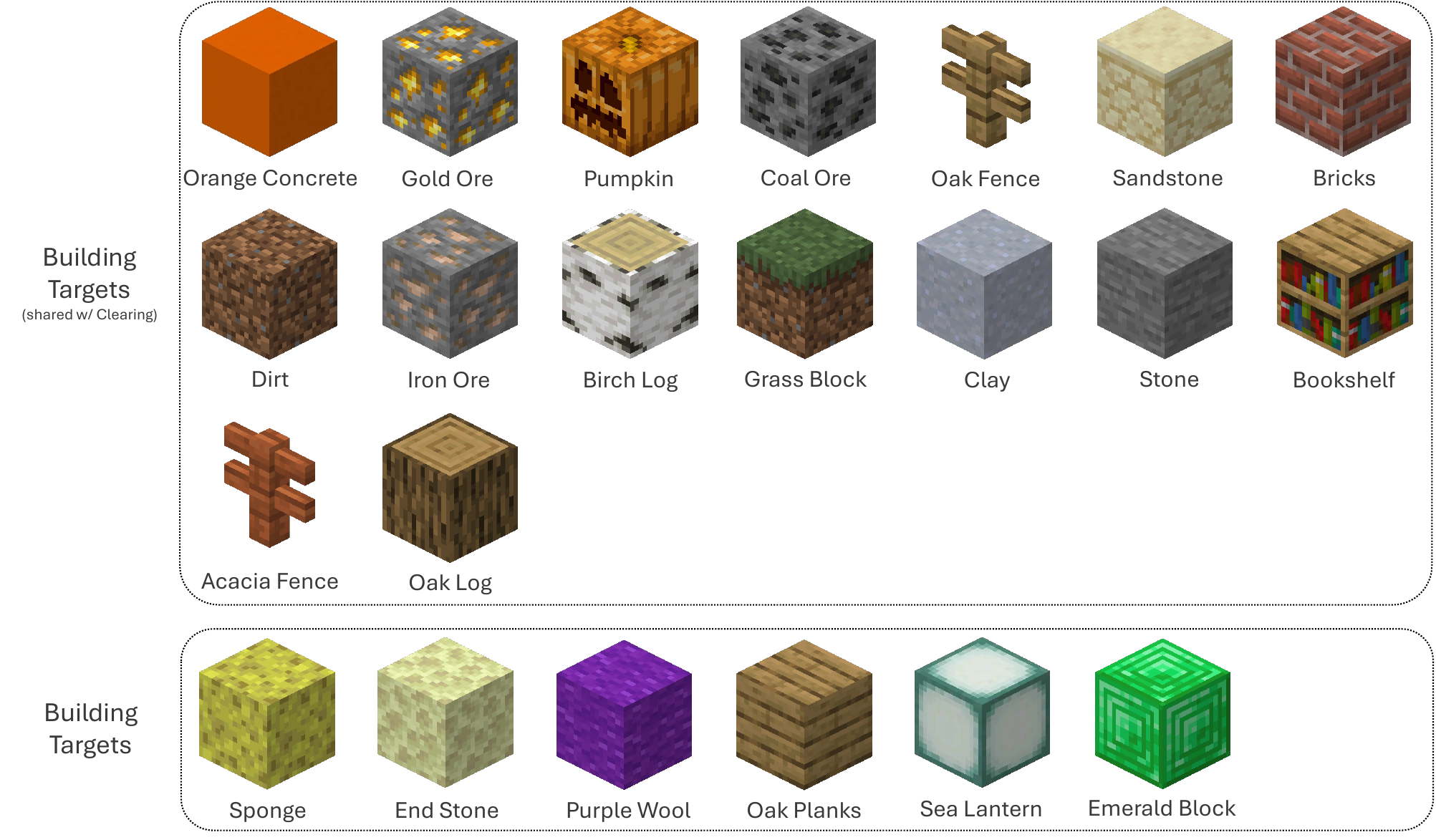}
        \caption{A close-up view of the visual diversity in building tasks.}
        \label{fig:diversity_3}
\end{figure*}

\begin{figure*}
\vspace{-5mm}
    \centering
        \includegraphics[width=0.97\linewidth]{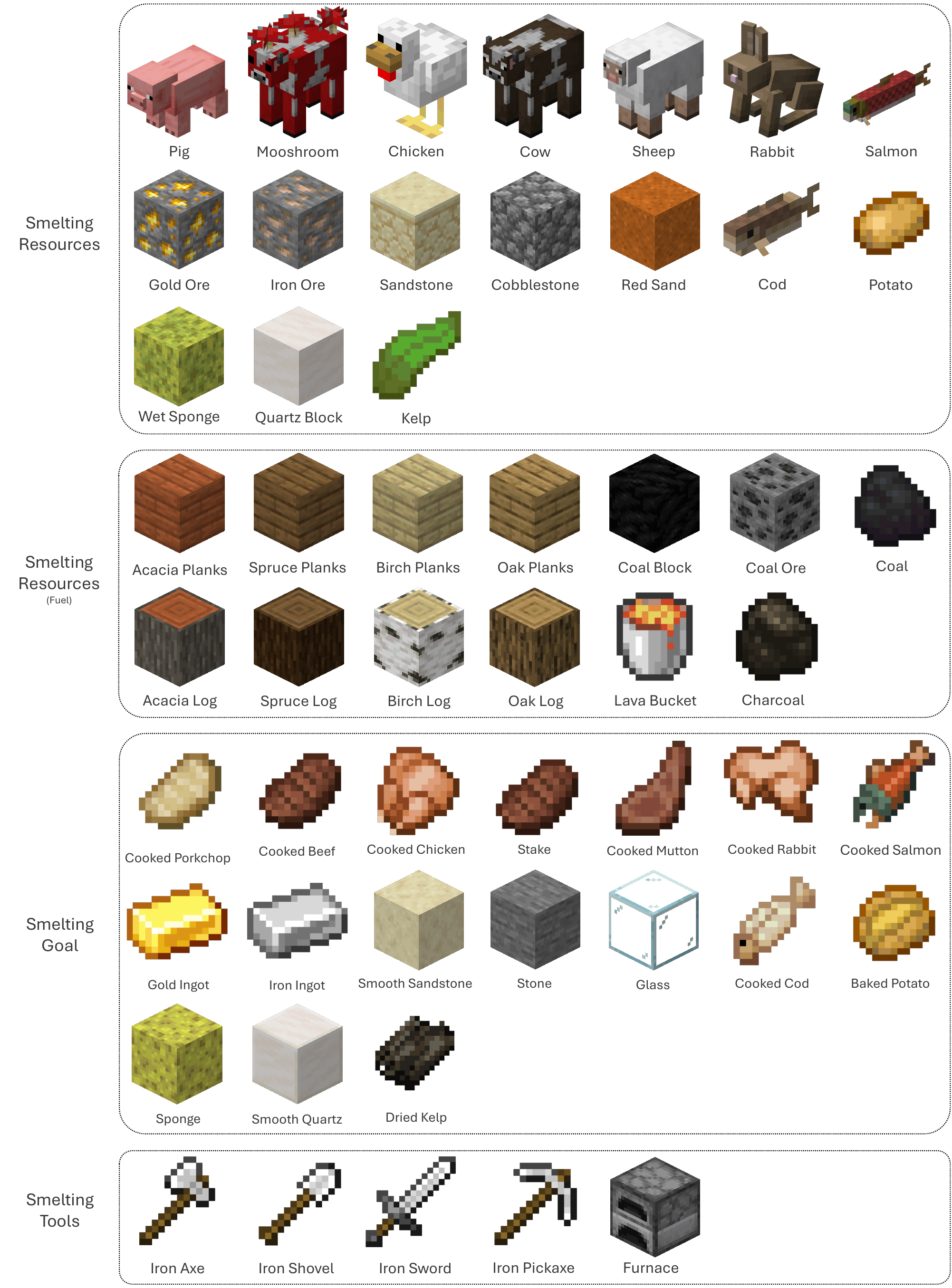}
        \caption{A close-up view of the visual diversity in smelting tasks.}
        \label{fig:diversity_5}
\end{figure*}

\noindent \textbf{Clearing} task uses a random set of blocks as its targets, illustrated in \autoref{fig:diversity_2}.
\noindent \textbf{Building} task also uses a random set of blocks as its target, with some blocks shared with clearing task, as illustrated in \autoref{fig:diversity_3}.
\noindent Unlike other tasks, the \textbf{Farming} task does not use a regular base. The playground is constructed from a combination of farmland for planting crops, water blocks, and randomly selected unfarmable blockers from the base that replace some of the farmland. An example is shown in \autoref{fig:expert_farm_1}. Each corps used in farming task has its own grown stage with different appearances, shown in \autoref{fig:diversity_4}.
\noindent \textbf{Smelting} task requires a wide varieties of resources to achieve its goal. Resources could be either entity, block, or item, shown in \autoref{fig:diversity_5}. 

Detailed statistics of each task are presented in \autoref{tab:statistics_building}, \autoref{tab:statistics_clearing}, \autoref{tab:statistics_farming}, and \autoref{tab:statistics_smelting} of \autoref{appendix:stat_table}. 

\section{Planner for Expert Demonstration}
\label{appendix:planner_details}

\TeamCraft employed a planner to assign actions to each agent at every time step, utilizing perfect knowledge of the task including goal object positions, agents' inventories, and each agent's efficiency in performing actions. The planner optimizes actions using a cost function designed to minimize the total time to complete the task, reduce idle times for agents, minimize action dependencies to prevent agents from waiting on others, maximize parallelism of actions, assign tasks to the most efficient agents, and eliminate redundant or unnecessary actions. The cost function considers the following components:

\paragraph{Minimize Total Task Completion Time $T$:} Denoted by $\min T$, our primary objective is to reduce the overall time required to complete the task, measured in time steps until the last agent finishes their final action.

\paragraph{Minimize Idle Actions for Each Agent $E$:} Denoted by $\min \sum_{i=1}^N E_i$, we minimize the total empty actions, the sum of empty action $E_i$ preformed by agent $i$.

\paragraph{Minimize Action Dependencies Across Agents $D$:} Denoted by $\min D$, we minimize dependencies cause agents to wait for others to complete certain actions. 

\paragraph{Minimize Redundant or Useless Actions $U$:} Denoted by $\min U$, we minimize the total number of redundant or unnecessary actions performed by all agents. 

\paragraph{Maximize Action Efficiency:} We assign actions to agents with higher capabilities to minimize the overall cost, $\min \sum_{i=1}^N \sum_{j \in A_i} c_{ij}$, where $c_{ij}$ be the cost (inverse of efficiency) for agent $i$ to perform action $j$. 

\bigskip
We assign each component a weight as follows:
\begin{equation}
    C = w_1 T + w_2 \sum_{i=1}^N E_i + w_3 D + w_4 \sum_{i=1}^N \sum_{j \in A_i} c_{ij} + w_5 U
\end{equation}
where $w_1$, $w_2$, $w_3$, $w_4$, and $w_5$ are weighting coefficients that are adjusted for each task as described below:

\paragraph{Building:} In the building task, where dependencies are moderate and parallelization is preferred, we place greater emphasis on minimizing idle actions by setting $w_2 = 1.4$ and assign a weight of 0.9 to the other components. This encourages agents to remain active and reduces idle time, enhancing overall efficiency.

\paragraph{Clearing:} In the clearing task, using the correct tools can significantly speed up block removal (up to a threefold increase). Therefore, we assign a higher weight of $w_4 = 1.8$ to maximize action efficiency by assigning tasks to the most capable agents. The other weights are set to 0.8 to maintain overall performance while focusing on efficient tool usage.

\paragraph{Farming:} Farming task is not heavily constrained by action dependencies, we assign equal weights of 1 to all components, ensuring a balanced consideration of time minimization, idle actions, action dependencies, action efficiency, and redundancy elimination.

\paragraph{Smelting:} In the smelting task, which involves comparatively long and highly dependent action sequences, we prioritize minimizing action dependencies by setting $w_3 = 1.8$. The other weights are assigned a value of 0.8 to support this focus, facilitating smoother coordination among agents and reducing waiting times.

\begin{figure*}[!ht]
    \centering
\begin{tcolorbox}[colback=gray!5!white,colframe=gray!75!black]
\textbf{System Prompt}\newline
Three bots need to build a building on the platform. Target building is: {Put sea\_lantern on [0 ,1 ,0]. Put oak\_fence on [-1 ,1 ,0]. Put sponge on [0 ,1 ,-1]. Put emerald\_block on [-1 ,1 ,-1]. Put dirt on [0 ,0 ,0]. Put bricks on [-1 ,0 ,0]. Put emerald\_block on [0 ,0 ,-1]. Put clay on [-1 ,0 ,-1]. }.  
\end{tcolorbox}
\begin{tcolorbox}[colback=gray!5!white,colframe=gray!75!black]
\textbf{User Prompt}\newline
bot1 has 4 dirt. bot1 has 3 clay. bot1 has 7 emerald\_block. bot1 has 1 oak\_fence. bot1 has 3 sponge. bot1 has 1 bricks. bot1 has 3 sea\_lantern. bot2 has 4 bricks. bot2 has 2 sponge. bot2 has 6 sea\_lantern. bot2 has 2 oak\_fence. bot2 has 4 emerald\_block. bot2 has 1 dirt. bot2 has 3 clay. bot3 has 6 emerald\_block. bot3 has 4 oak\_fence. bot3 has 2 dirt. bot3 has 2 sponge. bot3 has 3 clay. bot3 has 2 sea\_lantern. bricks is on [-1 ,0 ,0]. dirt is on [0 ,0 ,0]. Write the actions for bot1, bot2 and bot3 based on this given observation.
\end{tcolorbox}
\caption{Prompt example for Building task under the grid-world setting.}
\label{fig:build-grid}
\end{figure*}

\begin{figure*}[!ht]
    \centering
\begin{tcolorbox}[colback=gray!5!white,colframe=gray!75!black]
\textbf{System Prompt}\newline
Three bots need to break everything on the platform. clay is on [-2 ,0 ,-2]. birch\_log is on [-2 ,0 ,0]. dirt is on [-1 ,0 ,-2]. crafting\_table is on [-1 ,0 ,1]. anvil is on [-1 ,1 ,1]. anvil is on [0 ,0 ,-2]. iron\_ore is on [0 ,0 ,1]. cobweb is on [1 ,0 ,1].  
\end{tcolorbox}
\begin{tcolorbox}[colback=gray!5!white,colframe=gray!75!black]
\textbf{User Prompt}\newline
bot1 has 1 stone\_pickaxe. bot1 has 1 anvil. bot2 has 1 stone\_axe. bot2 has 1 crafting\_table. bot3 has 1 stone\_pickaxe. bot3 has 1 dirt. clay is on [-2 ,0 ,-2]. birch\_log is on [-2 ,0 ,0]. iron\_ore is on [0 ,0 ,1]. cobweb is on [1 ,0 ,1]. Write the actions for bot1, bot2 and bot3 based on this given observation.
\end{tcolorbox}
\caption{Prompt example for Clearing task under the grid-world setting.}
\label{fig:break-grid}
\end{figure*}

\subsection{Example Expert Demonstrations}

\begin{figure*}
    \centering
        \includegraphics[width=1.0\linewidth]{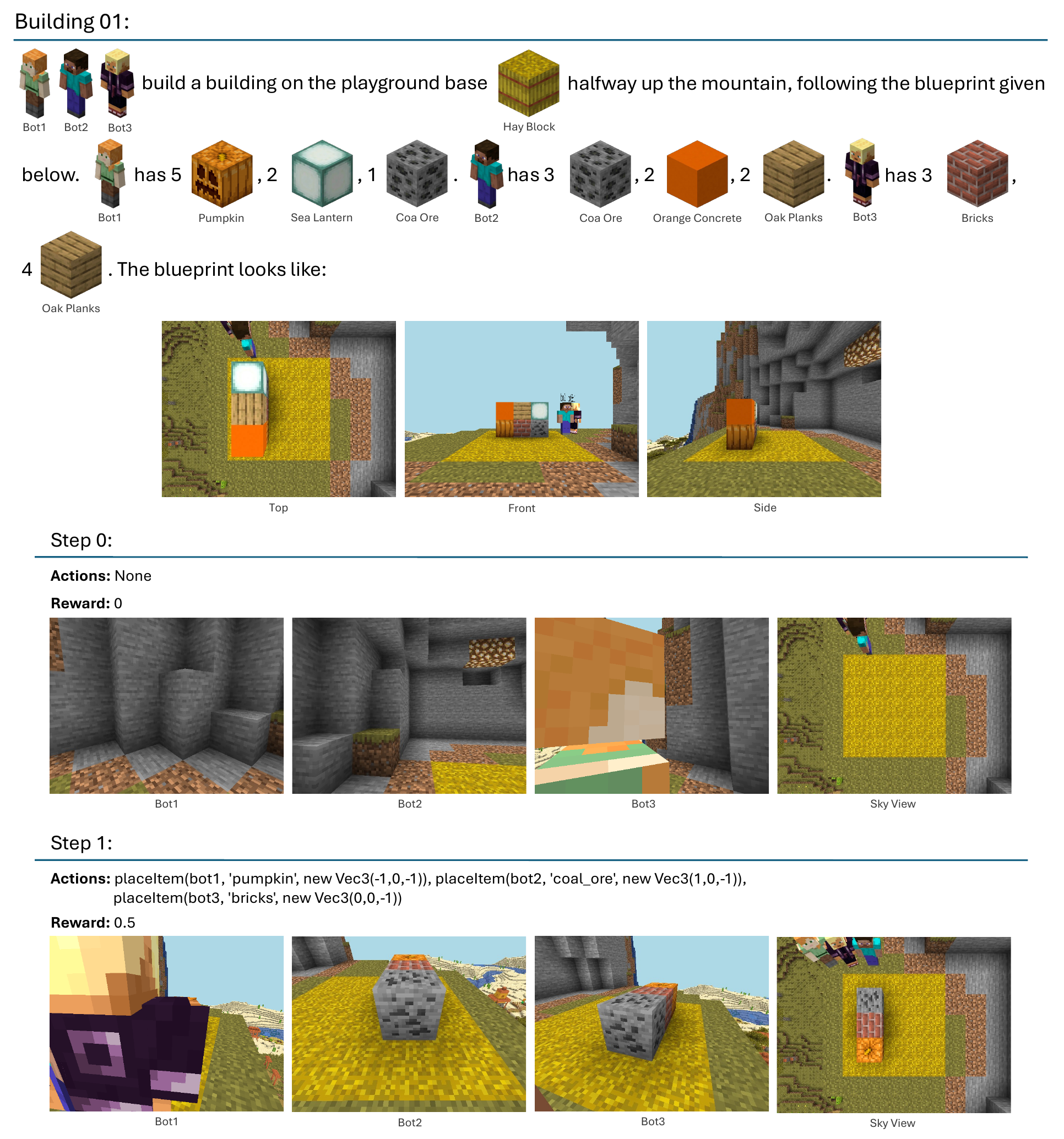}
        \caption{An example demonstration in the building task, Part~I.}
        \label{fig:expert_build_1}
\end{figure*}

\begin{figure*}
    \centering
        \includegraphics[width=1.0\linewidth]{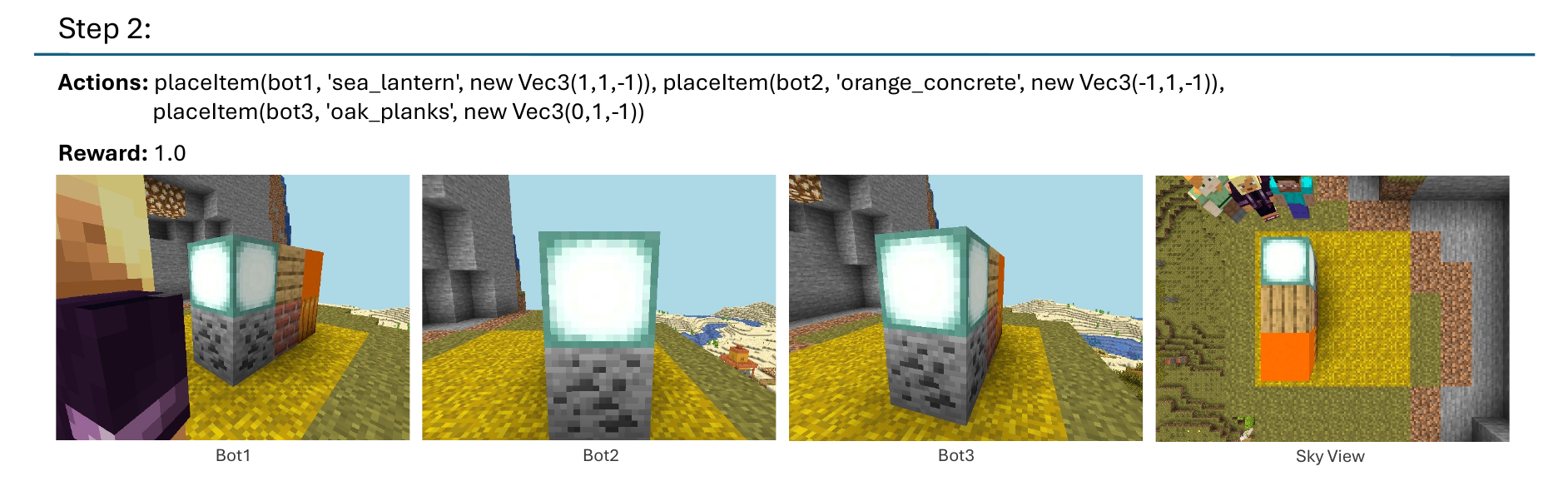}
        \caption{An example demonstration in the building task, Part~II.}
        \label{fig:expert_build_2}
\end{figure*}

\autoref{fig:expert_build_1} and \autoref{fig:expert_build_2} show a classic example of the building task, which involves three agents building a $2\times3$ building on the mountain half. Each of the agents has some of the needed blocks in their inventory to build the building. For every time step after Step~0, each of the three agents build one block from the bottom level to the second level.

\begin{figure*}
    \centering
        \includegraphics[width=1.0\linewidth]{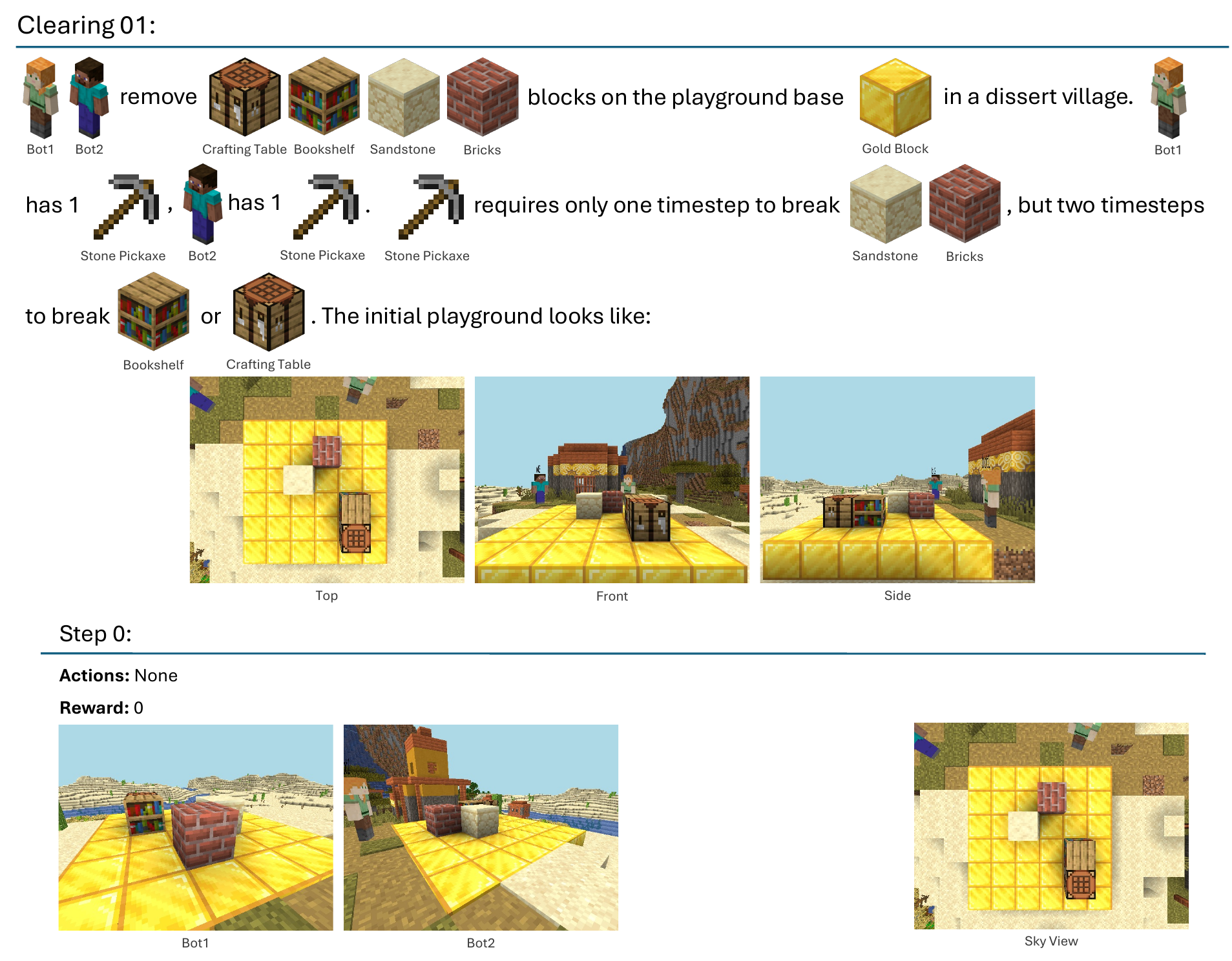}
        \caption{An example demonstration in the clearing task, Part~I.}
        \label{fig:expert_break_1}
\end{figure*}

\begin{figure*}
    \centering
        \includegraphics[width=0.972\linewidth]{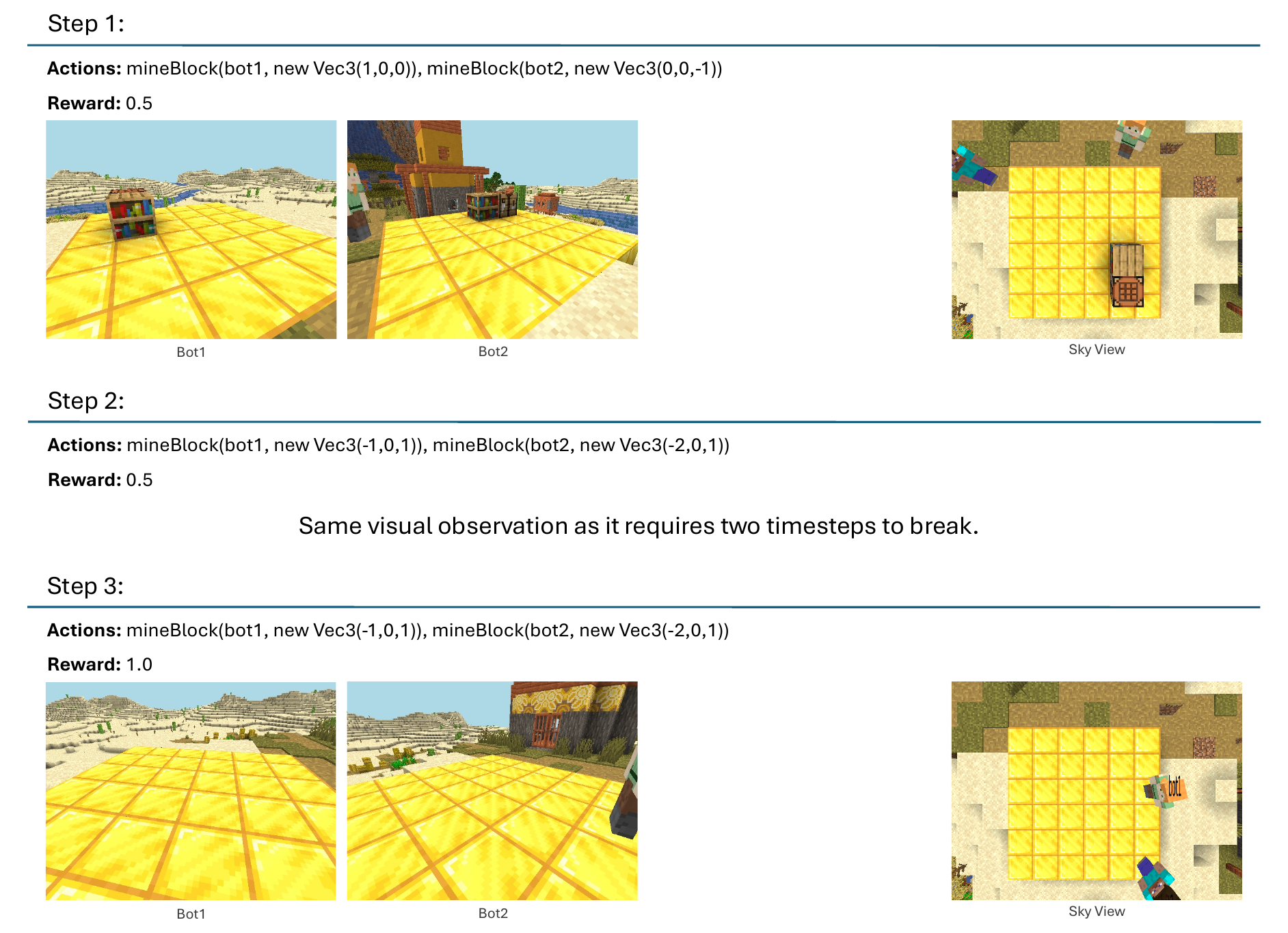}
        \caption{An example demonstration in the clearing task, Part~II.}
        \label{fig:expert_break_2}
\end{figure*}

\autoref{fig:expert_break_1} shows an example of the clearing task. Two agents are assigned to clean the blocks on a $6\times6$ platform. Each of them has a stone pickaxe in their inventory, which is the efficient tool for breaking "stone-like" blocks. In this case, they are able to break brick and sandstone in just one time step with pickaxe but require two time steps to break "wood-made" blocks such as bookshelf and crafting table. This resulted in time Step~2 and Step~3 having exactly the same visual observation, shown in \autoref{fig:expert_break_2}.

\begin{figure*}
    \centering
        \includegraphics[width=0.972\linewidth]{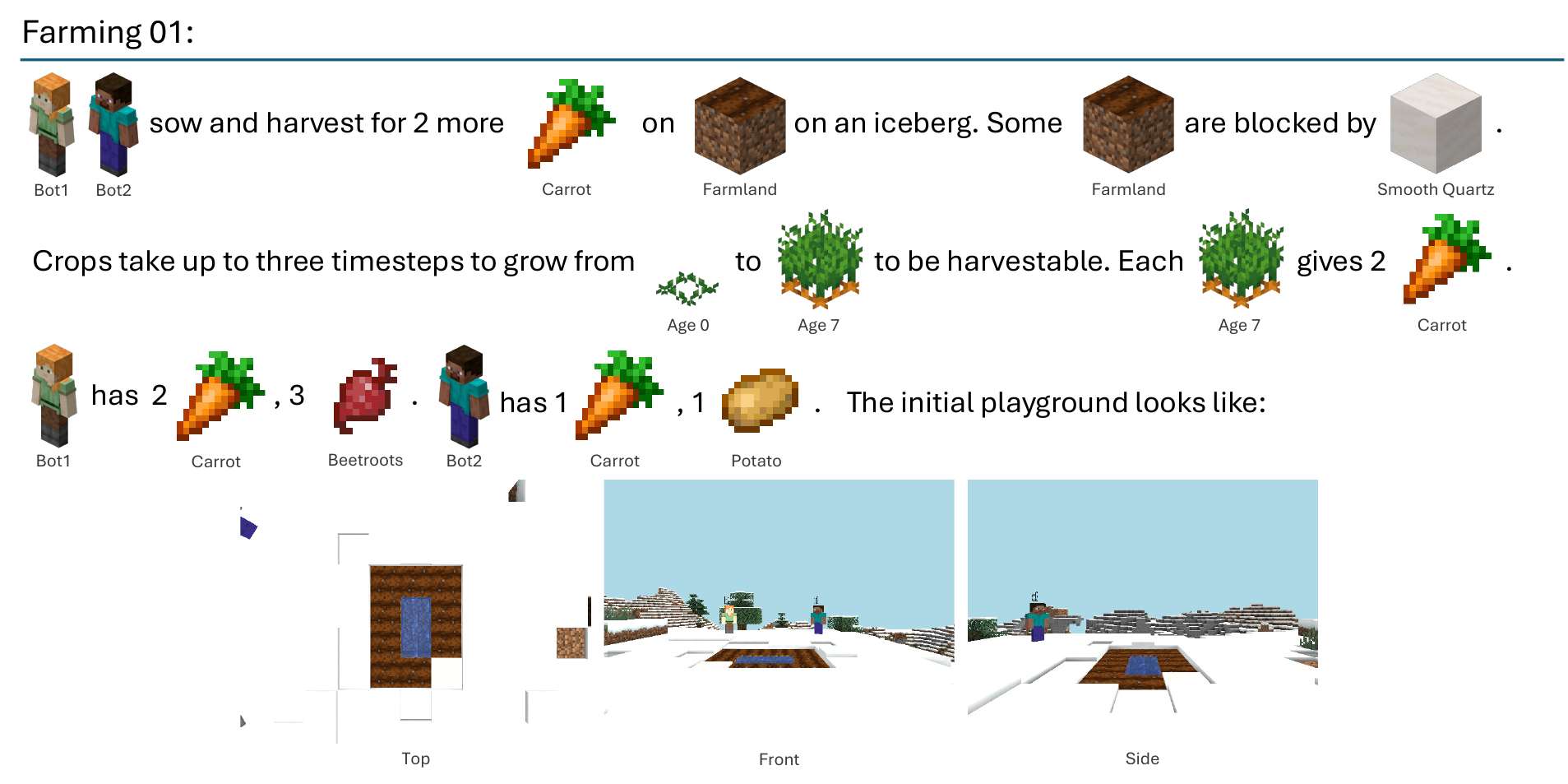}
        \caption{An example demonstration in the farming task, Part~I.}
        \label{fig:expert_farm_1}
\end{figure*}

\begin{figure*}
    \centering
        \includegraphics[width=1.0\linewidth]{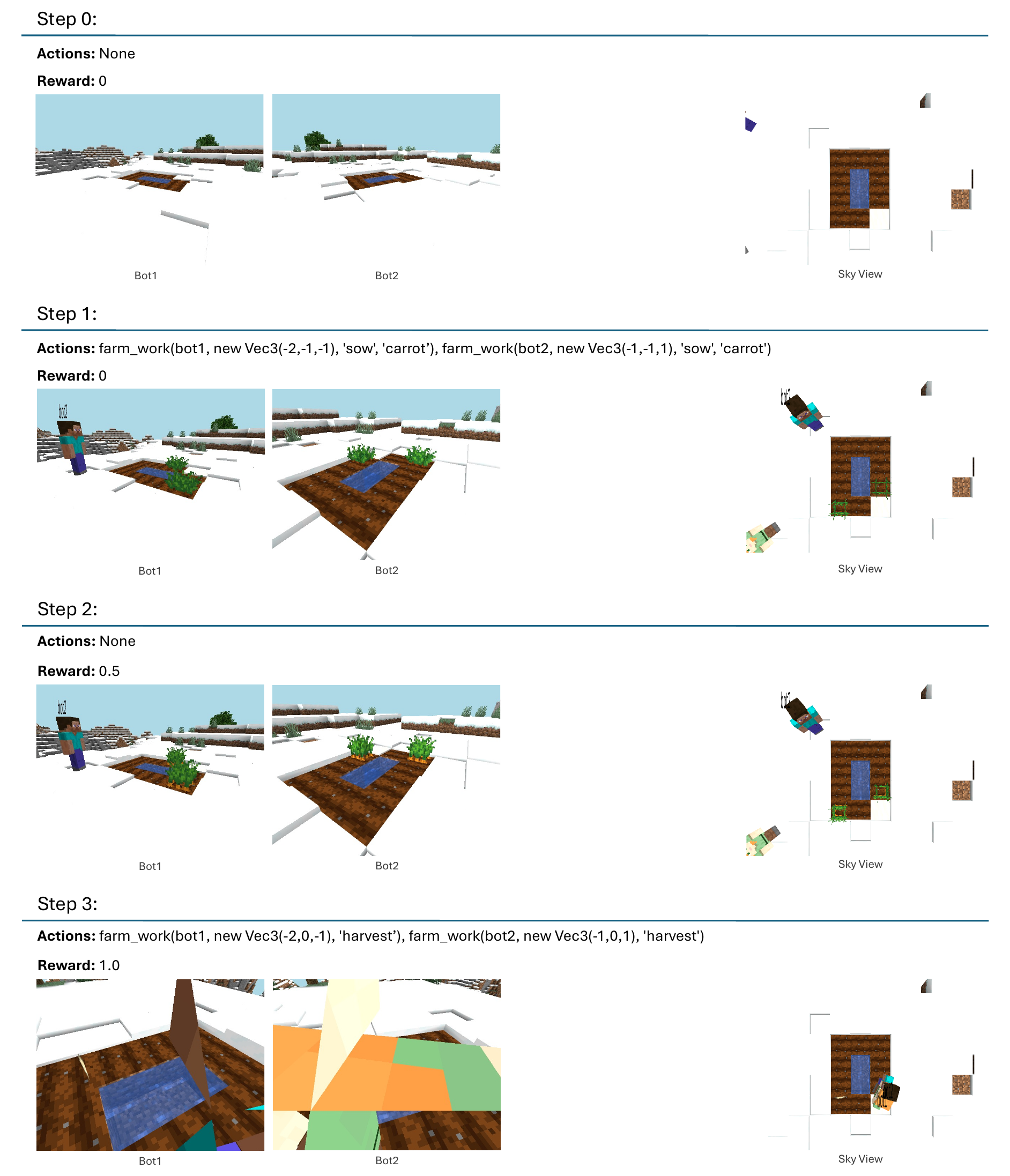}
        \caption{An example demonstration in the farming task, Part~II.}
        \label{fig:expert_farm_2}
\end{figure*}

\autoref{fig:expert_farm_1} and \autoref{fig:expert_farm_2} shows an example of two agents farming on a snow mountain for two extra carrots. In Step~1, agent1 and agent2 both sow the carrots on the open ground. In Step~2 they saw that the carrots are ready to collect and they both collected one carrot in Step~3 and eventually they collected two carrots. 

\begin{figure*}
    \centering
        \includegraphics[width=1.0\linewidth]{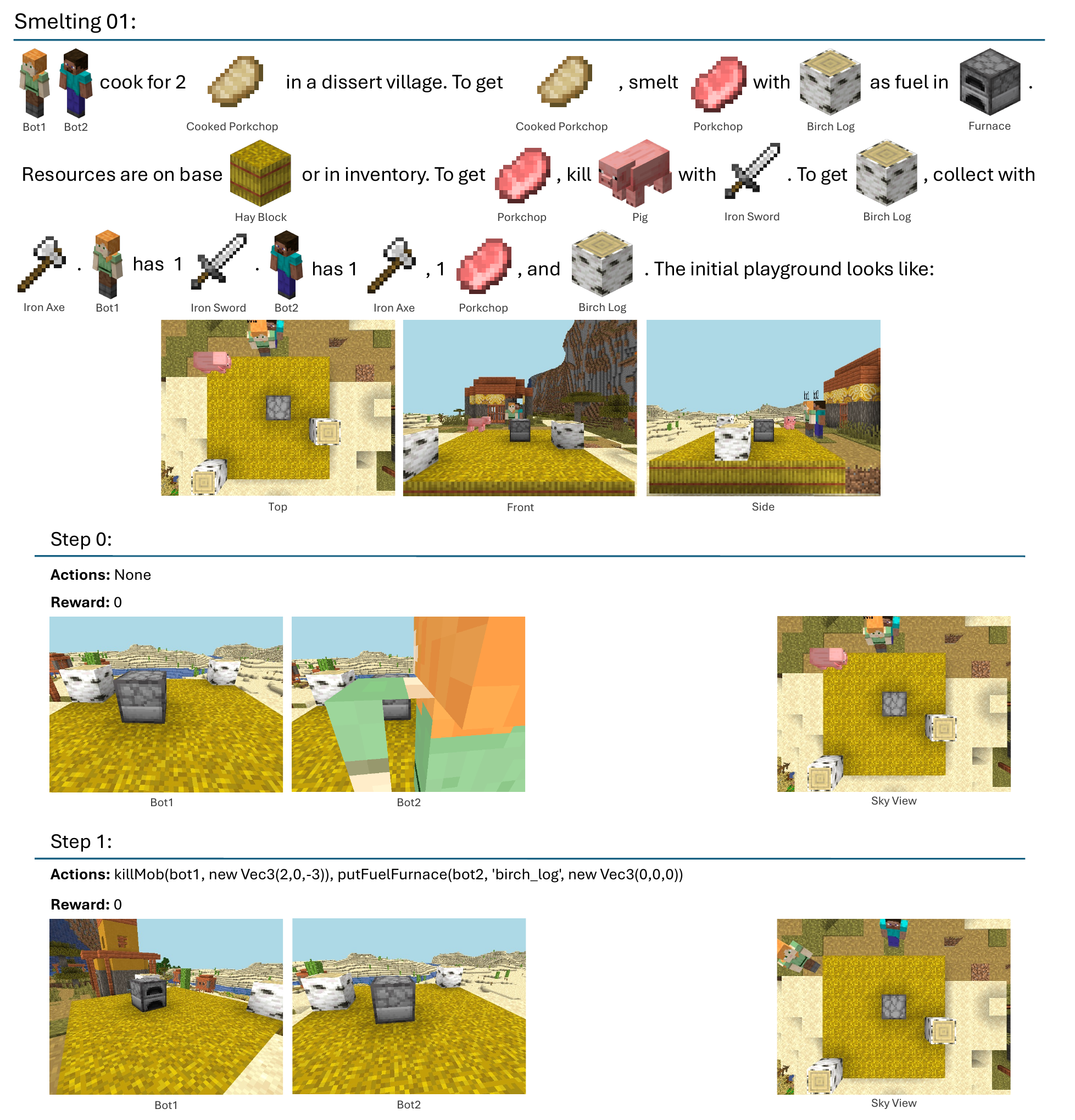}
        \caption{An example demonstration in the smelting task, Part~I.}
        \label{fig:expert_smelt_1}
\end{figure*}

\begin{figure*}
    \centering
        \includegraphics[width=1.0\linewidth]{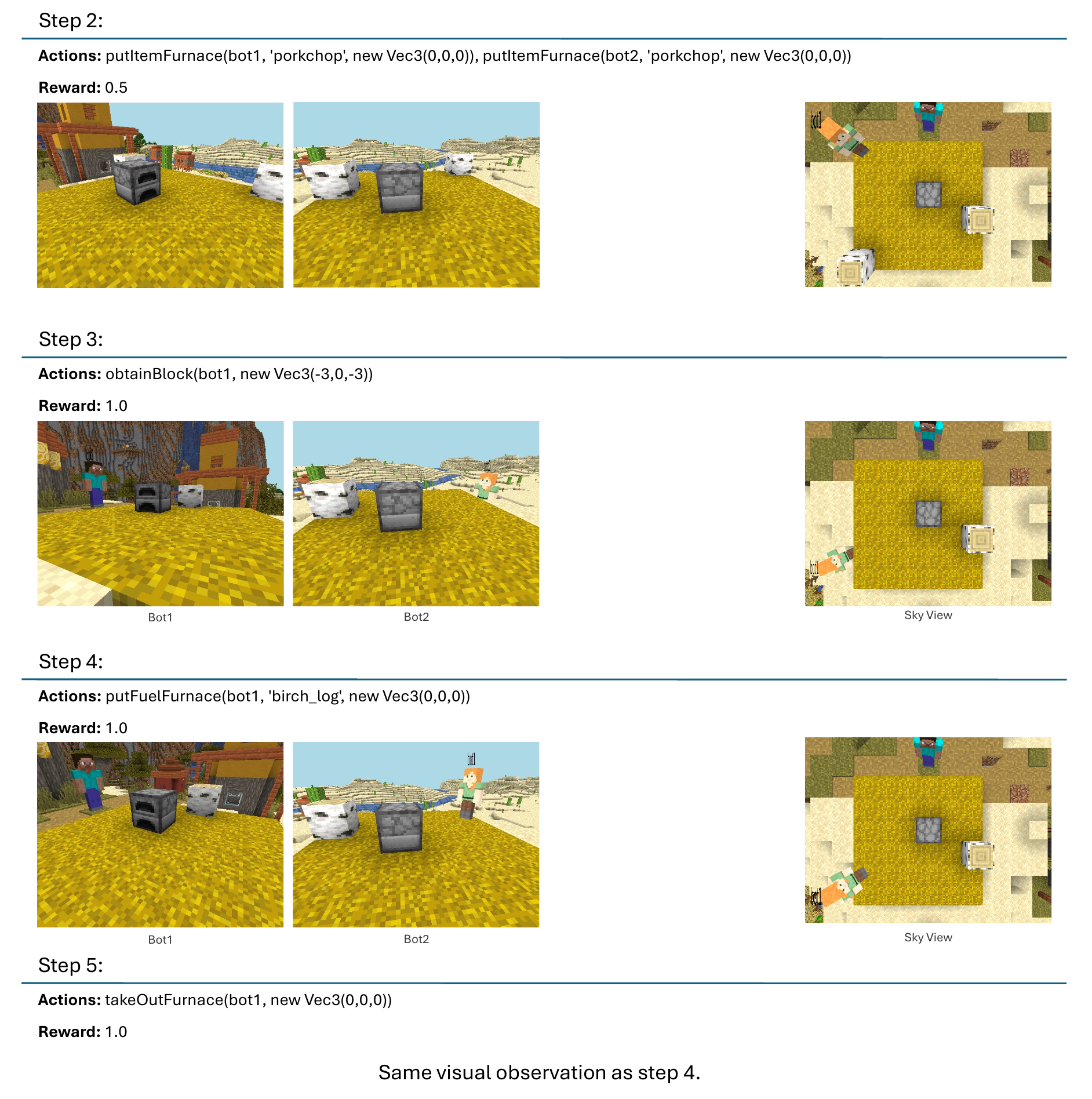}
        \caption{An example demonstration in the smelting task, Part~II.}
        \label{fig:expert_smelt_2}
\end{figure*}

\autoref{fig:expert_smelt_1} and \autoref{fig:expert_smelt_2} shows an example of the smelting task, where two agents need to cook two porkchops. In Step~1, one agent is in charge of adding the fuel to the furnace and the other agent tries to kill the pig to obtain a raw porkchop. Since bot2 already has one porkchop, it only requires one additional porkchop. In Step~2, both agents put the porkchop into the furnace and in Step~3, they have 2 cooked porkcops.

\section{Grid-World Settings}

Under the grid-world setting, we replace the three orthographic view images and first person view images with text descriptions of the task goal and current environment states, and provide them as input to the model. Here we show one example of the prompt construction in each task.

\paragraph{Building:} As shown in \autoref{fig:build-grid}, the system prompt consists of both task description and the target building coordination of each block. The user prompt consists of the built blocks and the inventories of the agents. 

\paragraph{Clearing:} As shown in \autoref{fig:break-grid}, the system prompt consists of both task description and the blocks that appeared on the platform initially. The user prompt consists of the blocks that appeared on the platform at current time step and the inventories of the agents.

\paragraph{Farming:} As shown in \autoref{fig:farm-grid}, the system prompt consists of both task description and the blocks in the farmland. The user prompt consists of the blocks in the farmland and crops information at current time step and the inventories of the agents.

\paragraph{Smelting:} As shown in \autoref{fig:smelt-grid}, the system prompt consists of both task description, instructions to craft different items and the blocks in the field. The user prompt consists of the blocks locations at current time step and the inventories of the agents.

\begin{figure*}[!ht]
    \centering
\begin{tcolorbox}[colback=gray!5!white,colframe=gray!75!black]
\textbf{System Prompt}\newline
Two bots need to grow on the platform. The goal is to get 5 carrot. farmland is on [-3 ,-1 ,-2] with value of 7. cyan\_concrete is on [-3 ,-1 ,-1]. water is on [-3 ,-1 ,0]. cyan\_concrete is on [-3 ,-1 ,1]. cyan\_concrete is on [-3 ,-1 ,2]. farmland is on [-2 ,-1 ,-2] with value of 7. cyan\_concrete is on [-2 ,-1 ,-1]. water is on [-2 ,-1 ,0]. farmland is on [-2 ,-1 ,1] with value of 7. cyan\_concrete is on [-2 ,-1 ,2]. cyan\_concrete is on [-1 ,-1 ,-2]. cyan\_concrete is on [-1 ,-1 ,-1]. water is on [-1 ,-1 ,0]. farmland is on [-1 ,-1 ,1] with value of 7. farmland is on [-1 ,-1 ,2] with value of 7. cyan\_concrete is on [0 ,-1 ,-2]. farmland is on [0 ,-1 ,-1] with value of 7. water is on [0 ,-1 ,0]. cyan\_concrete is on [0 ,-1 ,1]. cyan\_concrete is on [0 ,-1 ,2]. cyan\_concrete is on [1 ,-1 ,-2]. cyan\_concrete is on [1 ,-1 ,-1]. water is on [1 ,-1 ,0]. farmland is on [1 ,-1 ,1] with value of 7. cyan\_concrete is on [1 ,-1 ,2]. cyan\_concrete is on [2 ,-1 ,-2]. cyan\_concrete is on [2 ,-1 ,-1]. water is on [2 ,-1 ,0]. cyan\_concrete is on [2 ,-1 ,1]. farmland is on [2 ,-1 ,2] with value of 7. cyan\_concrete is on [3 ,-1 ,-2]. farmland is on [3 ,-1 ,-1] with value of 7. water is on [3 ,-1 ,0]. cyan\_concrete is on [3 ,-1 ,1]. farmland is on [3 ,-1 ,2] with value of 7. 
\end{tcolorbox}
\begin{tcolorbox}[colback=gray!5!white,colframe=gray!75!black]
\textbf{User Prompt}\newline
 bot1 has 5 carrot. bot1 has 2 beetroot. bot1 has 3 potato. bot2 has 2 carrot. bot2 has 2 beetroot. bot2 has 2 wheat\_seeds. farmland is on [-3 ,-1 ,-2] with value of 7. cyan\_concrete is on [-3 ,-1 ,-1]. water is on [-3 ,-1 ,0]. cyan\_concrete is on [-3 ,-1 ,1]. cyan\_concrete is on [-3 ,-1 ,2]. farmland is on [-2 ,-1 ,-2] with value of 7. cyan\_concrete is on [-2 ,-1 ,-1]. water is on [-2 ,-1 ,0]. farmland is on [-2 ,-1 ,1] with value of 7. cyan\_concrete is on [-2 ,-1 ,2]. cyan\_concrete is on [-1 ,-1 ,-2]. cyan\_concrete is on [-1 ,-1 ,-1]. water is on [-1 ,-1 ,0]. farmland is on [-1 ,-1 ,1] with value of 7. farmland is on [-1 ,-1 ,2] with value of 7. cyan\_concrete is on [0 ,-1 ,-2]. farmland is on [0 ,-1 ,-1] with value of 7. water is on [0 ,-1 ,0]. cyan\_concrete is on [0 ,-1 ,1]. cyan\_concrete is on [0 ,-1 ,2]. cyan\_concrete is on [1 ,-1 ,-2]. cyan\_concrete is on [1 ,-1 ,-1]. water is on [1 ,-1 ,0]. farmland is on [1 ,-1 ,1] with value of 7. cyan\_concrete is on [1 ,-1 ,2]. cyan\_concrete is on [2 ,-1 ,-2]. cyan\_concrete is on [2 ,-1 ,-1]. water is on [2 ,-1 ,0]. cyan\_concrete is on [2 ,-1 ,1]. farmland is on [2 ,-1 ,2] with value of 7. cyan\_concrete is on [3 ,-1 ,-2]. farmland is on [3 ,-1 ,-1] with value of 7. water is on [3 ,-1 ,0]. cyan\_concrete is on [3 ,-1 ,1]. farmland is on [3 ,-1 ,2] with value of 7. carrots is on [3 ,0 ,-1] with value of 0. carrots is on [3 ,0 ,2] with value of 0. Write the actions for bot1, bot2 based on this given observation.
\end{tcolorbox}
\caption{Prompt example for Farming task under the grid-world setting.}
\label{fig:farm-grid}
\end{figure*}

\begin{figure*}[!ht]
    \centering
\begin{tcolorbox}[colback=gray!5!white,colframe=gray!75!black]
\textbf{System Prompt}\newline
Two bots need to craft 2 stone. here are the instructions: Cooking Food:        1. To cook a 'cooked\_beef'... cobblestone is on [-2 ,0 ,2]. furnace is on [0 ,0 ,1]. spruce\_planks is on [2 ,0 ,-3]. cobblestone is on [2 ,0 ,-1].
\end{tcolorbox}
\begin{tcolorbox}[colback=gray!5!white,colframe=gray!75!black]
\textbf{User Prompt}\newline
bot1 has 1 iron\_sword. bot1 has 1 iron\_shovel. bot1 has 1 iron\_pickaxe. bot1 has 1 cobblestone. bot1 has 1 spruce\_planks. bot2 has 1 spruce\_planks. bot2 has 1 iron\_shovel. bot2 has 2 iron\_pickaxe. cobblestone is on [-2 ,0 ,2]. furnace is on [0 ,0 ,1]. spruce\_planks is on [2 ,0 ,-3]. cobblestone is on [2 ,0 ,-1]. Write the actions for bot1, bot2 based on this given observation.
\end{tcolorbox}
\caption{Prompt example for Smelting task under the grid-world setting.}
\label{fig:smelt-grid}
\end{figure*}

\clearpage

\section{TeamCraft-VLA Implementation Details}

We use Vicuna-v1.5 as the LLM backbone. For the visual encoder, we employ CLIP ViT-L/14 to process all input images, including three orthogonal views and the first-person view of the agents. The image embeddings are then projected into the LLM space with a linear projection layer and concatenated with the text embeddings. The combined embeddings are fed into the LLM, which outputs the final action. During training, we froze the visual encoder and projector and only finetune the LLM. All image embeddings are positioned before the text embeddings, separated by "image start" and "image end" tokens. In centralized settings, where the number of images varies depending on the number of agents, we pad a dummy image at the end for training stability if the task involves only two agents. In decentralized settings, the number of image inputs remains unaffected, as the model processes only the first-person view of the current agent, excluding views from others.

We train each model for 3 epochs using the training split, leveraging 8 A100 GPUs with a global batch size of 16. In the centralized setting, training the 7B model takes 36 hours, while the 13B model requires 72 hours. In the decentralized setting, the training duration doubles, with the 7B model requiring 72 hours and the 13B model taking 144 hours. In the grid-world setting, training the 7B model takes 20 hours.

\subsection{Arrangement of Three Orthogonal Views}

\begin{figure}
    \centering
        \includegraphics[width=0.9\linewidth]{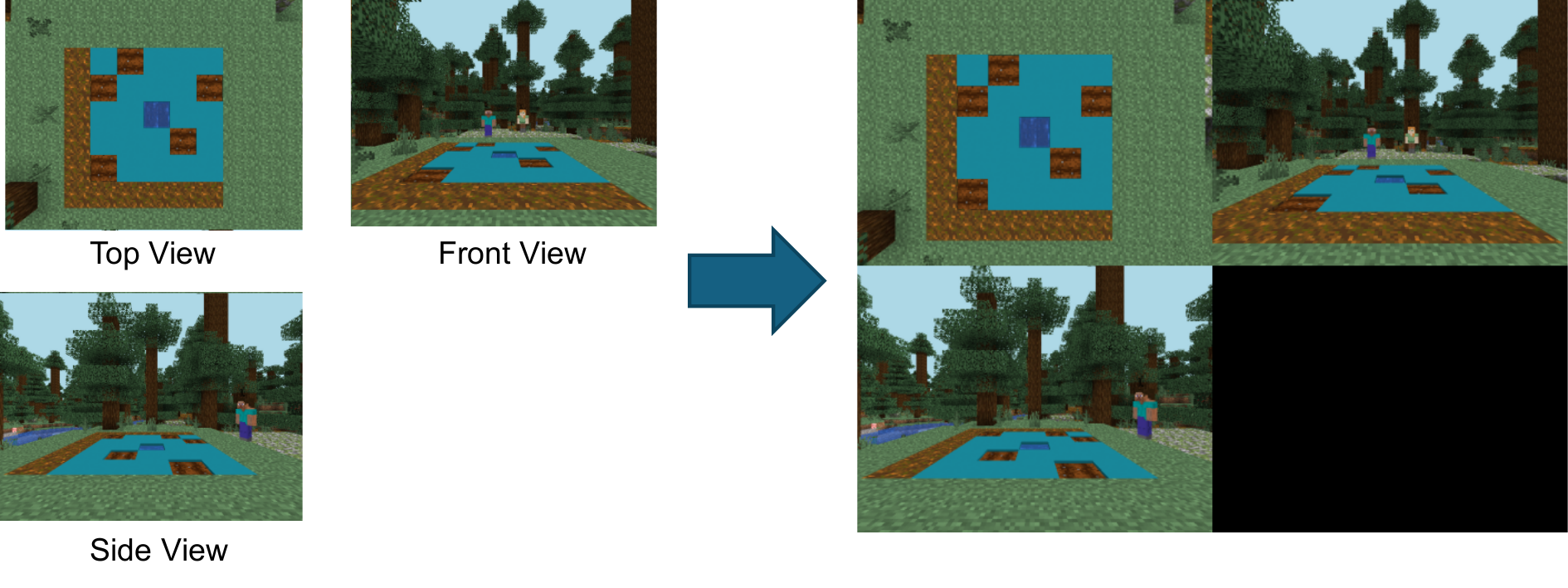}
        \caption{Combining three orthogonal view images into a single composite image as model input.}
        \label{fig:three}
\end{figure}

For training and evaluation, we combine the three orthogonal view images into a single composite image by arranging them to the upper-left top-left corner, top-right corner, and the lower-left corner of the composite image. An example of this arrangement is shown in \autoref{fig:three}. This process is to reduce the number of images provided to the model to conform with the 4096 context length limit.

\subsection{Hyperparameters}
We present the hyperparameters for VLA training in \autoref{t:hyper}.

\subsection{Model Output Parsing}

The output of the model is a string which will be parsed into the pre-defined high level skills. The string will be first processed by removing special sentence begin token, \verb|<s>|, and ending token \verb|</s>|. It will then be split into a list, where each item is parsed as the skill of one agent.

\begin{table*}
    \centering
    \caption{Hyperparameters for TeamCraft-VLA}
    \resizebox{\textwidth}{!}{\begin{tabular}{ccccccccc}
    \toprule
                lr & model max length & vision tower & patch size & resolution & language model & optimizer &lr scheduler type & warmup ratio\\ \midrule
          2e-5 & 4096 & openai-clip-vit-large & 14 & 336*336 & Vicuna-v1.5 & AdamW & constant\_with\_warmup & 0.03\\
        
        \bottomrule
    \end{tabular}}
    \label{t:hyper}
\end{table*}

\begin{table*}
\centering
\caption{Task success rates and subgoal success rates of the TeamCraft-VLA-7B-Cen and TeamCraft-VLA-7B-Dec models. Subgoal success rates are given in parentheses.}
\label{table:result_all}
\resizebox{0.8\linewidth}{!}{
\begin{tabular}{llcccccc}
\toprule
\multirow{3}{*}{Tasks} & \multirow{3}{*}{Condition} & \multicolumn{3}{c}{Centralized} & \multicolumn{3}{c}{Decentralized} \\
\cmidrule(lr){3-5} \cmidrule(lr){6-8}
 & & \multicolumn{1}{c}{10\%} & \multicolumn{1}{c}{50\%} & \multicolumn{1}{c}{100\%} & \multicolumn{1}{c}{10\%} & \multicolumn{1}{c}{50\%} & \multicolumn{1}{c}{100\%}  \\ 
\midrule  
\multirow{5}{*}{Building} 
 & Test       & 0.00 (0.12) & 0.38 (0.76) & 0.42 (0.81) & 0.00 (0.18) & 0.00 (0.28) & 0.00 (0.38) \\
 & Shape       & 0.00 (0.12) & 0.20 (0.67) & 0.30 (0.75) & 0.00 (0.15) & 0.00 (0.25) & 0.00 (0.40) \\
 & Material     & 0.00 (0.13) & 0.18 (0.64) & 0.30 (0.74) & 0.00 (0.13) & 0.00 (0.20) & 0.00 (0.34) \\
 & Scene     & 0.00 (0.15) & 0.36 (0.73) & 0.40 (0.83) & 0.00 (0.16) & 0.00 (0.21) & 0.00 (0.36) \\
 & Agents & 0.00 (0.18) & 0.02 (0.50) & 0.02 (0.57) & 0.00 (0.12) & 0.00 (0.20) & 0.00 (0.14) \\
\midrule
\multirow{5}{*}{Clearing} 
 & Test       & 0.00 (0.13) & 0.08 (0.43) & 0.64 (0.91) & 0.00 (0.45) & 0.02 (0.35) & 0.20 (0.68) \\
 & Shape       & 0.00 (0.09) & 0.08 (0.34) & 0.56 (0.91) & 0.00 (0.47) & 0.02 (0.27) & 0.16 (0.74) \\
 & Material     & 0.00 (0.10) & 0.12 (0.45) & 0.56 (0.90) & 0.00 (0.48) & 0.00 (0.22) & 0.16 (0.67) \\
 & Scene     & 0.00 (0.11) & 0.10 (0.44) & 0.58 (0.92) & 0.00 (0.41) & 0.04 (0.37) & 0.10 (0.64) \\
 & Agents & 0.00 (0.16) & 0.14 (0.64) & 0.36 (0.81) & 0.02 (0.50) & 0.02 (0.54) & 0.12 (0.60) \\
\midrule
\multirow{5}{*}{Farming}  
 & Test       & 0.14 (0.43) & 0.34 (0.60) & 0.36 (0.63) & 0.02 (0.07) & 0.02 (0.14) & 0.00 (0.09) \\
 & Crop       & 0.00 (0.00) & 0.00 (0.00) & 0.00 (0.00) & 0.00 (0.00) & 0.00 (0.00) & 0.00 (0.00) \\
 & Scene     & 0.16 (0.39) & 0.34 (0.65) & 0.38 (0.67) & 0.00 (0.05) & 0.00 (0.11) & 0.02 (0.07) \\
 & Agents     & 0.02 (0.18) & 0.18 (0.61) & 0.38 (0.68) & 0.00 (0.08) & 0.00 (0.11) & 0.04 (0.27) \\
\midrule
\multirow{5}{*}{Smelting}  
 & Test       & 0.06 (0.17) & 0.20 (0.36) & 0.24 (0.28) & 0.08 (0.13) & 0.08 (0.09) & 0.16 (0.29) \\
 & Goal       & 0.08 (0.21) & 0.04 (0.07) & 0.00 (0.00) & 0.08 (0.17) & 0.00 (0.00) & 0.00 (0.00) \\
 & Furnace     & 0.10 (0.28) & 0.10 (0.20) & 0.18 (0.20) & 0.06 (0.07) & 0.06 (0.06) & 0.06 (0.16) \\
 & Scene     & 0.08 (0.19) & 0.14 (0.28) & 0.18 (0.23) & 0.08 (0.19) & 0.14 (0.19) & 0.12 (0.28) \\
 & Agents & 0.00 (0.15) & 0.02 (0.24) & 0.06 (0.13) & 0.04 (0.05) & 0.00 (0.02) & 0.02 (0.28) \\
\bottomrule
\end{tabular}
}
\end{table*}

\begin{table*}
\centering
\caption{Task success rates and subgoal success rates of the TeamCraft-VLA-13B-Cen and TeamCraft-VLA-13B-Dec models. Subgoal success rates are given in parentheses.}
\label{table:13b-result_all}
\resizebox{0.8\linewidth}{!}{
\begin{tabular}{llcccccc}
\toprule
\multirow{3}{*}{Tasks} & \multirow{3}{*}{Condition} & \multicolumn{3}{c}{Centralized} & \multicolumn{3}{c}{Decentralized} \\
\cmidrule(lr){3-5} \cmidrule(lr){6-8}
 & & \multicolumn{1}{c}{10\%} & \multicolumn{1}{c}{50\%} & \multicolumn{1}{c}{100\%} & \multicolumn{1}{c}{10\%} & \multicolumn{1}{c}{50\%} & \multicolumn{1}{c}{100\%}  \\ 
\midrule  
\multirow{5}{*}{Building} 
 & Test       & 0.00 (0.18) & 0.46 (0.80) & 0.48 (0.79) & 0.00 (0.13) & 0.00 (0.18) & 0.00 (0.31) \\
 & Shape       & 0.00 (0.16) & 0.30 (0.73) & 0.26 (0.69) & 0.00 (0.15) & 0.00 (0.15) & 0.00 (0.32) \\
 & Material     & 0.00 (0.15) & 0.24 (0.65) & 0.08 (0.63) & 0.00 (0.14) & 0.00 (0.14) & 0.00 (0.31) \\
 & Scene     & 0.00 (0.16) & 0.38 (0.75) & 0.48 (0.83) & 0.00 (0.17) & 0.00 (0.17) & 0.00 (0.28) \\
 & Agents & 0.00 (0.16) & 0.00 (0.49) & 0.04 (0.59) & 0.00 (0.14) & 0.00 (0.16) & 0.00 (0.23) \\
\midrule
\multirow{5}{*}{Clearing} 
 & Test       & 0.04 (0.37) & 0.42 (0.83) & 0.64 (0.94) & 0.00 (0.46) & 0.02 (0.62) & 0.02 (0.60) \\
 & Shape       & 0.00 (0.26) & 0.42 (0.85) & 0.78 (0.96) & 0.00 (0.47) & 0.00 (0.57) & 0.04 (0.58) \\
 & Material     & 0.04 (0.36) & 0.36 (0.83) & 0.56 (0.92) & 0.02 (0.53) & 0.00 (0.60) & 0.02 (0.58) \\
 & Scene     & 0.06 (0.35) & 0.44 (0.88) & 0.48 (0.90) & 0.00 (0.55) & 0.02 (0.59) & 0.08 (0.64) \\
 & Agents & 0.02 (0.55) & 0.16 (0.65) & 0.16 (0.77) & 0.02 (0.50) & 0.02 (0.52) & 0.02 (0.50) \\
\midrule
\multirow{5}{*}{Farming}  
 & Test       & 0.4 (0.72) & 0.62 (0.79) & 0.46 (0.73) & 0.08 (0.39) & 0.04 (0.23) & 0.02 (0.33) \\
 & Crop       & 0.00 (0.00) & 0.00 (0.00) & 0.00 (0.00) & 0.00 (0.00) & 0.00 (0.00) & 0.00 (0.00) \\
 & Scene     & 0.30 (0.69) & 0.52 (0.76) & 0.44 (0.75) & 0.04 (0.32) & 0.06 (0.29) & 0.10 (0.33) \\
 & Agents     & 0.12 (0.54) & 0.44 (0.79) & 0.36 (0.72) & 0.02 (0.22) & 0.00 (0.19) & 0.02 (0.23) \\
\midrule
\multirow{5}{*}{Smelting}  
 & Test       & 0.06 (0.08) & 0.22 (0.44) & 0.32 (0.59) & 0.10 (0.25) & 0.06 (0.09) & 0.10 (0.19) \\
 & Goal       & 0.00 (0.00) & 0.00 (0.00) & 0.00 (0.00) & 0.00 (0.10) & 0.00 (0.00) & 0.00 (0.00) \\
 & Furnace     & 0.06 (0.08) & 0.20 (0.40) & 0.18 (0.38) & 0.06 (0.12) & 0.04 (0.08) & 0.04 (0.07) \\
 & Scene     & 0.04 (0.08) & 0.16 (0.43) & 0.24 (0.56) & 0.12 (0.28) & 0.04 (0.09) & 0.08 (0.18) \\
 & Agents & 0.00 (0.03) & 0.00 (0.26) & 0.04 (0.37) & 0.00 (0.02) & 0.00 (0.01) & 0.00 (0.00) \\
\bottomrule
\end{tabular}
}
\end{table*}

\begin{table*}
\centering
\caption{Task success rates and subgoal success rates of various centralized models. Subgoal success rates are given in parentheses. All models are trained with the full training data except GPT-4o.}
\label{table:ablation-model}
\resizebox{0.8\linewidth}{!}{
\begin{tabular}{llccccccc}
\toprule
\multirow{1}{*}{Tasks} & \multirow{1}{*}{Condition} & \multicolumn{1}{c}{TeamCraft-VLA-7B} & \multicolumn{1}{c}{TeamCraft-VLA-13B} & \multicolumn{1}{c}{GPT-4o} & \multicolumn{1}{c}{TeamCraft-7B-GridWorld}\\
\midrule  
\multirow{5}{*}{Building} 
 & Test       & 0.42 (0.81) & 0.48 (0.79) & 0.00 (0.07) & 0.42 (0.88)\\
 & Shape       & 0.30 (0.75) & 0.26 (0.69) & 0.00 (0.08) & 0.50 (0.90) \\
 & Material     & 0.30 (0.74) & 0.08 (0.63) & 0.00 (0.07) & 0.26 (0.82) \\
 & Scene     & 0.40 (0.83) & 0.48 (0.83) & 0.00 (0.07) & 0.48 (0.89) \\
 & Agents & 0.02 (0.57) & 0.04 (0.59) &   0.00 (0.00) & 0.12 (0.71)\\
\midrule
\multirow{5}{*}{Clearing} 
 & Test       & 0.64 (0.91) & 0.64 (0.94) & 0.00 (0.03)   & 1.00 (1.00)\\
 & Shape       & 0.56 (0.91) & 0.78 (0.96) & 0.00 (0.04) & 1.00 (1.00) \\
 & Material    & 0.56 (0.91) & 0.56 (0.92) & 0.00 (0.12) & 1.00 (1.00) \\
 & Scene     & 0.58 (0.92) & 0.48 (0.90) & 0.00 (0.06) & 1.00 (1.00)\\
 & Agents  & 0.36 (0.81) & 0.16 (0.77) &   0.00 (0.00) & 0.84 (0.97)\\
\midrule
\multirow{5}{*}{Farming}  
 & Test       & 0.36 (0.64) & 0.46 (0.73) & 0.00 (0.00) & 0.78 (0.86) \\
 & Crop       & 0.00 (0.00) & 0.00 (0.00) & 0.00 (0.00) & 0.00 (0.00) \\
 & Scene     & 0.38 (0.67) & 0.44 (0.75) &  0.00 (0.00) & 0.90 (0.96)\\
 & Agents     & 0.38 (0.68) & 0.36 (0.72) & 0.00 (0.00) & 0.40 (0.73) \\
\midrule
\multirow{5}{*}{Smelting}  
 & Test       & 0.24 (0.28) & 0.32 (0.59) & 0.02 (0.02)  & 0.24 (0.51)\\
 & Goal       & 0.00 (0.00) & 0.00 (0.00) & 0.08 (0.08)  & 0.00 (0.00)\\
 & Furnace     & 0.18 (0.20) & 0.18 (0.38) & 0.00 (0.00) & 0.24 (0.39) \\
 & Scene     & 0.18 (0.23) & 0.24 (0.56) & 0.00 (0.00)  & 0.36 (0.58) \\
 & Agents & 0.06 (0.13) & 0.04 (0.37) &   0.00 (0.00) & 0.00 (0.31)\\
\bottomrule
\end{tabular}
}
\end{table*}

\section{Additional Results of TeamCraft-VLA}

\subsection{Task Success Rate and Subgoal Success Rate}
We show task success rate and subgoal success rate of centralized and decentralized 7B models with different data scales in \autoref{table:result_all}, and those of 13B models in \autoref{table:13b-result_all}. We compare among different centralized models in \autoref{table:ablation-model}.

\subsection{Redundancy Rate}
We present a more detailed analysis of redundancy rates, including the 13B models, in \autoref{t:redundancy_2}. Both the 7B and 13B models exhibit redundancy issues in decentralized settings. Increasing model size alone does not resolve the redundancy problem in such scenarios.

\subsection{Action Sequence Length}
We compared the average action lengths across different splits between the 7B and 13B models under both centralized and decentralized settings, as shown in \autoref{t:action-length}. In general, decentralized settings require longer action sequences to complete tasks. Among the splits, the \textit{Goal} split is the most challenging, as it demands more actions to accomplish the tasks.

\begin{table}
    \centering
    \caption{Comparison of TeamCraft-VLA redundancy rates.}
    \resizebox{\linewidth}{!}{\begin{tabular}{lcccccc}
    \toprule
               & Test & Goal & Scene & Agents & Average \\ 
               \midrule
        TeamCraft-VLA-7B-Cen & 0.01 & 0.02 & 0.01& 0.01 & 0.01 \\
        TeamCraft-VLA-13B-Cen & 0.01 & 0.00 & 0.01& 0.01 & 0.01 \\
        TeamCraft-VLA-7B-Dec & 0.13 & 0.12 & 0.13& 0.24 & 0.15 \\
        TeamCraft-VLA-13B-Dec & 0.11 & 0.11 & 0.12& 0.22 & 0.14 \\
         
        \bottomrule
    \end{tabular}}
    \label{t:redundancy_2}
\end{table}

\begin{table}
    \centering
    \caption{Comparison of TeamCraft-VLA action sequence length.}
    \resizebox{\linewidth}{!}{\begin{tabular}{lcccccc}
    \toprule
               & Test & Goal & Scene & Agents & Average \\ 
               \midrule
        TeamCraft-VLA-7B-Cen & 6.62 & 7.63 & 5.93& 6.35 & 6.63 \\
        TeamCraft-VLA-13B-Cen & 6.25 & 7.44 & 6.46& 6.47 & 6.65 \\
        TeamCraft-VLA-7B-Dec & 8.42 & 8.53 & 8.06& 7.38 & 8.1 \\
        TeamCraft-VLA-13B-Dec & 8.62 & 8.46 & 8.41& 6.71 & 8.04 \\
         
        \bottomrule
    \end{tabular}}
    \label{t:action-length}
\end{table}

\subsection{Case Study}

\begin{figure*}[!ht]
\centering
\begin{tcolorbox}[colback=gray!5!white,colframe=gray!75!black]
\textbf{System Prompt}\newline
Two bots need to grow on the platform. The goal is to get 10 beetroot. Write the actions for bot1, bot2 based on this given observation."\newline
\begin{subfigure}[b]{0.23\textwidth}
        \includegraphics[width=\textwidth]{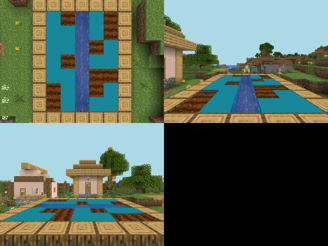}
        \caption{orthographic views of the task}
    \end{subfigure}
\end{tcolorbox}
\begin{tcolorbox}
\textbf{Step~0:}
\begin{tcolorbox}[colback=gray!5!white,colframe=gray!75!black]
\textbf{User Prompt}\newline
bot1 has 7 beetroot\_seeds. bot1 has 1 carrot. bot1 has 3 potato. bot2 has 5 beetroot\_seeds. bot2 has 2 wheat\_seeds. Write the actions for bot1, bot2 based on this given observation.\newline
    \begin{subfigure}[b]{0.23\textwidth}
        \includegraphics[width=\textwidth]{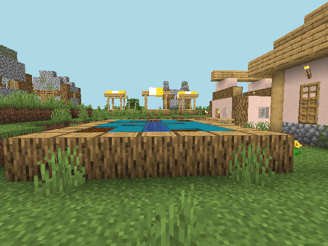}
        \caption{Agent1 observation}
    \end{subfigure}
    \begin{subfigure}[b]{0.23\textwidth}
        \includegraphics[width=\textwidth]{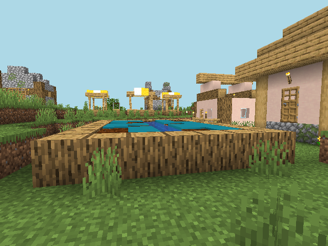}
        \caption{Agent2 observation}
    \end{subfigure}    
\end{tcolorbox}
\begin{tcolorbox}[colback=gray!5!white,colframe=gray!75!black]
\textbf{Actions}\newline
["farm\_work(bot1, new Vec3(-1,-1,2), 'sow', 'beat\_seeds')",\\
"farm\_work(bot2, new Vec3(-2,-1,-2), , 'sow', 'beat\_seeds')"]
\end{tcolorbox}
\end{tcolorbox}
\begin{tcolorbox}
\textbf{Step~1:}
\begin{tcolorbox}[colback=gray!5!white,colframe=gray!75!black]
\textbf{User Prompt}\newline
bot1 has 7 beetroot\_seeds. bot1 has 1 carrot. bot1 has 3 potato. bot2 has 5 beetroot\_seeds. bot2 has 2 wheat\_seeds. Write the actions for bot1, bot2 based on this given observation.\newline
    \begin{subfigure}[b]{0.23\textwidth}
        \includegraphics[width=\textwidth]{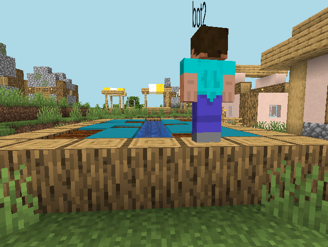}
        \caption{Agent1 observation}
    \end{subfigure}
    \begin{subfigure}[b]{0.23\textwidth}
        \includegraphics[width=\textwidth]{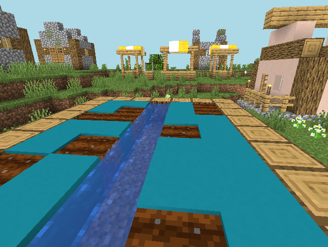}
        \caption{Agent2 observation}
    \end{subfigure}   
\end{tcolorbox}
\begin{tcolorbox}[colback=gray!5!white,colframe=gray!75!black]
\textbf{Actions}\newline
["farm\_work(bot1, new Vec3(3,-1,-1), 'sow', 'beat\_seeds')",\\
"farm\_work(bot2, new Vec3(0,-1,-1), 'sow', 'beat\_seeds')"]
\end{tcolorbox}
\end{tcolorbox}
\caption{Object mismatching Example 1.}
\label{fig:e1}
\end{figure*}

\begin{figure*}[!ht]
\centering
\begin{tcolorbox}[colback=gray!5!white,colframe=gray!75!black]
\textbf{System Prompt}\newline
"Two bots need to craft 2 dried\_kelp. here are the instructions: … To cook a 'dried\_kelp', I need a 'kelp'. ... ",
\begin{subfigure}[b]{0.23\textwidth}
        \includegraphics[width=\textwidth]{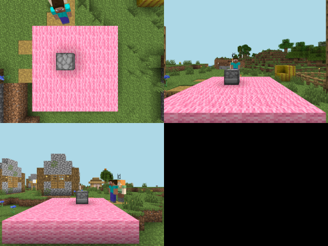}
        \caption{orthographic views of the task}
    \end{subfigure}
\end{tcolorbox}
\begin{tcolorbox}
\textbf{Step~0:}
\begin{tcolorbox}[colback=gray!5!white,colframe=gray!75!black]
\textbf{User Prompt}\newline
bot1 has 1 iron\_pickaxe. bot1 has 1 iron\_axe. bot2 has 2 charcoal. bot2 has 1 iron\_shovel. bot2 has 1 iron\_sword. Write the actions for bot1, bot2 based on this given observation.\newline
    \begin{subfigure}[b]{0.23\textwidth}
        \includegraphics[width=\textwidth]{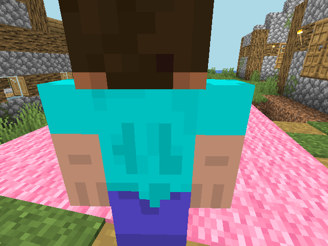}
        \caption{Agent1 observation}
    \end{subfigure}
    \begin{subfigure}[b]{0.23\textwidth}
        \includegraphics[width=\textwidth]{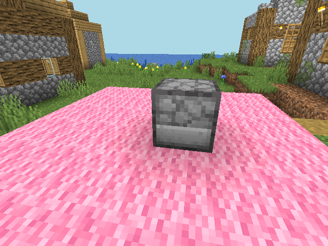}
        \caption{Agent2 observation}
    \end{subfigure} 
\end{tcolorbox}
\begin{tcolorbox}[colback=gray!5!white,colframe=gray!75!black]
\textbf{Actions}\newline
[putItemFurnace(bot1, 'cobbl1', new Vec3(0,0,-1)),\\
putFuelFurnace(bot2, 'charcoal', new Vec3(0,0,-1))]
\end{tcolorbox}
\end{tcolorbox}
\begin{tcolorbox}
\textbf{Step~1:}
\begin{tcolorbox}[colback=gray!5!white,colframe=gray!75!black]
\textbf{User Prompt}\newline
        bot1 has 2 kelp. bot1 has 1 iron\_pickaxe. bot1 has 1 iron\_axe. bot2 has 1 charcoal. bot2 has 1 iron\_shovel. bot2 has 1 iron\_sword. Write the actions for bot1, bot2 based on this given observation.\newline
    \begin{subfigure}[b]{0.23\textwidth}
        \includegraphics[width=\textwidth]{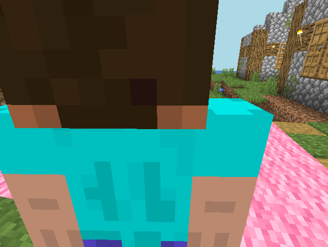}
        \caption{Agent1 observation}
    \end{subfigure}
    \begin{subfigure}[b]{0.23\textwidth}
        \includegraphics[width=\textwidth]{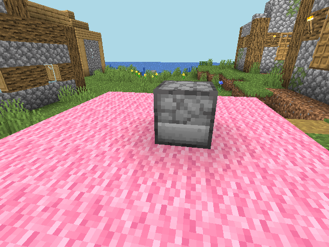}
        \caption{Agent2 observation}
    \end{subfigure}  
\end{tcolorbox}
\begin{tcolorbox}[colback=gray!5!white,colframe=gray!75!black]
\textbf{Actions}\newline
[putItemFurnace(bot1, 'cobbl1', new Vec3(0,0,-1)),\\
putFuelFurnace(bot2, 'charcoal', new Vec3(0,0,-1))]
\end{tcolorbox}
\end{tcolorbox}
\caption{Object mismatching Example 2.}
\label{fig:e2}
\end{figure*}

We next present a detailed failure case analysis by categories:

\paragraph{Object Mismatching:} As an example (\autoref{fig:e1}), in the farming tasks two agents need to get 10 beetroot. In Step~0, the actions involve a mismatch in the objects; the agents mistakenly sow "beet\_seeds" instead of "beetroot\_seeds." Consequently, in Step~1, due to the object mismatch, no crops grow on the farmland. As another example, two agents need to get 2 dried kelp in the smelting task (\autoref{fig:e2}). The task requires one bot to put the kelp and the other put the fuel. However, in this example bot1 mistake the object "kelt" to "cobbl1".

\begin{figure*}[!ht]
\centering
\begin{tcolorbox}[colback=gray!5!white,colframe=gray!75!black]
\textbf{System Prompt}\newline
"Four bots need to break everything on the platform."
",\newline
\begin{subfigure}[b]{0.23\textwidth}
        \includegraphics[width=\textwidth]{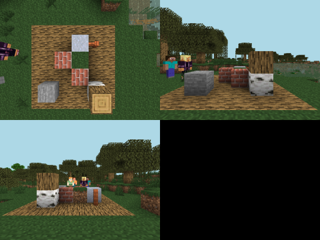}
        \caption{orthographic views of the task}
    \end{subfigure}
\end{tcolorbox}
\begin{tcolorbox}
\textbf{Step~0:}
\begin{tcolorbox}[colback=gray!5!white,colframe=gray!75!black]
\textbf{User Prompt}\newline
bot0 has 1 stone\_axe. bot1 has 1 stone\_shovel. bot2 has 1 stone\_pickaxe. bot3 has 1 stone\_sword. Write the actions for bot0, bot1, bot2, bot3 based on this given observation.\newline
    \begin{subfigure}[b]{0.23\textwidth}
        \includegraphics[width=\textwidth]{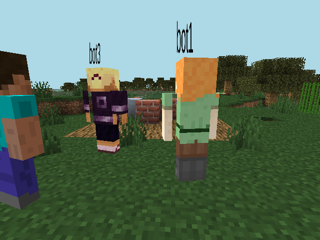}
        \caption{Agent0 observation}
    \end{subfigure}
    \begin{subfigure}[b]{0.23\textwidth}
        \includegraphics[width=\textwidth]{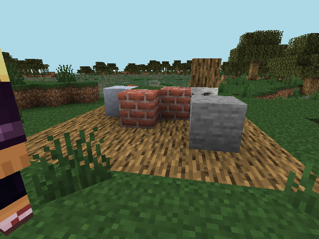}
        \caption{Agent1 observation}
    \end{subfigure}
    \begin{subfigure}[b]{0.23\textwidth}
        \includegraphics[width=\textwidth]{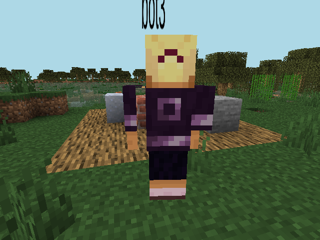}
        \caption{Agent2 observation}
    \end{subfigure}
    \begin{subfigure}[b]{0.23\textwidth}
        \includegraphics[width=\textwidth]{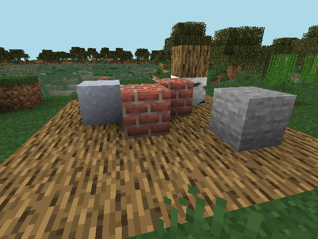}
        \caption{Agent3 observation}
    \end{subfigure}
\end{tcolorbox}
\begin{tcolorbox}[colback=gray!5!white,colframe=gray!75!black]
\textbf{Actions}\newline
[" mineBlock(bot1, new Vec3(-2,0,1)),\\
mineBlock(bot2, new Vec3(-1,0,0)),\\
mineBlock(bot3, new Vec3(1,0,1))", ]
\end{tcolorbox}
\end{tcolorbox}
\begin{tcolorbox}
\textbf{Step~1:}
\begin{tcolorbox}[colback=gray!5!white,colframe=gray!75!black]
\textbf{User Prompt}\newline
      Write the actions for bot0, bot1, bot2, bot3 based on this given observation."\newline
    \begin{subfigure}[b]{0.23\textwidth}
        \includegraphics[width=\textwidth]{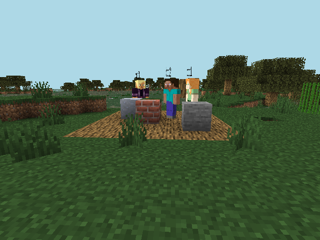}
        \caption{Agent0 observation}
    \end{subfigure}
    \begin{subfigure}[b]{0.23\textwidth}
        \includegraphics[width=\textwidth]{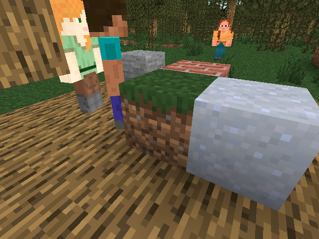}
        \caption{Agent1 observation}
    \end{subfigure}
    \begin{subfigure}[b]{0.23\textwidth}
        \includegraphics[width=\textwidth]{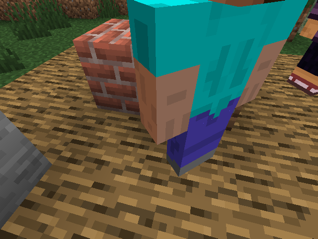}
        \caption{Agent2 observation}
    \end{subfigure}
    \begin{subfigure}[b]{0.23\textwidth}
        \includegraphics[width=\textwidth]{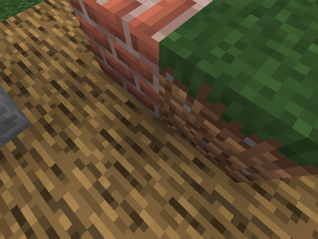}
        \caption{Agent3 observation}
    \end{subfigure}
\end{tcolorbox}
\begin{tcolorbox}[colback=gray!5!white,colframe=gray!75!black]
\textbf{Actions}\newline
[" mineBlock(bot1, new Vec3(1,0,0)),\\
mineBlock(bot2, new Vec3(-2,0,-2)),\\
mineBlock(bot3, new Vec3(1,0,1))",
]
\end{tcolorbox}
\end{tcolorbox}
\caption{Task allocation failure Example 3.}
\label{fig:e3}
\end{figure*}

\begin{figure*}[!ht]
    \centering
\begin{tcolorbox}[colback=gray!5!white,colframe=gray!75!black]
\textbf{System Prompt}\newline
"Four bots need to build a building on the platform. "\newline
\begin{subfigure}[b]{0.23\textwidth}
        \includegraphics[width=\textwidth]{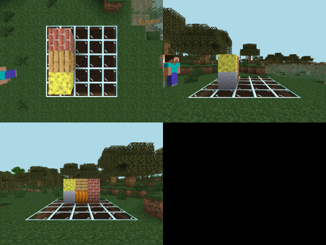}
        \caption{orthographic views of the task}
    \end{subfigure}
\end{tcolorbox}
\begin{tcolorbox}
\textbf{Step~0:}
\begin{tcolorbox}[colback=gray!5!white,colframe=gray!75!black]
\textbf{User Prompt}\newline
        bot0 has 4 bricks. bot0 has 5 sponge. bot0 has 2 oak\_planks...bot1 has 1 pumpkin... bot2 has 3 bricks... bot3 has 1 clay... Write the actions for bot0, bot1, bot2, bot3 based on this given observation.\newline
    \begin{subfigure}[b]{0.23\textwidth}
        \includegraphics[width=\textwidth]{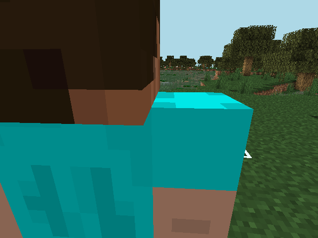}
        \caption{Agent0 observation}
    \end{subfigure}
    \begin{subfigure}[b]{0.23\textwidth}
        \includegraphics[width=\textwidth]{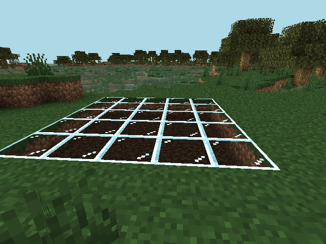}
        \caption{Agent1 observation}
    \end{subfigure}
    \begin{subfigure}[b]{0.23\textwidth}
        \includegraphics[width=\textwidth]{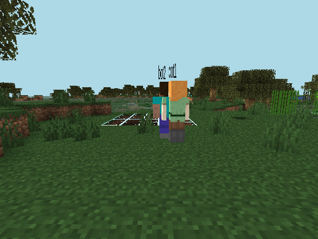}
        \caption{Agent2 observation}
    \end{subfigure}
    \begin{subfigure}[b]{0.23\textwidth}
        \includegraphics[width=\textwidth]{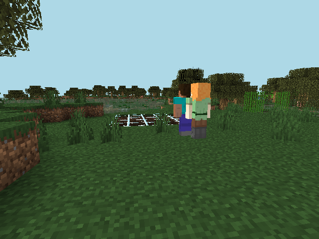}
        \caption{Agent3 observation}
    \end{subfigure}   
\end{tcolorbox}
\begin{tcolorbox}[colback=gray!5!white,colframe=gray!75!black]
\textbf{Actions}\newline
[        " placeItem(bot1, 'bricks', new Vec3(0,0,-1)),\\ placeItem(bot2, 'clay', new Vec3(-1,0,-1)),\\ placeItem(bot3, 'clay', new Vec3(1,0,-1))"
]
\end{tcolorbox}
\end{tcolorbox}
\begin{tcolorbox}
\textbf{Step~1:}
\begin{tcolorbox}[colback=gray!5!white,colframe=gray!75!black]
\textbf{User Prompt}\newline
      Write the actions for bot0, bot1, bot2, bot3 based on this given observation.\newline
    \begin{subfigure}[b]{0.23\textwidth}
        \includegraphics[width=\textwidth]{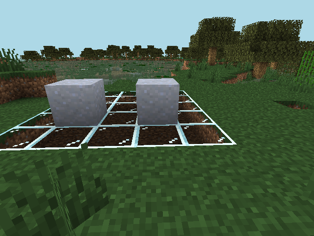}
        \caption{Agent0 observation}
    \end{subfigure}
    \begin{subfigure}[b]{0.23\textwidth}
        \includegraphics[width=\textwidth]{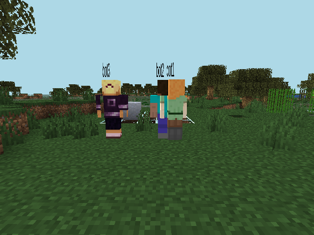}
        \caption{Agent1 observation}
    \end{subfigure}
    \begin{subfigure}[b]{0.23\textwidth}
        \includegraphics[width=\textwidth]{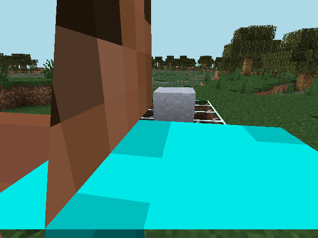}
        \caption{Agent2 observation}
    \end{subfigure}
    \begin{subfigure}[b]{0.23\textwidth}
        \includegraphics[width=\textwidth]{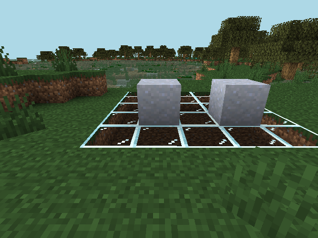}
        \caption{Agent3 observation}
    \end{subfigure}
\end{tcolorbox}
\begin{tcolorbox}[colback=gray!5!white,colframe=gray!75!black]
\textbf{Actions}\newline
["placeItem(bot1, 'sponge', new Vec3(-1,1,-1)),\\placeItem(bot2, 'bricks', new Vec3(1,1,-1)),\\placeItem(bot3, 'sponge', new Vec3(0,1,-1))",
]
\end{tcolorbox}
\end{tcolorbox}
\caption{Task allocation failure Example 4.}
\label{fig:e4}
\end{figure*}

\paragraph{Task Allocation Failure:} This occurs when a task requires four agents. As two examples, four agents must break everything on the platform in the clearing task (\autoref{fig:e3}), and construct on the platform in the building task (\autoref{fig:e4}). Only three agents are assigned distinct actions, leaving the fourth agent idle.

\begin{figure*}[!ht]
    \centering
\begin{tcolorbox}[colback=gray!5!white,colframe=gray!75!black]
\textbf{System Prompt}\newline
Two bots need to grow on the platform. The goal is to get 4 carrot.\newline
\begin{subfigure}[b]{0.23\textwidth}
        \includegraphics[width=\textwidth]{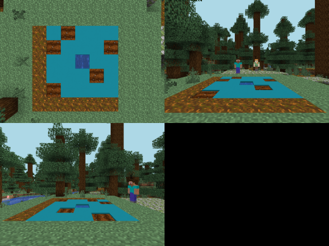}
        \caption{orthographic views of the task}
    \end{subfigure}
\end{tcolorbox}
\begin{tcolorbox}
\textbf{Step~0:}
\begin{tcolorbox}[colback=gray!5!white,colframe=gray!75!black]
\textbf{User Prompt}\newline
        "bot1 has 4 carrot. bot1 has 1 potato. bot1 has 1 beetroot. bot2 has 5 carrot. bot2 has 2 beetroot. bot2 has 3 wheat\_seeds. bot2 has 2 potato. Write the actions for bot1, bot2 based on this given observation."\newline
    \begin{subfigure}[b]{0.23\textwidth}
        \includegraphics[width=\textwidth]{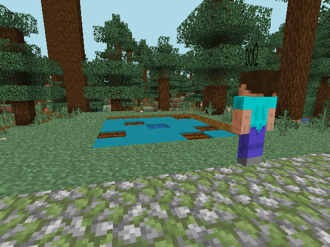}
        \caption{Agent1 observation}
    \end{subfigure}
    \begin{subfigure}[b]{0.23\textwidth}
        \includegraphics[width=\textwidth]{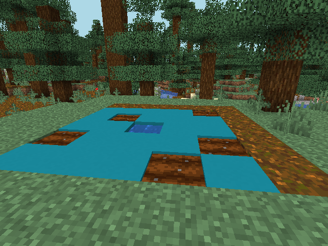}
        \caption{Agent2 observation}
    \end{subfigure}
\end{tcolorbox}
\begin{tcolorbox}[colback=gray!5!white,colframe=gray!75!black]
\textbf{Actions}\newline
[" farm\_work(bot1, new Vec3(1,-1,-2), 'sow', 'carrot'),\\farm\_work(bot2, new Vec3(2,-1,-2), 'sow', 'carrot')",
]
\end{tcolorbox}
\end{tcolorbox}
\begin{tcolorbox}
\textbf{Step~2:}
\begin{tcolorbox}[colback=gray!5!white,colframe=gray!75!black]
\textbf{User Prompt}\newline
               "bot1 has 3 carrot. bot1 has 1 potato. bot1 has 1 beetroot. bot2 has 5 carrot. bot2 has 2 beetroot. bot2 has 3 wheat\_seeds. bot2 has 2 potato. Write the actions for bot1, bot2 based on this given observation."\newline
    \begin{subfigure}[b]{0.23\textwidth}
        \includegraphics[width=\textwidth]{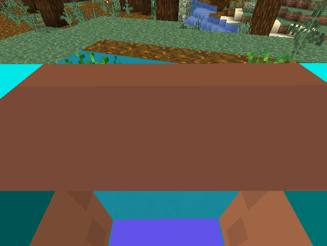}
        \caption{Agent1 observation}
    \end{subfigure}
    \begin{subfigure}[b]{0.23\textwidth}
        \includegraphics[width=\textwidth]{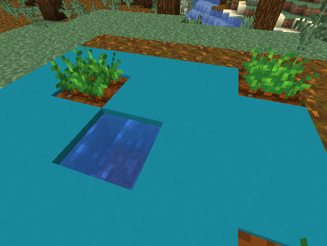}
        \caption{Agent2 observation}
    \end{subfigure} 
\end{tcolorbox}
\begin{tcolorbox}[colback=gray!5!white,colframe=gray!75!black]
\textbf{Actions}\newline
[" farm\_work(bot1, new Vec3(-2,-1,-2), 'sow', 'carrot'),\\farm\_work(bot2, new Vec3(-1,-1,1), 'sow', 'carrot')",
]
\end{tcolorbox}
\end{tcolorbox}
\caption{Object state recognition failure Example 5.}
\label{fig:e5}
\end{figure*}

\begin{figure*}
    \centering
        \includegraphics[width=1.0\linewidth]{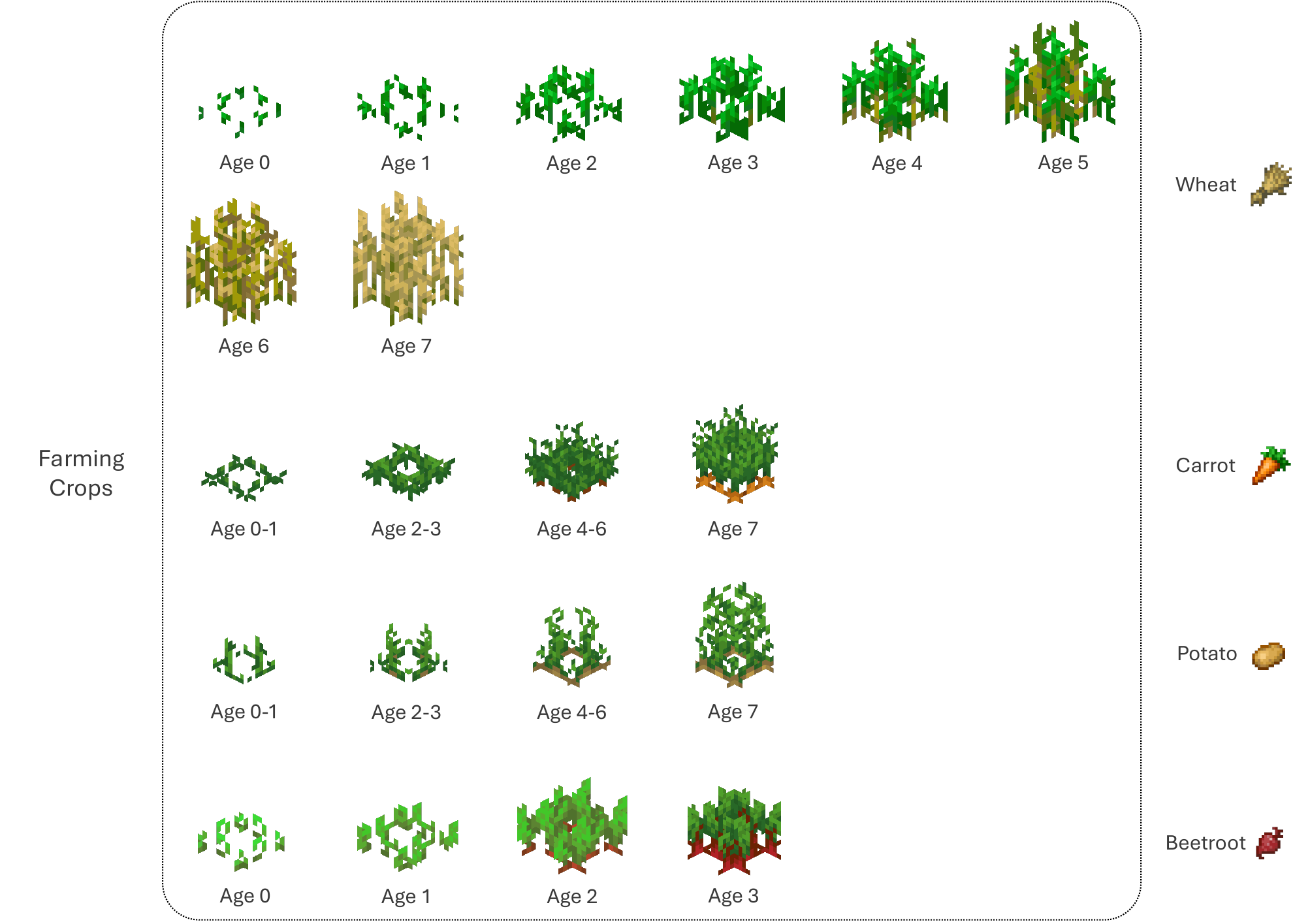}
        \caption{A close-up view of crops appearances across various growing stages in farming tasks.}
        \label{fig:diversity_4}
\end{figure*}

\paragraph{Object State Recognition Failure:} As an example (\autoref{fig:e5}), a farming task requires two agents to collect four additional carrots. In Step~0, \textit{bot1} and \textit{bot2} both sow carrots and attempt to harvest them in Step~2. However, at that time, the carrots are still immature and not ready for collection. The mature state of the carrot is shown in \autoref{fig:diversity_4}.

\section{GPT-4o Implementation}
\label{appendix:gpt4o}

We use \verb|gpt-4o-2024-08-06| as the proprietary VLA. Specifically, we use similar prompt structures as the centralized finetuned TeamCraft-VLA model, with additional task information in the initial system prompt as shown in \autoref{fig:gpt4o-obs}, \autoref{fig:gpt4o-obs3} and \autoref{fig:gpt4o-obs5} to provide background knowledge of the task. The system prompt contains recipes, input, output formats, all available blocks, items, workspace limitations, and one successful rollout of a similar task in the same task family. At the first step, we additionally provide the first user prompt, where the model is given a specific multi-modal task specification accompanied by initial visual observations and inventory details of the agents. Based on the system prompts and user prompts, the model predicts the actions. As the interaction progresses with subsequent prompts, the context is maintained and expanded with the addition of prior responses and updated visual data as shown in \autoref{fig:gpt4o-obs2}.

\section{GPT-4o Error Analysis}
\begin{figure*}[!ht]
    \centering
\begin{tcolorbox}[colback=gray!5!white,colframe=gray!75!black]
\textbf{System Prompt}\newline
You are controlling 3 bots in a Minecraft world. The goal is to build a specific structure on a platform.
\newline
Please review the images provided below, which include the current state of the world and the goal structure (the final image is the three orthographic views of the goal). Based on these observations, generate actions for each bot to help build the structure.
\newline
**Instructions:**
\newline
- **Action Format:**
\newline
- **Bots:**
\newline
- `botID` can be one of: 'bot1', 'bot2', 'bot3', 'bot4' (depending on the number of bots).
\newline
- **Blocks:**
\newline
- `"block"` is the type of block to place.
\newline
- **Available Blocks:**
\newline
- `oak fence", `birch log", `coal ore", `bricks", `sandstone", `stone", `iron ore", `gold ore", `sponge", `sea lantern", dirt, `grass block", `clay", `oak planks", `emerald block", `pumpkin", `orange concrete", `purple wool", `end stone", `bookshelf", `acacia fence", `oak log".
\newline
- **Constraints:**
\newline
- **Inventory Awareness:** Ensure each bot has the necessary blocks in their inventory.
\newline
- **No Overlapping Blocks:** Do not place more than one block at the same position.
\newline
- **Workspace Dimensions:** The center of the workspace is at (0, 0, 0), and it spans 3 units along the x-axis, 3 units along the z-axis, and 2 units along the y-axis.
\newline
- **One Action per Bot:** Each bot can place only one block at a time.
\newline
**Submission Guidelines:**
\newline
- Provide only the list of action commands for all bots.
\newline
- Do not include any additional text, explanations, or formatting (e.g., no code blocks or markdown).
\newline
- Example:
\newline
[ "placeItem(bot1, 'stone', new Vec3(1, 0, 0))", "placeItem(bot2, 'oak planks', new Vec3(0, 0, 1))" ]
  \newline      
You need to put "" for each entry in the list.
\newline
Please generate the list of commands based on the current observations and the goal image.
\end{tcolorbox}
\begin{tcolorbox}[colback=gray!5!white,colframe=gray!75!black]
\textbf{User Prompt}\newline
Here are the observations at the current time step for each agent. The final image is the goal image.
bot1 has 5 dirt. bot1 has 3 clay. bot1 has 7 emerald block. bot1 has 1 oak fence. bot1 has 3 sponge. bot1 has 1 bricks. bot1 has 3 sea lantern. bot2 has 5 bricks. bot2 has 2 sponge. bot2 has 6 sea lantern. bot2 has 2 oak fence. bot2 has 4 emerald block. bot2 has 1 dirt. bot2 has 3 clay. bot3 has 6 emerald block. bot3 has 4 oak fence. bot3 has 2 dirt. bot3 has 2 sponge. bot3 has 3 clay. bot3 has 2 sea lantern.\newline
    \begin{subfigure}[b]{0.23\textwidth}
        \includegraphics[width=\textwidth]{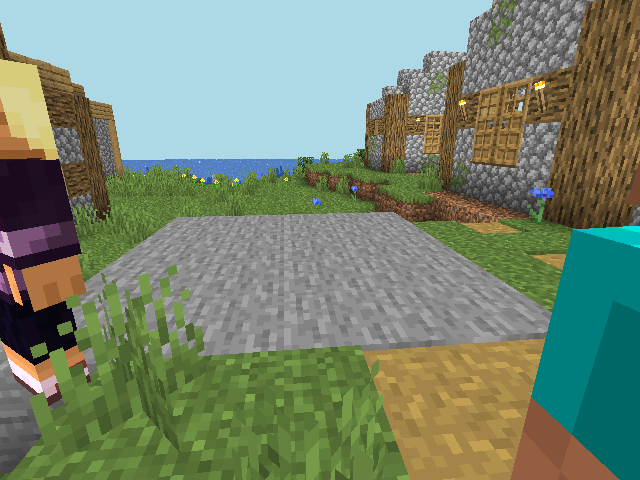}
        \caption{Agent1 observation}
    \end{subfigure}
    \hfill
    \begin{subfigure}[b]{0.23\textwidth}
        \includegraphics[width=\textwidth]{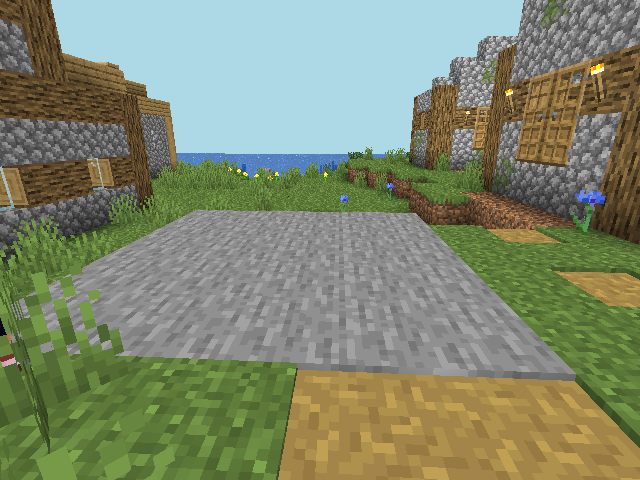}
        \caption{Agent2 observation}
    \end{subfigure}
    \hfill
    \begin{subfigure}[b]{0.23\textwidth}
        \includegraphics[width=\textwidth]{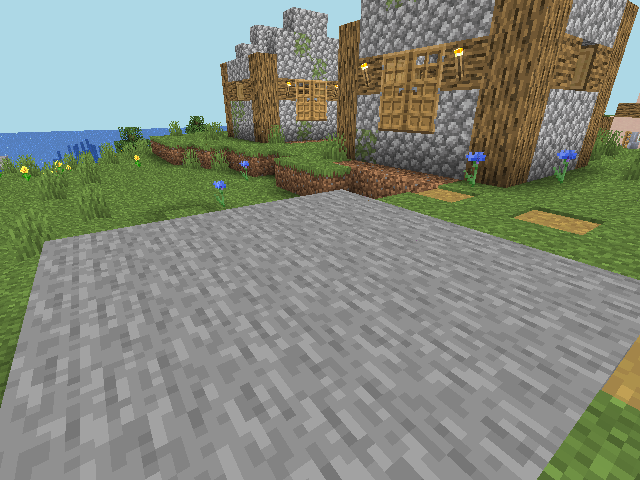}
        \caption{Agent3 observation}
    \end{subfigure}
    \hfill
    \begin{subfigure}[b]{0.23\textwidth}
        \includegraphics[width=\textwidth]{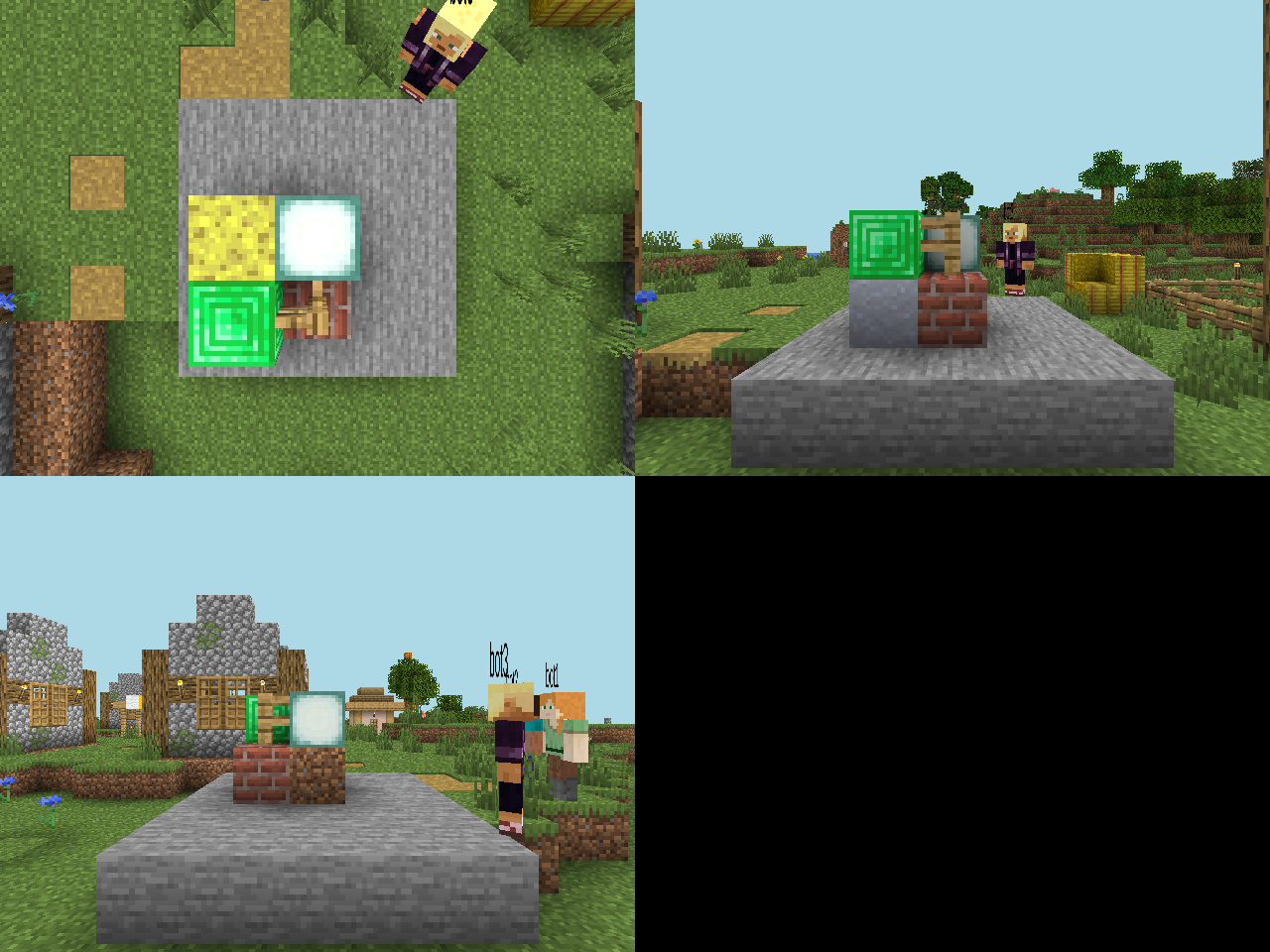}
        \caption{orthographic views of the goal}
    \end{subfigure}  
\end{tcolorbox}
\begin{tcolorbox}[colback=gray!5!white,colframe=gray!75!black]
\textbf{Assistant}\newline
[placeItem(bot1, emerald block, new Vec3(-1,0,-1)), placeItem(bot2, bricks, new Vec3(0,0,0)), \\ placeItem(bot3, sea lantern, new Vec3(-1,0,0))]
\end{tcolorbox}
\caption{Sample rollouts of GPT-4o in a building task, Part~I.}
\label{fig:gpt4o-obs}
\end{figure*}

\begin{figure*}
    \begin{tcolorbox}[colback=gray!5!white,colframe=gray!75!black]
\textbf{User Prompt}\newline
Here are the observations at the current time step for each agent. The final image is the goal image. bot1 has ...\newline
    \begin{subfigure}[b]{0.23\textwidth}
        \includegraphics[width=\textwidth]{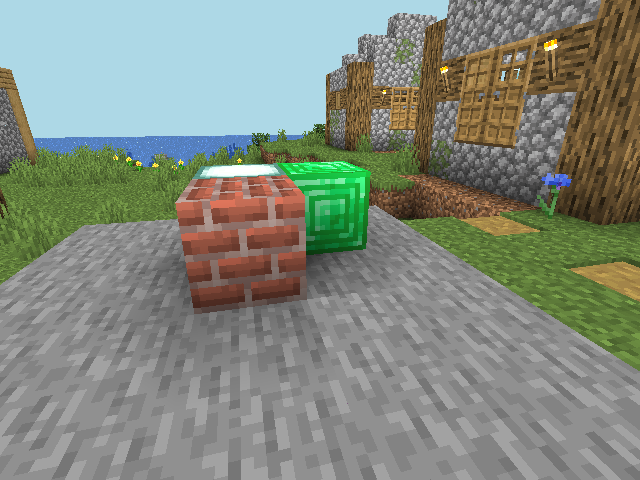}
        \caption{Agent1 observation}
    \end{subfigure}
    \hfill
    \begin{subfigure}[b]{0.23\textwidth}
        \includegraphics[width=\textwidth]{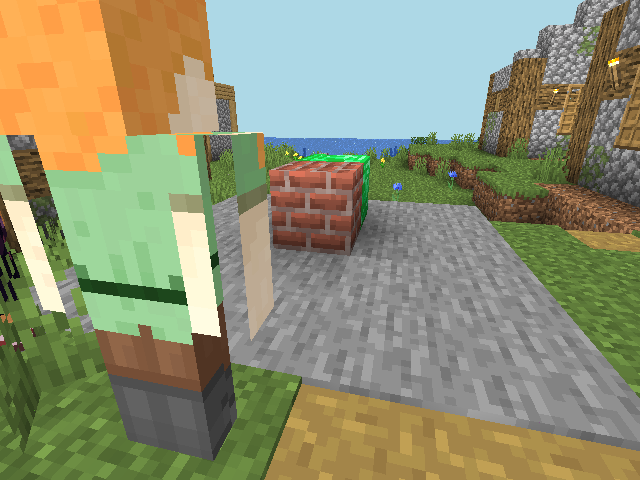}
        \caption{Agent2 observation}
    \end{subfigure}
    \hfill
    \begin{subfigure}[b]{0.23\textwidth}
        \includegraphics[width=\textwidth]{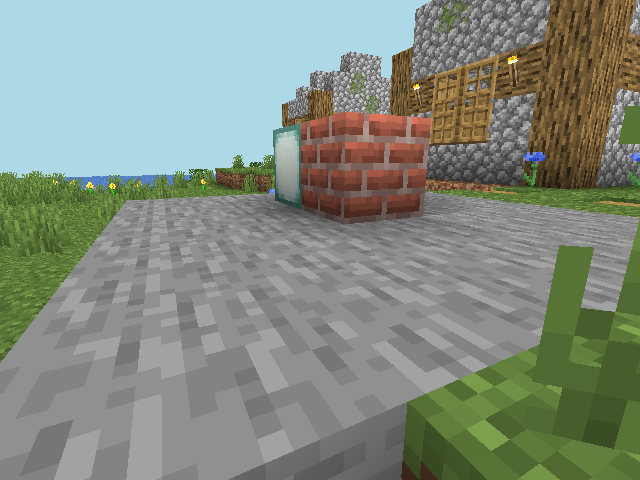}
        \caption{Agent3 observation}
    \end{subfigure}
    \hfill
    \begin{subfigure}[b]{0.23\textwidth}
        \includegraphics[width=\textwidth]{gpt4figures/obs.png}
        \caption{orthographic views of the goal}
    \end{subfigure}
\end{tcolorbox}
\begin{tcolorbox}[colback=gray!5!white,colframe=gray!75!black]
\textbf{Assistant}\newline
[placeItem(bot1, sponge, new Vec3(1,0,-1)), placeItem(bot2, oak fence, new Vec3(0,0,1))]
\end{tcolorbox}
 \begin{tcolorbox}[colback=gray!5!white,colframe=gray!75!black]
\textbf{User Prompt}\newline
Here are the observations at the current time step for each agent. The final image is the goal image. bot1 has ...\newline
    \begin{subfigure}[b]{0.23\textwidth}
        \includegraphics[width=\textwidth]{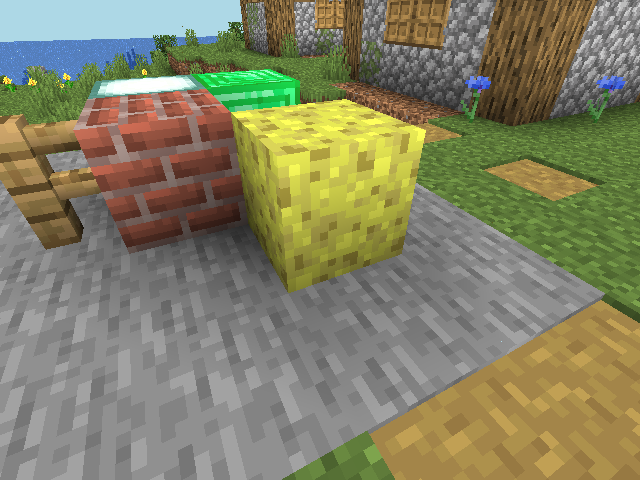}
        \caption{Agent1 observation}
    \end{subfigure}
    \hfill
    \begin{subfigure}[b]{0.23\textwidth}
        \includegraphics[width=\textwidth]{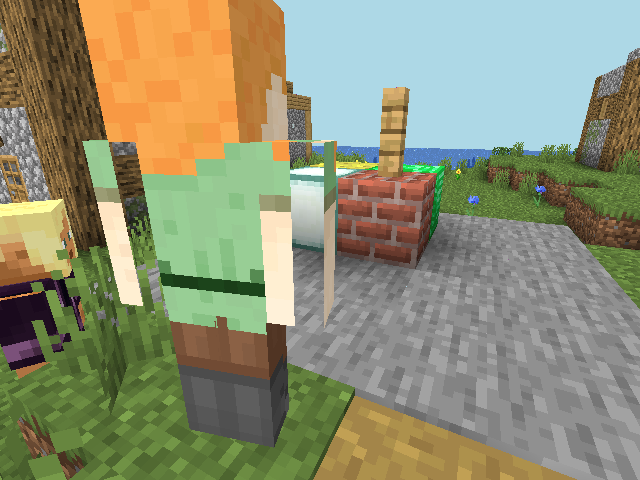}
        \caption{Agent2 observation}
    \end{subfigure}
    \hfill
    \begin{subfigure}[b]{0.23\textwidth}
        \includegraphics[width=\textwidth]{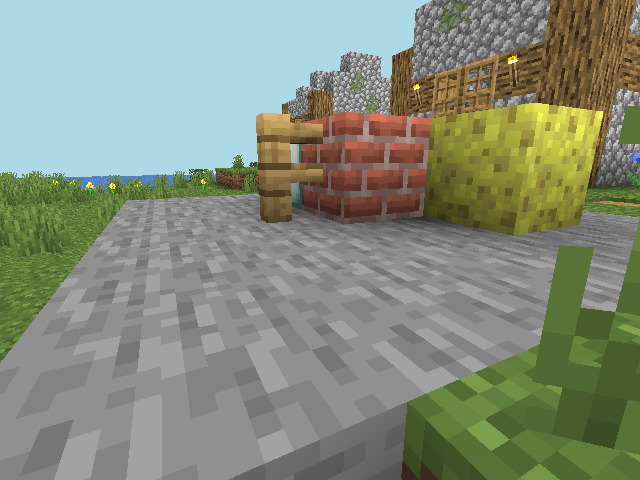}
        \caption{Agent3 observation}
    \end{subfigure}
    \hfill
    \begin{subfigure}[b]{0.23\textwidth}
        \includegraphics[width=\textwidth]{gpt4figures/obs.png}
        \caption{orthographic views of the goal}
    \end{subfigure}
\end{tcolorbox}
\begin{tcolorbox}[colback=gray!5!white,colframe=gray!75!black]
\textbf{Assistant}\newline
[placeItem(bot3,  sea lantern , new Vec3(0,0,-1))]
\end{tcolorbox}
\begin{tcolorbox}[colback=gray!5!white,colframe=gray!75!black]
\textbf{Comment}\newline
GPT-4o failed to place the blocks in the correct location. The agents placed an additional fence that is not required by the goal. The sponge (yellow) block should be placed next to a sea lattern (white) block. 
\end{tcolorbox}
\caption{Sample rollouts of GPT-4o in a building task, Part~II.}
\label{fig:gpt4o-obs2}
\end{figure*}

\begin{figure*}[!ht]
    \centering
\begin{tcolorbox}[colback=gray!5!white,colframe=gray!75!black]
\textbf{System Prompt}\newline
You are controlling 3 bots in a Minecraft world. The goal is to break all blocks in the world.
\newline
    Please review the images provided below, which include the current state of the world. Based on these observations, generate actions for each bot to help break the required blocks.
\newline
    **Instructions:**
\newline
    - **Action Format:**
        - `mineBlock(botID, new Vec3(x,y,z))` to break a block at the specified coordinates.
\newline      
        - The up axis is the y-axis. x, y, z should be greater than 1.
\newline
    - **Bots:**
        - `botID` can be one of: 'bot1', 'bot2', 'bot3' (depending on the number of bots).
\newline
    - **Constraints:**
\newline    
        - **Tool Awareness:** Each bot has tools in their inventory to break blocks. Use the `item dict durability` below to determine how many actions are required based on the tool and block type.
\newline           
            - `item dict durability`: {item dict durbility}
\newline           
            - For example, if a bot has a stone axe, it can break a 'bookshelf' in 1 action; otherwise, it will take 2 actions.
\newline          
        - **No Duplicate Actions:** Ensure that no bot attempts to break the same block more than once.
\newline        
        - **Workspace Dimensions:** The center of the workspace is at (0, 0, 0), and it spans 5 units along the x-axis (-2, 2), 2 units along the y-axis, and 5 units along the z-axis (-2, 2). So the action can have negative numbers
 \newline       
        - **One Action per Bot:** Each bot can attempt to break only one block at a time unless specified by `item dict durability`.
\newline
    **Submission Guidelines:**
\newline
    - Provide only the list of action commands for all bots.
\newline  
    - Do not include any additional text, explanations, or formatting (e.g., no code blocks or markdown).
\newline  
    - Example:
    [ "mineBlock(bot1, new Vec3(1,0,-1))", "mineBlock(bot2, new Vec3(-1,0,1))" ]
\newline
    Please generate the list of commands based on the current observations, the goal image, and tool durability requirements.
Please generate the list of commands based on the current observations and the goal image.
\end{tcolorbox}
\begin{tcolorbox}[colback=gray!5!white,colframe=gray!75!black]
\textbf{User Prompt}\newline
Here are the observations at the current time step for each agent.\newline
    \begin{subfigure}[b]{0.23\textwidth}
        \includegraphics[width=\textwidth]{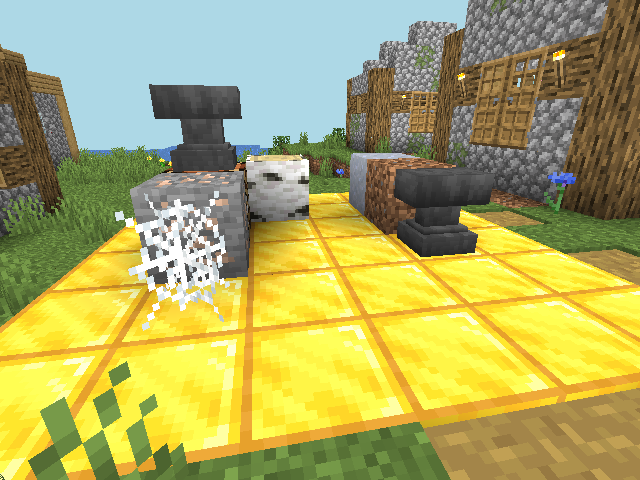}
        \caption{Agent1 observation}
    \end{subfigure}
    \hfill
    \begin{subfigure}[b]{0.23\textwidth}
        \includegraphics[width=\textwidth]{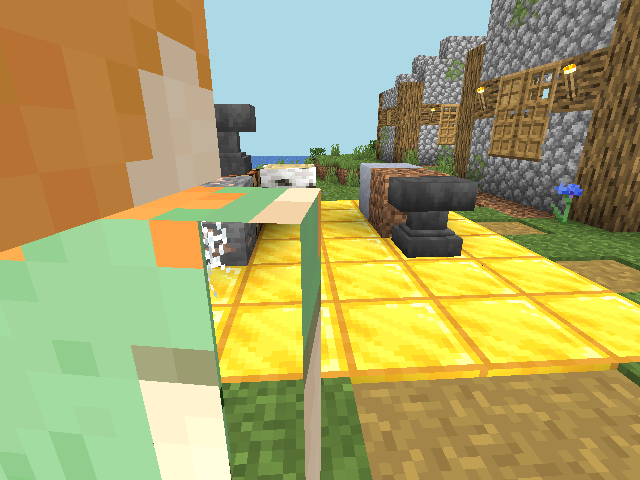}
        \caption{Agent2 observation}
    \end{subfigure}
    \hfill
    \begin{subfigure}[b]{0.23\textwidth}
        \includegraphics[width=\textwidth]{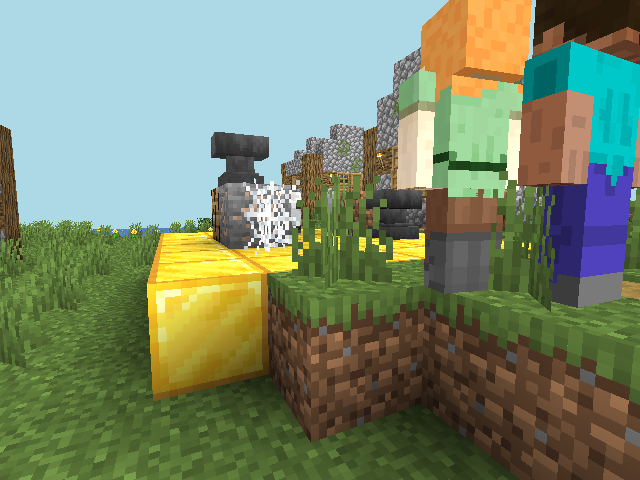}
        \caption{Agent3 observation}
    \end{subfigure}
    \hfill
    \begin{subfigure}[b]{0.23\textwidth}
        \includegraphics[width=\textwidth]{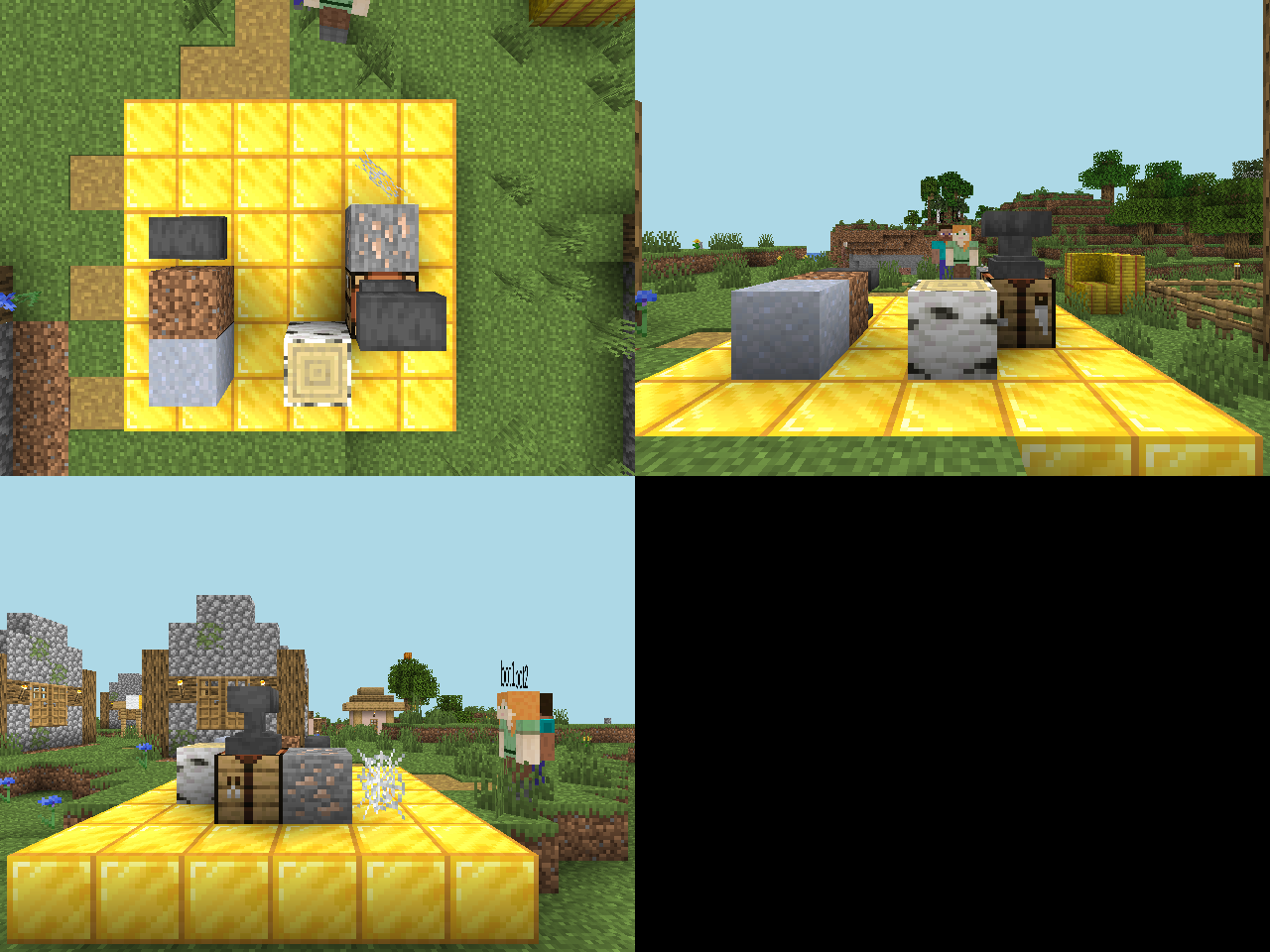}
        \caption{orthographic views }
    \end{subfigure}
\end{tcolorbox}
\begin{tcolorbox}[colback=gray!5!white,colframe=gray!75!black]
\textbf{Assistant}\newline
[mineBlock(bot1, new Vec3(3,0,2)), mineBlock(bot2, new Vec3(2,0,1))]
\end{tcolorbox}
\caption{Sample rollouts of GPT-4o in a clearing task, Part~I.}
\label{fig:gpt4o-obs3}
\end{figure*}

\begin{figure*}
    \begin{tcolorbox}[colback=gray!5!white,colframe=gray!75!black]
\textbf{User Prompt}\newline
Here are the observations at the current time step for each agent. The final image is the goal image. bot1 has ...\newline
    \begin{subfigure}[b]{0.23\textwidth}
        \includegraphics[width=\textwidth]{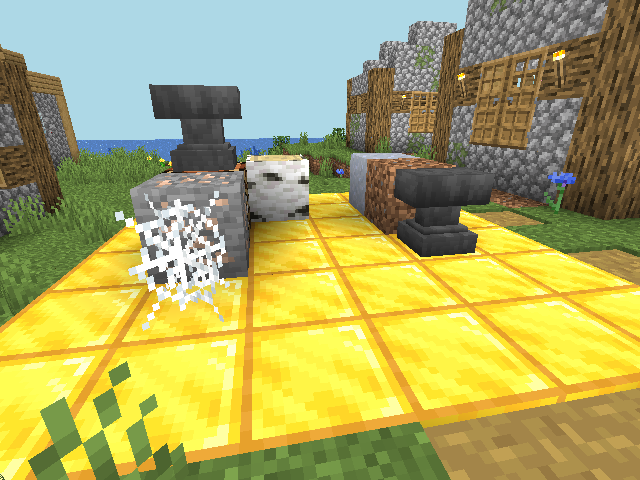}
        \caption{Agent1 observation}
    \end{subfigure}
    \hfill
    \begin{subfigure}[b]{0.23\textwidth}
        \includegraphics[width=\textwidth]{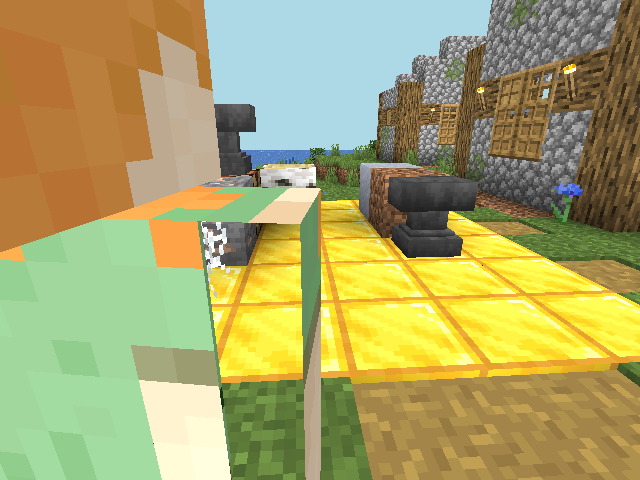}
        \caption{Agent2 observation}
    \end{subfigure}
    \hfill
    \begin{subfigure}[b]{0.23\textwidth}
        \includegraphics[width=\textwidth]{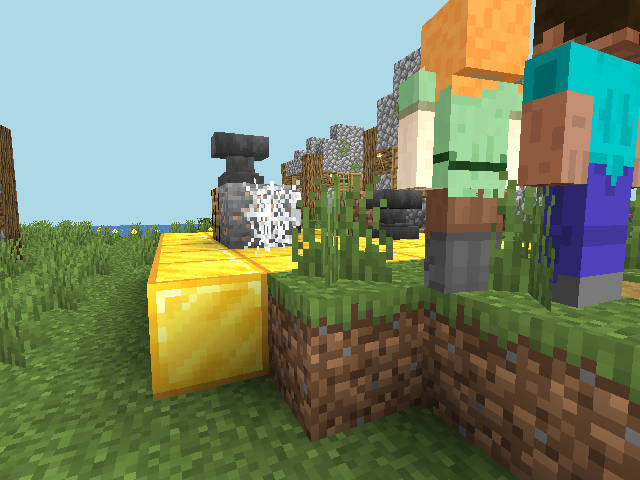}
        \caption{Agent3 observation}
    \end{subfigure}
    \hfill
    \begin{subfigure}[b]{0.23\textwidth}
        \includegraphics[width=\textwidth]{task_break_figures/obs.png}
        \caption{orthographic views }
    \end{subfigure}
\end{tcolorbox}
\begin{tcolorbox}[colback=gray!5!white,colframe=gray!75!black]
\textbf{Assistant}\newline
[mineBlock(bot1, new Vec3(3,0,2)), mineBlock(bot2, new Vec3(2,0,1))]
\end{tcolorbox}
\begin{tcolorbox}[colback=gray!5!white,colframe=gray!75!black]
\textbf{User Prompt}\newline
Here are the observations at the current time step for each agent. The final image is the goal image. bot1 has ...\newline
    \begin{subfigure}[b]{0.23\textwidth}
        \includegraphics[width=\textwidth]{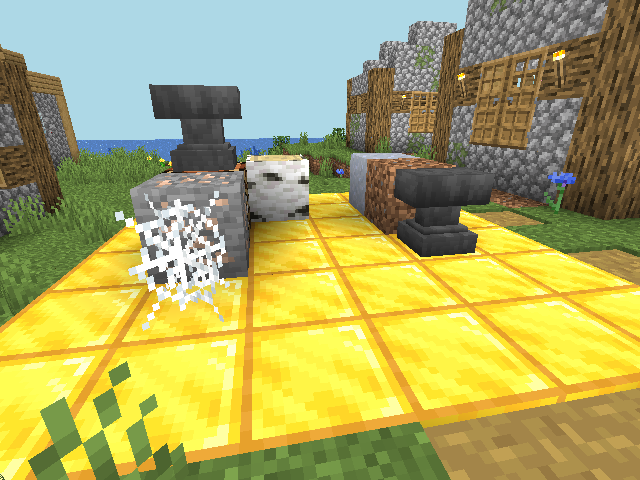}
        \caption{Agent1 observation}
    \end{subfigure}
    \hfill
    \begin{subfigure}[b]{0.23\textwidth}
        \includegraphics[width=\textwidth]{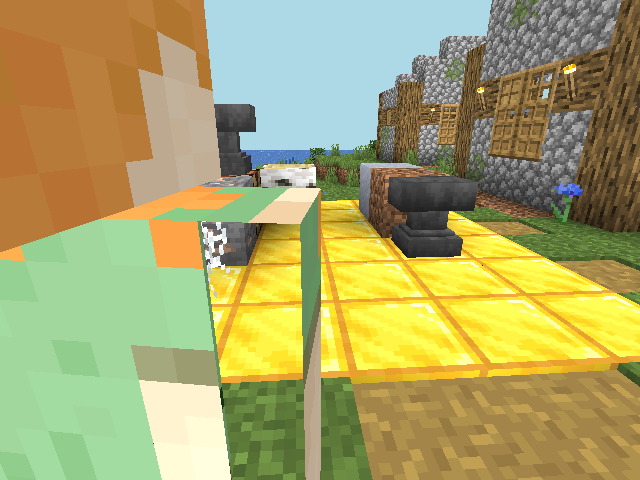}
        \caption{Agent2 observation}
    \end{subfigure}
    \hfill
    \begin{subfigure}[b]{0.23\textwidth}
        \includegraphics[width=\textwidth]{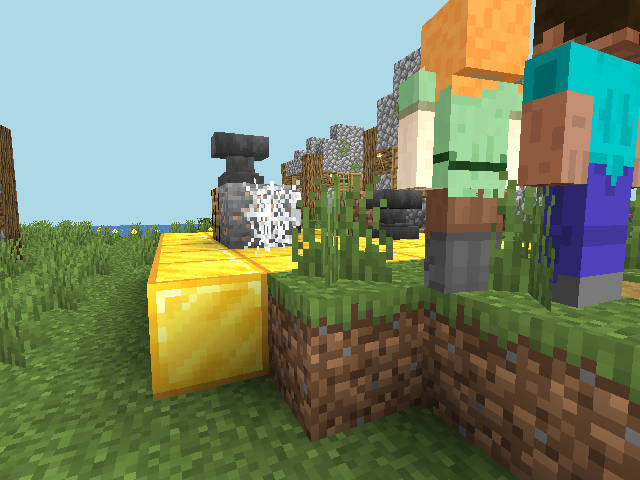}
        \caption{Agent3 observation}
    \end{subfigure}
    \hfill
    \begin{subfigure}[b]{0.23\textwidth}
        \includegraphics[width=\textwidth]{task_break_figures/obs.png}
        \caption{orthographic views}
    \end{subfigure}
\end{tcolorbox}
\begin{tcolorbox}[colback=gray!5!white,colframe=gray!75!black]
\textbf{Assistant}\newline
[mineBlock(bot1, new Vec3(1,0,2)), mineBlock(bot2, new Vec3(2,0,1))]
\end{tcolorbox}
\begin{tcolorbox}[colback=gray!5!white,colframe=gray!75!black]
\textbf{Comment}\newline
The model did not predict correct block locations and failed to mine any blocks. 
\end{tcolorbox}
\caption{Sample rollouts of GPT-4o in a clearing task, Part~II.}
\label{fig:gpt4o-obs4}
\end{figure*}

\begin{figure*}[!ht]
    \centering
\begin{tcolorbox}[colback=gray!5!white,colframe=gray!75!black]
\textbf{System Prompt}\newline
You are controlling 3 bots in a Minecraft farming environment. The bots are responsible for planting, growing, and harvesting crops like wheat, carrots, and potatoes in a designated farming area.
\newline
        Please review the images provided below, which include the current state of the farm and the goal farm layout (the final image is the 3-view of the goal). Based on these observations, generate farming commands for each bot to help manage the farming tasks efficiently.
\newline
        **Instructions:**
\newline
        1. **Action Format:**
        - The actions are in the format:
            farmwork(botID, new Vec3(x, y, z), 'action', 'crop')
            - `botID` can be one of: 'bot1', 'bot2', or 'bot3'.
\newline            
            - `'action'` is the type of action the bot will perform, such as `'sow'` (for planting) or `'harvest'`.
\newline            
            - `'crop'` is the type of crop the bot will interact with, such as `'wheat'`, `'carrot'`, `'potato', or 'beetroots'.
\newline
        2. **Bots:**
        - Each bot has its own inventory and task assignment for sowing seeds, watering, or harvesting the crops.
\newline        
        3. **Crops:**
        - The available crops are wheat, carrots, and potatoes. Ensure each bot is assigned to the correct action and crop.
\newline        
        4. **Commands:**
        - Example commands:
        ["farmwork(bot1, new Vec3(0, 0, -1), 'sow', 'carrot')", "farmwork(bot2, new Vec3(1, 0, -1), 'harvest')"]
\newline
        **Constraints:**
        - **Inventory Awareness:** Ensure each bot has the necessary items (seeds or tools) in their inventory.
\newline        
        - **No Overlapping Actions:** Do not assign multiple bots to perform the same action on the same crop plot simultaneously.
\newline        
        - **Workspace Dimensions:** The center of the workspace is at (0, 0, 0), and it spans 5 units along the x-axis (-3,3), 5 units along the z-axis(-3,3), and 3 units along the y-axis.
\newline       
        - **One Action per Bot:** Each bot can perform only one action at a time.
\newline
        **Submission Guidelines:**
\newline
        - Provide only the list of farming commands for all bots.
\newline        
        - Do not include any additional text, explanations, or formatting (e.g., no code blocks or markdown).
 \newline       
        - Example:
        [ "farmwork(bot1, new Vec3(1, 0, 0), 'sow', 'wheat')", "farmwork(bot2, new Vec3(0, 0, 1), 'harvest')" ]
        Please generate the list of commands based on the current observations and the goal farm layout.
\newline        
        *** You need to harvest when the crop is ready to finish the task ***
\end{tcolorbox}
\begin{tcolorbox}[colback=gray!5!white,colframe=gray!75!black]
\textbf{User Prompt}\newline
Here are the observations at the current time step for each agent.         Please harvest the crops when they are ready! You can only harvest what you sow. You can only sow seeds on empty plots, and you can only sow what you have in your inventory. When harvesting, do not specify the crop type. Here is the inventory status for each bot: bot1 has 7 xxx\newline
    \begin{subfigure}[b]{0.23\textwidth}
        \includegraphics[width=\textwidth]{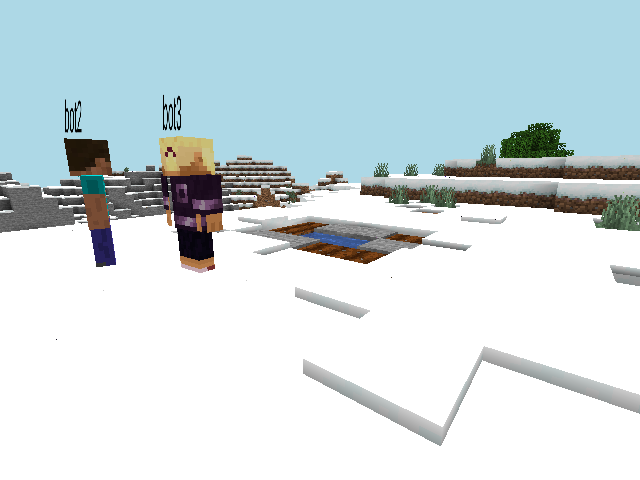}
        \caption{Agent1 observation}
    \end{subfigure}
    \hfill
    \begin{subfigure}[b]{0.23\textwidth}
        \includegraphics[width=\textwidth]{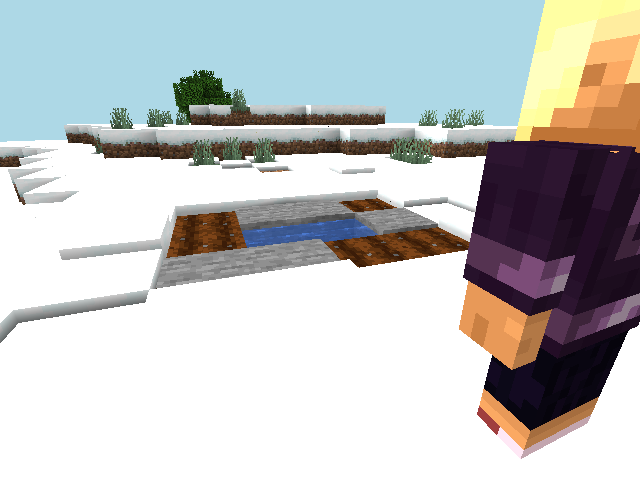}
        \caption{Agent2 observation}
    \end{subfigure}
    \hfill
    \begin{subfigure}[b]{0.23\textwidth}
        \includegraphics[width=\textwidth]{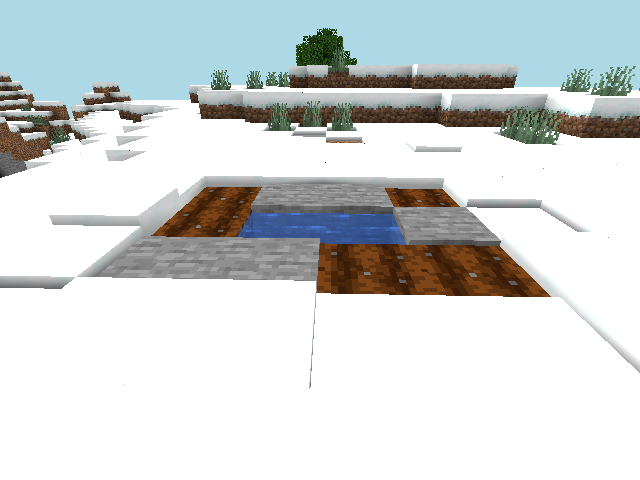}
        \caption{Agent3 observation}
    \end{subfigure}
    \hfill
    \begin{subfigure}[b]{0.23\textwidth}
        \includegraphics[width=\textwidth]{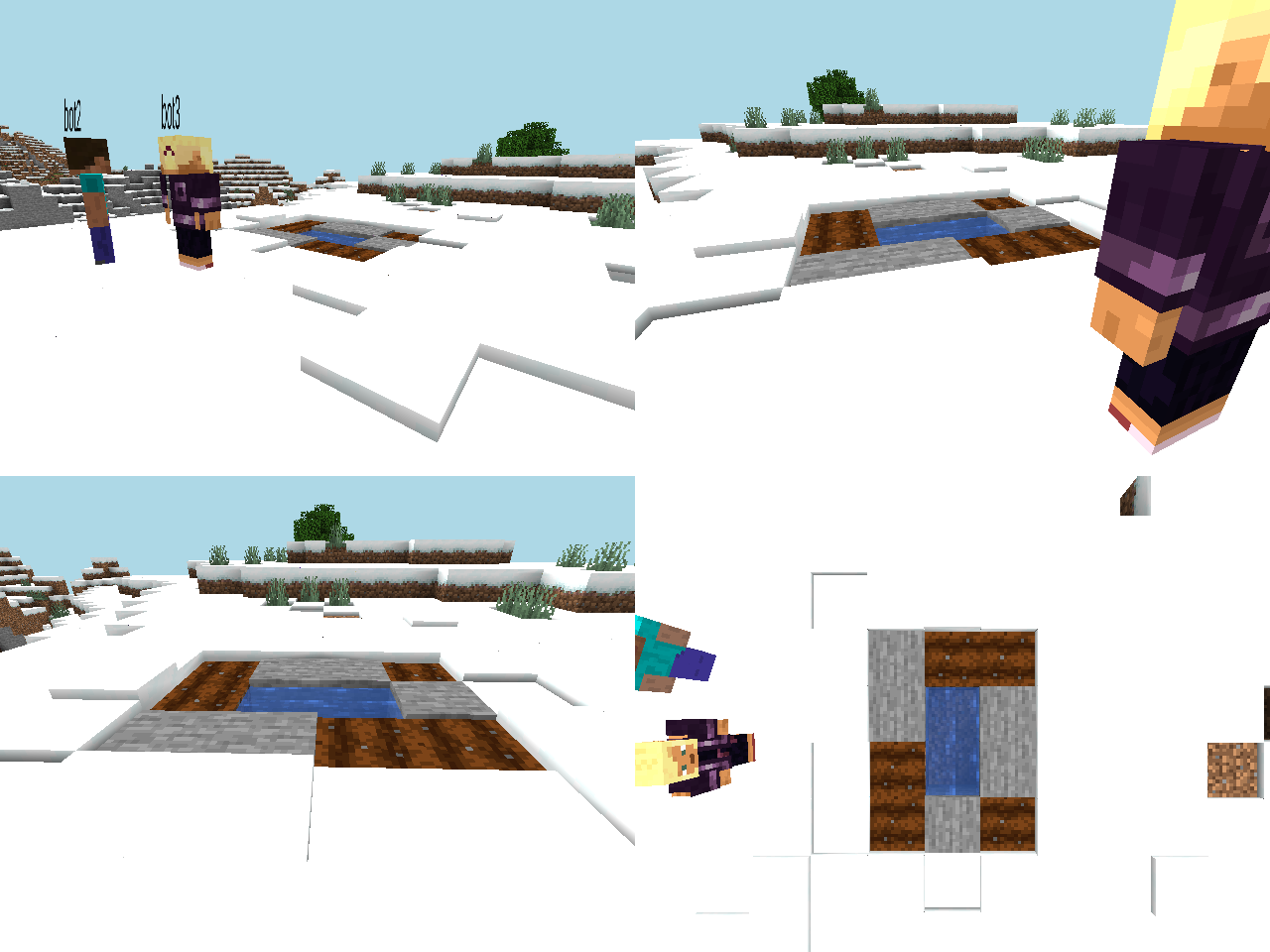}
        \caption{orthographic views }
    \end{subfigure}  
\end{tcolorbox}
\begin{tcolorbox}[colback=gray!5!white,colframe=gray!75!black]
\textbf{Assistant}\newline
farmwork(bot1, new Vec3(0, 0, -1), 'sow', 'potato')\\
farmwork(bot2, new Vec3(1, 0, 0), 'sow', 'potato')\\
farmwork(bot3, new Vec3(-1, 0, 1), 'sow', 'potato')
\end{tcolorbox}
\caption{Sample rollouts of GPT-4o in a farming task, Part~I.}
\label{fig:gpt4o-obs5}
\end{figure*}

\begin{figure*}
    \begin{tcolorbox}[colback=gray!5!white,colframe=gray!75!black]
\textbf{User Prompt}\newline
Here are the observations at the current time step for each agent. bot1 has ...\newline
    \begin{subfigure}[b]{0.23\textwidth}
        \includegraphics[width=\textwidth]{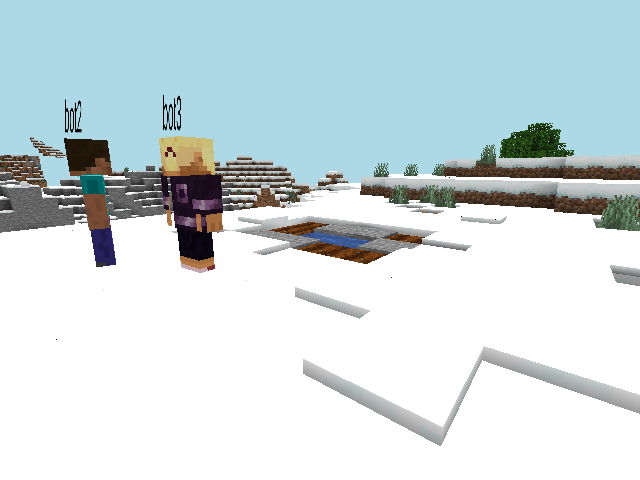}
        \caption{Agent1 observation}
    \end{subfigure}
    \hfill
    \begin{subfigure}[b]{0.23\textwidth}
        \includegraphics[width=\textwidth]{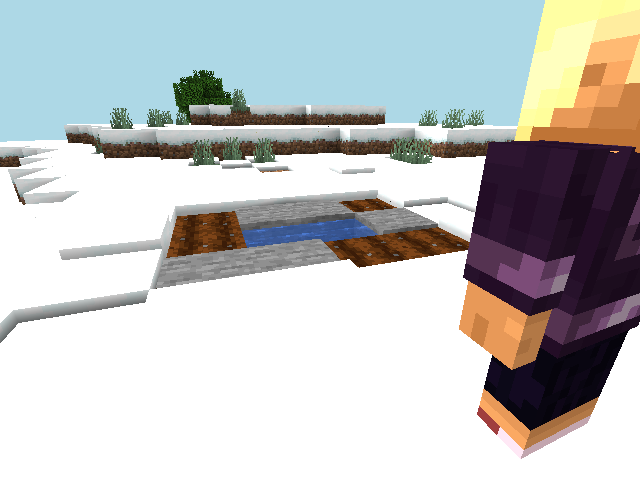}
        \caption{Agent2 observation}
    \end{subfigure}
    \hfill
    \begin{subfigure}[b]{0.23\textwidth}
        \includegraphics[width=\textwidth]{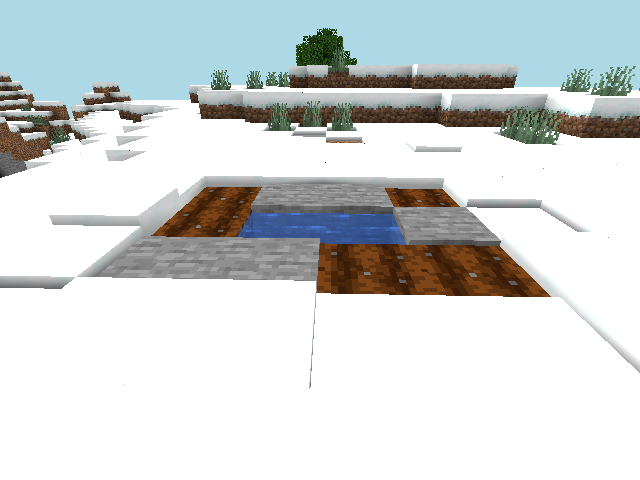}
        \caption{Agent3 observation}
    \end{subfigure}
    \hfill
    \begin{subfigure}[b]{0.23\textwidth}
        \includegraphics[width=\textwidth]{task_farm_figures/obs.png}
        \caption{orthographic views }
    \end{subfigure}
\end{tcolorbox}
\begin{tcolorbox}[colback=gray!5!white,colframe=gray!75!black]
\textbf{Assistant}\newline
farmwork(bot1, new Vec3(0, 0, 1), 'sow', 'potato'),\\
farmwork(bot2, new Vec3(0, 0, -1), 'sow', 'potato'),\\
farmwork(bot3, new Vec3(1, 0, 0), 'sow', 'potato')
\end{tcolorbox}
\begin{tcolorbox}[colback=gray!5!white,colframe=gray!75!black]
\textbf{User Prompt}\newline
Here are the observations at the current time step for each agent.  bot1 has ...\newline
    \begin{subfigure}[b]{0.23\textwidth}
        \includegraphics[width=\textwidth]{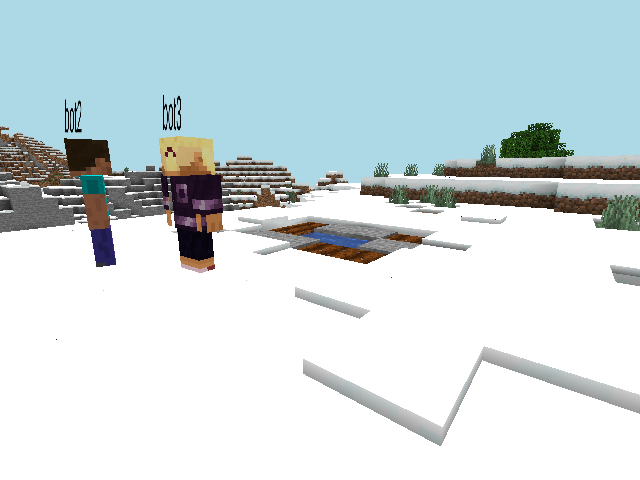}
        \caption{Agent1 observation}
    \end{subfigure}
    \hfill
    \begin{subfigure}[b]{0.23\textwidth}
        \includegraphics[width=\textwidth]{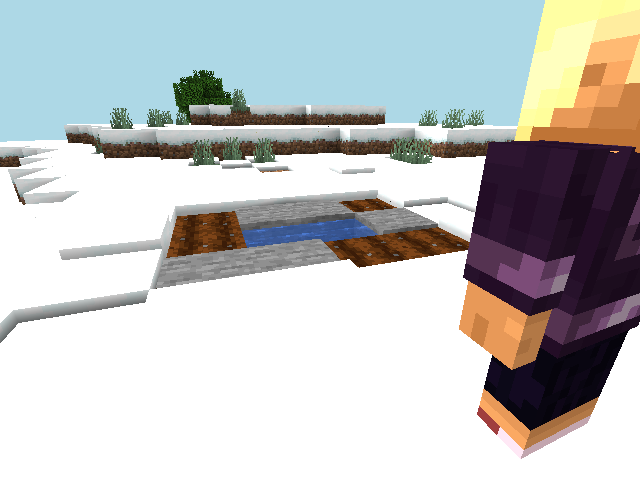}
        \caption{Agent2 observation}
    \end{subfigure}
    \hfill
    \begin{subfigure}[b]{0.23\textwidth}
        \includegraphics[width=\textwidth]{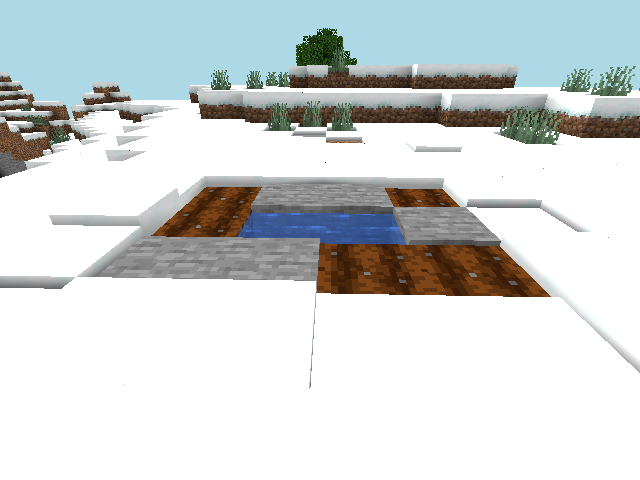}
        \caption{Agent3 observation}
    \end{subfigure}
    \hfill
    \begin{subfigure}[b]{0.23\textwidth}
        \includegraphics[width=\textwidth]{task_farm_figures/obs.png}
        \caption{orthographic views}
    \end{subfigure}
\end{tcolorbox}
\begin{tcolorbox}[colback=gray!5!white,colframe=gray!75!black]
\textbf{Assistant}\newline
farmwork(bot1, new Vec3(-1, 0, 0), 'sow', 'potato'),\\
farmwork(bot2, new Vec3(1, 0, 0), 'sow', 'potato'),\\
farmwork(bot3, new Vec3(0, 0, 1), 'sow', 'potato')
\end{tcolorbox}
\begin{tcolorbox}[colback=gray!5!white,colframe=gray!75!black]
\textbf{Comment}\newline
The model does not follow instruction, and keeps predicting the sowing actions without harvesting.
\end{tcolorbox}
\caption{Sample rollouts of GPT-4o in a farming task, Part~II.}
\label{fig:gpt4o-obs6}
\end{figure*}

We visualized some failure cases in \autoref{fig:gpt4o-obs}, \autoref{fig:gpt4o-obs2}, \autoref{fig:gpt4o-obs3}, \autoref{fig:gpt4o-obs4}, \autoref{fig:gpt4o-obs5} and \autoref{fig:gpt4o-obs6}. In general, GPT4-o fail to understand spatial relations and often chooses the incorrect coordinates as the locations for placing and mining actions. In addition, the model sometimes fails to follow instructions and does not harvest the crop in the farming task, as shown in \autoref{fig:gpt4o-obs5} and \autoref{fig:gpt4o-obs6}.

\clearpage

\clearpage

\clearpage

\clearpage

\clearpage

\clearpage

\clearpage

\section{Dataset Statistics Tables}
\label{appendix:stat_table}

\begin{table}[ht]
\scriptsize
\begin{minipage}[t]{\linewidth}
\centering
\caption{Building Task Diversity Statistics}
\label{tab:statistics_building}
\begin{tabular}{llrr}
\toprule
\textbf{Diversity} & \textbf{Type} & \textbf{Count} & \textbf{Percentage} \\
\midrule
\multicolumn{4}{l}{\textbf{Action Sequences}} \\
 & 3 & 7,777 & 51.85\% \\
 & 2 & 3,207 & 21.38\% \\
 & 4 & 3,091 & 20.61\% \\
 & 5 & 483 & 3.22\% \\
 & 6 & 440 & 2.93\% \\
\midrule
\multicolumn{4}{l}{\textbf{Agents}} \\
 & 3 & 7,505 & 50.03\% \\
 & 2 & 7,493 & 49.97\% \\
\midrule
\multicolumn{4}{l}{\textbf{Scenes}} \\
 & ice\_on\_water & 2,555 & 17.04\% \\
 & mountain\_half & 2,553 & 17.03\% \\
 & village & 2,482 & 16.55\% \\
 & desert\_village & 2,480 & 16.53\% \\
 & snow\_mountain & 2,478 & 16.52\% \\
 & swamp & 2,450 & 16.34\% \\
\midrule
\multicolumn{4}{l}{\textbf{Background Types}} \\
 & stone & 1,530 & 10.20\% \\
 & pink\_wool & 1,527 & 10.19\% \\
 & glowstone & 1,522 & 10.15\% \\
 & obsidian & 1,511 & 10.08\% \\
 & glass & 1,509 & 10.07\% \\
 & smooth\_quartz & 1,499 & 10.00\% \\
 & hay\_block & 1,494 & 9.96\% \\
 & gold\_block & 1,473 & 9.82\% \\
 & oak\_wood & 1,471 & 9.81\% \\
 & cyan\_concrete & 1,462 & 9.75\% \\
\midrule
\multicolumn{4}{l}{\textbf{Target Types}} \\
 & bricks & 10,391 & 9.92\% \\
 & sponge & 5,438 & 5.19\% \\
 & coal\_ore & 5,370 & 5.13\% \\
 & grass\_block & 5,327 & 5.09\% \\
 & clay & 5,318 & 5.08\% \\
 & sea\_lantern & 5,296 & 5.06\% \\
 & orange\_concrete & 5,287 & 5.05\% \\
 & pumpkin & 5,269 & 5.03\% \\
 & purple\_wool & 5,257 & 5.02\% \\
 & gold\_ore & 5,247 & 5.01\% \\
 & oak\_fence & 5,234 & 5.00\% \\
 & oak\_planks & 5,216 & 4.98\% \\
 & birch\_log & 5,184 & 4.95\% \\
 & stone & 5,182 & 4.95\% \\
 & sandstone & 5,176 & 4.94\% \\
 & emerald\_block & 5,164 & 4.93\% \\
 & iron\_ore & 5,160 & 4.93\% \\
 & dirt & 5,124 & 4.89\% \\
 & end\_stone & 5,119 & 4.89\% \\
\midrule
\multicolumn{4}{l}{\textbf{Target Counts}} \\
 & 6 & 5,653 & 37.69\% \\
 & 7 & 2,625 & 17.50\% \\
 & 8 & 2,573 & 17.15\% \\
 & 5 & 2,122 & 14.15\% \\
 & 10 & 526 & 3.51\% \\
 & 12 & 515 & 3.43\% \\
 & 9 & 496 & 3.31\% \\
 & 11 & 488 & 3.25\% \\
\midrule
\multicolumn{4}{l}{\textbf{Dimensional Shapes}} \\
 & [3, 1, 2] & 3,859 & 25.73\% \\
 & [4, 1, 2] & 3,770 & 25.14\% \\
 & [2, 3, 2] & 3,695 & 24.63\% \\
 & [2, 2, 2] & 3,674 & 24.49\% \\
\bottomrule
\end{tabular}
\end{minipage}
\end{table}

\begin{table}[b]
\scriptsize
\begin{minipage}[h]{\linewidth}
\vspace{11mm}
\centering
\caption{Clearing Task Diversity Statistics}
\label{tab:statistics_clearing}
\begin{tabular}{llrr}
\toprule
\textbf{Diversity} & \textbf{Type} & \textbf{Count} & \textbf{Percentage} \\
\midrule
\multicolumn{4}{l}{\textbf{Action Sequences}} \\
 & 4 & 4,027 & 27.51\% \\
 & 5 & 3,751 & 25.61\% \\
 & 6 & 3,270 & 22.32\% \\
 & 3 & 1,561 & 10.66\% \\
 & 7 & 1,396 & 9.53\% \\
 & 8 & 424 & 2.89\% \\
 & 9 & 133 & 0.91\% \\
 & 2 & 79 & 0.54\% \\
\midrule
\multicolumn{4}{l}{\textbf{Agents}} \\
 & 2 & 7,358 & 50.28\% \\
 & 3 & 7,283 & 49.72\% \\
\midrule
\multicolumn{4}{l}{\textbf{Scenes}} \\
 & desert\_village & 3,012 & 20.56\% \\
 & snow\_mountain & 2,948 & 20.13\% \\
 & swamp & 2,929 & 20.00\% \\
 & ice\_on\_water & 2,894 & 19.76\% \\
 & village & 2,858 & 19.54\% \\
\midrule
\multicolumn{4}{l}{\textbf{Background Types}} \\
 & smooth\_quartz & 1,405 & 9.59\% \\
 & pink\_wool & 1,357 & 9.27\% \\
 & gold\_block & 1,353 & 9.24\% \\
 & oak\_wood & 1,334 & 9.10\% \\
 & hay\_block & 1,332 & 9.09\% \\
 & cyan\_concrete & 1,332 & 9.09\% \\
 & grass\_block & 1,328 & 9.06\% \\
 & glass & 1,325 & 9.04\% \\
 & glowstone & 1,309 & 8.93\% \\
 & stone & 1,302 & 8.89\% \\
 & obsidian & 1,264 & 8.63\% \\
\midrule
\multicolumn{4}{l}{\textbf{Target Counts}} \\
 & 6 & 4,310 & 29.43\% \\
 & 5 & 2,499 & 17.07\% \\
 & 4 & 2,436 & 16.64\% \\
 & 8 & 1,843 & 12.58\% \\
 & 7 & 1,803 & 12.31\% \\
 & 9 & 1,750 & 11.95\% \\
\midrule
\multicolumn{4}{l}{\textbf{Target Types}} \\
 & oak\_fence & 5,879 & 6.45\% \\
 & grass\_block & 5,836 & 6.40\% \\
 & clay & 5,816 & 6.38\% \\
 & oak\_log & 5,772 & 6.33\% \\
 & sandstone & 5,748 & 6.30\% \\
 & acacia\_fence & 5,744 & 6.30\% \\
 & birch\_log & 5,732 & 6.28\% \\
 & bookshelf & 5,726 & 6.28\% \\
 & stone & 5,709 & 6.26\% \\
 & bricks & 5,695 & 6.25\% \\
 & crafting\_table & 5,684 & 6.23\% \\
 & dirt & 5,671 & 6.22\% \\
 & cobweb & 5,605 & 6.15\% \\
 & iron\_ore & 5,603 & 6.14\% \\
 & coal\_ore & 5,555 & 6.09\% \\
 & anvil & 5,439 & 5.96\% \\
\midrule
\multicolumn{4}{l}{\textbf{Dimensional Shapes}} \\
 & 3 & 7,346 & 50.15\% \\
 & 2 & 7,295 & 49.84\% \\
\midrule
\multicolumn{4}{l}{\textbf{Tools}} \\
 & stone\_pickaxe & 9,329 & 25.51\% \\
 & stone\_sword & 9,180 & 25.10\% \\
 & stone\_axe & 9,150 & 24.99\% \\
 & stone\_shovel & 8,906 & 24.36\% \\
\bottomrule
\end{tabular}
\end{minipage}
\end{table}

\clearpage\newpage

\begin{table}[ht]
\scriptsize
\begin{minipage}[h]{\linewidth}
\centering
\caption{Farming Task Diversity Statistics}
\label{tab:statistics_farming}
\begin{tabular}{llrr}
\toprule
\textbf{Diversity} & \textbf{Type} & \textbf{Count} & \textbf{Percentage} \\
\midrule
\multicolumn{4}{l}{\textbf{Action Sequences}} \\
 & 4 & 7,458 & 50.33\% \\
 & 5 & 3,731 & 25.17\% \\
 & 3 & 3,264 & 22.02\% \\
 & 6 & 270 & 1.82\% \\
 & 2 & 81 & 0.55\% \\
 & 7 & 11 & 0.07\% \\
\midrule
\multicolumn{4}{l}{\textbf{Agents}} \\
 & 2 & 7,465 & 50.37\% \\
 & 3 & 7,350 & 49.63\% \\
\midrule
\multicolumn{4}{l}{\textbf{Scenes}} \\
 & snow\_mountain & 3,732 & 25.18\% \\
 & swamp & 3,722 & 25.11\% \\
 & ice\_on\_water & 3,707 & 25.01\% \\
 & village & 3,654 & 24.69\% \\
\midrule
\multicolumn{4}{l}{\textbf{Background Types}} \\
 & stone & 2,892 & 19.51\% \\
 & obsidian & 1,549 & 10.46\% \\
 & hay\_block & 1,527 & 10.30\% \\
 & oak\_wood & 1,524 & 10.28\% \\
 & cyan\_concrete & 1,492 & 10.06\% \\
 & glass & 1,465 & 9.88\% \\
 & smooth\_quartz & 1,462 & 9.86\% \\
 & pink\_wool & 1,455 & 9.81\% \\
 & dirt & 1,449 & 9.77\% \\
\midrule
\multicolumn{4}{l}{\textbf{Target Types}} \\
 & potato & 4,972 & 33.56\% \\
 & carrot & 4,955 & 33.45\% \\
 & wheat & 4,888 & 32.99\% \\
\midrule
\multicolumn{4}{l}{\textbf{Target Counts}} \\
 & 4 & 2,873 & 19.39\% \\
 & 3 & 2,269 & 15.31\% \\
 & 5 & 2,256 & 15.22\% \\
 & 6 & 2,151 & 14.51\% \\
 & 2 & 1,240 & 8.37\% \\
 & 8 & 1,112 & 7.50\% \\
 & 10 & 1,062 & 7.17\% \\
 & 7 & 933 & 6.29\% \\
 & 12 & 512 & 3.45\% \\
 & 14 & 407 & 2.75\% \\
\bottomrule
\end{tabular}
\end{minipage}
\end{table}

\begin{table}[b]
\scriptsize
\begin{minipage}[t]{\linewidth}
\centering
\caption{Smelting Task Diversity Statistics}
\label{tab:statistics_smelting}
\begin{tabular}{llrr}
\toprule
\textbf{Diversity} & \textbf{Type} & \textbf{Count} & \textbf{Percentage} \\
\midrule
\multicolumn{4}{l}{\textbf{Action Sequences}} \\
 & 5 & 3,261 & 30.20\% \\
 & 4 & 3,072 & 28.45\% \\
 & 6 & 2,041 & 18.89\% \\
 & 3 & 1,824 & 16.88\% \\
 & 2 & 358 & 3.31\% \\
 & 7 & 239 & 2.21\% \\
 & 8 & 8 & 0.07\% \\
\midrule
\multicolumn{4}{l}{\textbf{Agents}} \\
 & 3 & 5,480 & 50.75\% \\
 & 2 & 5,323 & 49.25\% \\
\midrule
\multicolumn{4}{l}{\textbf{Scenes}} \\
 & snow\_mountain & 2,272 & 21.04\% \\
 & desert\_village & 2,257 & 20.92\% \\
 & swamp & 2,171 & 20.08\% \\
 & ice\_on\_water & 2,059 & 19.09\% \\
 & village & 2,044 & 18.87\% \\
\bottomrule
\end{tabular}
\end{minipage}
\end{table}

\begin{table}[b]
\scriptsize
\begin{minipage}[t]{\linewidth}
\centering
\begin{tabular}{llrr}
\multicolumn{4}{l}{\autoref{tab:statistics_smelting}. (cont'd)}\\
\toprule
\textbf{Diversity} & \textbf{Type} & \textbf{Count} & \textbf{Percentage} \\
\midrule
\multicolumn{4}{l}{\textbf{Background Types}} \\
 & gold\_block & 1,014 & 9.22\% \\
 & smooth\_quartz & 1,010 & 9.19\% \\
 & cyan\_concrete & 995 & 9.02\% \\
 & glowstone & 981 & 8.92\% \\
 & pink\_wool & 990 & 8.99\% \\
 & glass & 978 & 8.89\% \\
 & oak\_wood & 987 & 8.98\% \\
 & grass\_block & 977 & 8.88\% \\
 & hay\_block & 968 & 8.80\% \\
 & stone & 964 & 8.76\% \\
 & obsidian & 939 & 8.54\% \\
\midrule
\multicolumn{4}{l}{\textbf{Furnace}} \\
 & 1 & 5,772 & 53.45\% \\
 & 2 & 5,031 & 46.55\% \\
\midrule
\multicolumn{4}{l}{\textbf{Fuel Types}} \\
 & coal\_block & 999 & 9.58\% \\
 & charcoal & 962 & 9.22\% \\
 & lava\_bucket & 940 & 9.01\% \\
 & coal & 921 & 8.84\% \\
 & spruce\_planks & 910 & 8.73\% \\
 & acacia\_planks & 906 & 8.69\% \\
 & oak\_planks & 861 & 8.26\% \\
 & birch\_log & 893 & 8.57\% \\
 & acacia\_log & 887 & 8.50\% \\
 & spruce\_log & 845 & 8.10\% \\
 & oak\_log & 840 & 8.05\% \\
 & birch\_planks & 839 & 8.04\% \\
 \midrule
\multicolumn{4}{l}{\textbf{Goal Types}} \\
 & food & 5,412 & 50.09\% \\
 & item & 5,391 & 49.91\% \\
\midrule
\multicolumn{4}{l}{\textbf{Target Types}} \\
 & glass & 1,144 & 10.26\% \\
 & gold\_ingot & 1,094 & 9.81\% \\
 & stone & 1,077 & 9.66\% \\
 & smooth\_sandstone & 1,040 & 9.32\% \\
 & iron\_ingot & 1,036 & 9.29\% \\
 & cooked\_salmon & 712 & 6.38\% \\
 & cooked\_cod & 708 & 6.35\% \\
 & baked\_potato & 758 & 6.80\% \\
 & cooked\_mutton & 664 & 5.95\% \\
 & cooked\_rabbit & 648 & 5.81\% \\
 & cooked\_porkchop & 668 & 5.99\% \\
 & cooked\_beef & 627 & 5.62\% \\
 & cooked\_chicken & 627 & 5.62\% \\
\midrule
\multicolumn{4}{l}{\textbf{Target Counts}} \\
 & 2 & 3,999 & 37.01\% \\
 & 3 & 3,363 & 31.13\% \\
 & 1 & 1,909 & 17.68\% \\
 & 4 & 1,532 & 14.18\% \\
\midrule
\multicolumn{4}{l}{\textbf{Tools}} \\
 & iron\_pickaxe & 18,633 & 29.69\% \\
 & iron\_shovel & 13,676 & 21.78\% \\
 & iron\_axe & 13,453 & 21.43\% \\
 & iron\_sword & 13,448 & 21.42\% \\
\midrule
\multicolumn{4}{l}{\textbf{Resource Types}} \\
 & red\_sand & 2,032 & 10.37\% \\
 & gold\_ore & 1,999 & 10.20\% \\
 & cobblestone & 1,915 & 9.77\% \\
 & sandstone & 1,818 & 9.28\% \\
 & iron\_ore & 1,780 & 9.08\% \\
 & coal\_ore & 1,714 & 8.75\% \\
 & acacia\_planks & 1,564 & 7.98\% \\
 & oak\_planks & 1,503 & 7.67\% \\
 & birch\_log & 1,486 & 7.58\% \\
 & spruce\_log & 1,477 & 7.54\% \\
 & oak\_log & 1,456 & 7.44\% \\
 & spruce\_planks & 1,471 & 7.51\% \\
 & birch\_planks & 1,344 & 6.86\% \\
 & sheep & 1,119 & 5.71\% \\
 & pig & 1,104 & 5.63\% \\
 & rabbit & 1,097 & 5.60\% \\
 & chicken & 1,081 & 5.52\% \\
 & cow & 700 & 3.57\% \\
 & mushroom & 675 & 3.44\% \\
\bottomrule
\end{tabular}
\end{minipage}
\end{table}

\clearpage

\section{Datasheet}

\datasheetsection{Motivation}

\textsf{For what purpose was the dataset created? Was there a specific task
in mind? Was there a specific gap that needed to be filled? Please provide
a description.}

The TeamCraft dataset was created to support development and evaluation for multi-modal multi-agent systems in MineCraft. 
\newline

\textsf{Who created the dataset (e.g., which team, research group) and on
behalf of which entity (e.g., company, institution, organization)?}

The dataset was created by the TeamCraft team.
\newline

\textsf{Who funded the creation of the dataset? If there is an associated
grant, please provide the name of the grantor and the grant name and number.}

The dataset was funded by the TeamCraft team.
\newline

\textsf{Any other comments?}

None.
\newline

\datasheetsection{Composition}

\textsf{What do the instances that comprise the dataset represent (e.g.,
documents, photos, people, countries)? Are there multiple types of
instances (e.g., movies, users, and ratings; people and interactions between them; nodes and edges)? Please provide a description.}

Each instance contains a ground-truth expert demonstration of a multi-agent team finishing a task in Minecraft, and the corresponding multi-modal prompts specifying the task. 
\newline

\textsf{How many instances are there in total (of each type, if appropriate)?}

There are in total 57,207 instances. 
\newline

\textsf{Does the dataset contain all possible instances or is it a sample
(not necessarily random) of instances from a larger set? If the
dataset is a sample, then what is the larger set? Is the sample representative of the larger set (e.g., geographic coverage)? If so, please describe how
this representativeness was validated/verified. If it is not representative
of the larger set, please describe why not (e.g., to cover a more diverse
range of instances, because instances were withheld or unavailable).}

The dataset contain all possible instances.
\newline

\textsf{What data does each instance consist of? “Raw” data (e.g., unprocessed text or images) or features? In either case, please provide a description.}

Each instance consists of a multi-modal task specification, agents observations and expert trajectories. Each task specification contains one raw language instruction and three orthographic views images. Agents observations contain the first-person view RGB images and the inventory information. 
\newline

\textsf{Is there a label or target associated with each instance? If so, please
provide a description.}

N/A.
\newline

\textsf{Is any information missing from individual instances? If so, please
provide a description, explaining why this information is missing (e.g.,
because it was unavailable). This does not include intentionally removed
information, but might include, e.g., redacted text.}

We intentionally removed the expert demonstration in the test set to prevent over-fitting. 
\newline

\textsf{Are relationships between individual instances made explicit
(e.g., users’ movie ratings, social network links)? If so, please describe how these relationships are made explicit.}

Each instance in the dataset corresponds to an individual task variant that belongs to one of the four task types (i.e. building, clearing, farming, smelting). The task type is explicitly specified in the file name. 
\newline

\textsf{Are there recommended data splits (e.g., training, development/validation, testing)? If so, please provide a description of these splits, explaining the rationale behind them.}

The dataset has been split into a training set (55,257 instances), a validation set (1,000 instances) and a test set (950 instances). The training set is designed for model training while the validation set is for hyperparameter tuning and checkpoint selection. The test set is designed to evaluate the model's generalization capabilities across novel scenes, novel goal states and novel agent numbers. 
\newline

\textsf{Are there any errors, sources of noise, or redundancies in the
dataset? If so, please provide a description.}

None as we know.
\newline

\textsf{Is the dataset self-contained, or does it link to or otherwise rely on
external resources (e.g., websites, tweets, other datasets)? If it links
to or relies on external resources, a) are there guarantees that they will
exist, and remain constant, over time; b) are there official archival versions
of the complete dataset (i.e., including the external resources as they
existed at the time the dataset was created); c) are there any restrictions
(e.g., licenses, fees) associated with any of the external resources that
might apply to a dataset consumer? Please provide descriptions of all
external resources and any restrictions associated with them, as well as
links or other access points, as appropriate.}

The dataset is entirely self-contained.
\newline

\textsf{Does the dataset contain data that might be considered confidential (e.g., data that is protected by legal privilege or by doctor–
patient confidentiality, data that includes the content of individuals’ non-public communications)? If so, please provide a description.}

None as we know.
\newline

\textsf{Does the dataset contain data that, if viewed directly, might be offensive, insulting, threatening, or might otherwise cause anxiety?
If so, please describe why.}

None as we know.
\newline

\textsf{Does the dataset identify any subpopulations (e.g., by age, gender)? If so, please describe how these subpopulations are identified and
provide a description of their respective distributions within the dataset.}

No.
\newline

\textsf{Is it possible to identify individuals (i.e., one or more natural persons), either directly or indirectly (i.e., in combination with other
data) from the dataset? If so, please describe how.}

No.
\newline

\textsf{Does the dataset contain data that might be considered sensitive
in any way (e.g., data that reveals race or ethnic origins, sexual orientations, religious beliefs, political opinions or union memberships, or locations; financial or health data; biometric or genetic
data; forms of government identification, such as social security
numbers; criminal history)? If so, please provide a description.}

None as we know.
\newline

\textsf{Any other comments?}

None.
\newline

\datasheetsection{Collection Process}

\textsf{How was the data associated with each instance acquired? Was
the data directly observable (e.g., raw text, movie ratings), reported by
subjects (e.g., survey responses), or indirectly inferred/derived from other
data (e.g., part-of-speech tags, model-based guesses for age or language)?
If the data was reported by subjects or indirectly inferred/derived from
other data, was the data validated/verified? If so, please describe how.}

In each data instance, the expert trajectory was generated programmatically via a planning algorithm. The language instruction was created by language templates. The orthographic views images and agent observations were collected in MineCraft. 
\newline

\textsf{What mechanisms or procedures were used to collect the data
(e.g., hardware apparatuses or sensors, manual human curation,
software programs, software APIs)? How were these mechanisms or
procedures validated?}

The data is automatically generated by running the data collection scripts. The procedure is further verified by the team via manual inspection.  
\newline

\textsf{If the dataset is a sample from a larger set, what was the sampling
strategy (e.g., deterministic, probabilistic with specific sampling
probabilities)?}

N/A.
\newline

\textsf{Who was involved in the data collection process (e.g., students,
crowdworkers, contractors) and how were they compensated (e.g.,
how much were crowdworkers paid)?}

Only the TeamCraft team members are voluntarily involved in the data collection process. 
\newline

\textsf{Over what timeframe was the data collected? Does this timeframe
match the creation timeframe of the data associated with the instances
(e.g., recent crawl of old news articles)? If not, please describe the timeframe in which the data associated with the instances was created.}

The data were collected between February 2024 and September 2024.
\newline

\textsf{Were any ethical review processes conducted (e.g., by an institutional review board)? If so, please provide a description of these review
processes, including the outcomes, as well as a link or other access point
to any supporting documentation.}

No.
\newline

\textsf{Did you collect the data from the individuals in question directly,
or obtain it via third parties or other sources (e.g., websites)?}

N/A.
\newline

\textsf{Were the individuals in question notified about the data collection? If so, please describe (or show with screenshots or other information) how notice was provided, and provide a link or other access point
to, or otherwise reproduce, the exact language of the notification itself.}

N/A.
\newline

\textsf{Did the individuals in question consent to the collection and use
of their data? If so, please describe (or show with screenshots or other
information) how consent was requested and provided, and provide a
link or other access point to, or otherwise reproduce, the exact language
to which the individuals consented.}

N/A.
\newline

\textsf{If consent was obtained, were the consenting individuals provided with a mechanism to revoke their consent in the future or for certain uses? If so, please provide a description, as well as a link or
other access point to the mechanism (if appropriate).}

N/A.
\newline

\textsf{Has an analysis of the potential impact of the dataset and its use
on data subjects (e.g., a data protection impact analysis) been conducted? If so, please provide a description of this analysis, including
the outcomes, as well as a link or other access point to any supporting
documentation.}

No.
\newline

\textsf{Any other comments?}

None.
\newline

\datasheetsection{Preprocessing/cleaning/labeling}

\textsf{Was any preprocessing/cleaning/labeling of the data done (e.g.,
discretization or bucketing, tokenization, part-of-speech tagging,
SIFT feature extraction, removal of instances, processing of missing values)? If so, please provide a description. If not, you may skip the
remaining questions in this section.}

Yes. In each data instance, the three orthographic views images rendered by MineCraft are manually concatenated as one image.
\newline

\textsf{Was the “raw” data saved in addition to the preprocessed/cleaned/labeled
data (e.g., to support unanticipated future uses)? If so, please provide a link or other access point to the “raw” data.}

No.
\newline

\textsf{Is the software that was used to preprocess/clean/label the data
available? If so, please provide a link or other access point.}

N/A.
\newline

\textsf{Any other comments?}

None.
\newline

\datasheetsection{Uses}

\textsf{Has the dataset been used for any tasks already? If so, please provide
a description.}

The dataset is used to develop the TeamCraft-VLA model, as described in this paper.  
\newline

\textsf{Is there a repository that links to any or all papers or systems that
use the dataset? If so, please provide a link or other access point.}

No.
\newline

\textsf{What (other) tasks could the dataset be used for?}

This dataset can be used for the development and evaluation of multi-modal multi-agent systems in MineCraft. 
\newline

\textsf{Is there anything about the composition of the dataset or the way
it was collected and preprocessed/cleaned/labeled that might impact future uses? For example, is there anything that a dataset consumer
might need to know to avoid uses that could result in unfair treatment of
individuals or groups (e.g., stereotyping, quality of service issues) or other
risks or harms (e.g., legal risks, financial harms)? If so, please provide a
description. Is there anything a dataset consumer could do to mitigate
these risks or harms?}

Unknown to the authors of the datasheet.
\newline

\textsf{Are there tasks for which the dataset should not be used? If so,
please provide a description.}

Unknown to the authors of the datasheet.
\newline

\textsf{Any other comments?}

None.
\newline

\datasheetsection{Distribution}

\textsf{Will the dataset be distributed to third parties outside of the entity (e.g., company, institution, organization) on behalf of which
the dataset was created? If so, please provide a description.}

Yes, the dataset is available on the Internet.
\newline

\textsf{How will the dataset will be distributed (e.g., tarball on website,
API, GitHub)? Does the dataset have a digital object identifier (DOI)?}

The dataset will be available on \href{https://huggingface.co/datasets/teamcraft/teamcraft_data}{Huggingface}. It does not have a DOI.
\newline

\textsf{When will the dataset be distributed?}

The dataset will be available online by 12/01/2024.
\newline

\balance

\textsf{Will the dataset be distributed under a copyright or other intellectual property (IP) license, and/or under applicable terms of use
(ToU)? If so, please describe this license and/or ToU, and provide a link
or other access point to, or otherwise reproduce, any relevant licensing
terms or ToU, as well as any fees associated with these restrictions.}

The dataset is under Apache 2.0 license.
\newline

\textsf{Have any third parties imposed IP-based or other restrictions on
the data associated with the instances? If so, please describe these
restrictions, and provide a link or other access point to, or otherwise
reproduce, any relevant licensing terms, as well as any fees associated
with these restrictions.}

None as we know. 
\newline

\textsf{Do any export controls or other regulatory restrictions apply to
the dataset or to individual instances? If so, please describe these
restrictions, and provide a link or other access point to, or otherwise
reproduce, any supporting documentation.}

Unknown to authors of the datasheet.
\newline

\textsf{Any other comments?}

None.
\newline

\datasheetsection{Maintenance}

\textsf{Who will be supporting/hosting/maintaining the dataset?}

The TeamCraft team will be maintaining the dataset. 
\newline

\textsf{How can the owner/curator/manager of the dataset be contacted
(e.g., email address)?}

Email: teamcraftbench@gmail.com
\newline

\textsf{Is there an erratum? If so, please provide a link or other access point.}

No.
\newline

\textsf{Will the dataset be updated (e.g., to correct labeling errors, add
new instances, delete instances)? If so, please describe how often, by
whom, and how updates will be communicated to dataset consumers
(e.g., mailing list, GitHub)?}

No planned updates at the time of preparing this datasheet.
\newline

\textsf{If the dataset relates to people, are there applicable limits on the
retention of the data associated with the instances (e.g., were the
individuals in question told that their data would be retained for
a fixed period of time and then deleted)? If so, please describe these
limits and explain how they will be enforced.}

Unknown to authors of the datasheet.
\newline

\textsf{Will older versions of the dataset continue to be supported/hosted/maintained?
If so, please describe how. If not, please describe how its obsolescence
will be communicated to dataset consumers.}

N/A.
\newline

\textsf{If others want to extend/augment/build on/contribute to the
dataset, is there a mechanism for them to do so? If so, please
provide a description. Will these contributions be validated/verified? If
so, please describe how. If not, why not? Is there a process for communicating/distributing these contributions to dataset consumers? If so,
please provide a description.}

Others may do so and should contact the original authors about incorporating fixes/extensions.
\newline

\textsf{Any other comments?}

None.
\newline